%% file: main.tex
\documentclass{article}

    \PassOptionsToPackage{numbers, compress}{natbib}

 \usepackage[main, final]{neurips_2026}

\usepackage[utf8]{inputenc} 
\usepackage[T1]{fontenc}    
\usepackage{hyperref}       
\usepackage{url}            
\usepackage{booktabs}       
\usepackage{amsfonts}       
\usepackage{nicefrac}       
\usepackage{microtype}      
\usepackage{xcolor}         

\input{preamble}
\title{Hierarchical Semantic Tree Anchoring for\\ CLIP-based Class-Incremental Learning}

\author{%
 {Tao Hu$^{1,2}$  \quad
Lan Li$^{1,2}$    \quad
Zhen-Hao Xie$^{1,2}$    \quad
Da-Wei Zhou$^{1,2}\textsuperscript{(\Letter)}$} \\
$^{1} $ School of Artificial Intelligence, Nanjing University \\
$^{2} $ State Key Laboratory for Novel Software Technology, Nanjing University \\
\texttt{\small \{hut, lil, wenzh, zhoudw\}@lamda.nju.edu.cn}
}

\begin{document}
\maketitle

\input{sec/0_abstract}

\input{sec/1_inroduction}

\input{sec/2_related_work}
\input{sec/3_preliminaries}

\input{sec/4_method}

\input{sec/5_experiments}

\input{sec/6_conclusion}

{
    \small
    \bibliographystyle{ieeenat_fullname}
    \bibliography{main}
}

\input{sec/X_suppl}
\input{checklist.tex}
\end{document}

%% file: preamble.tex
\newcommand{\name}{\textsc{Hasten}\xspace}
\newcommand{\mame}{\textsc{Hasten}\xspace}

\usepackage{multirow}
\usepackage{amsmath,amssymb}
\usepackage{mathtools}
\usepackage{xfrac}
\usepackage[english]{babel}
\usepackage{hyphenat}
\usepackage{tabularx} 
\usepackage{xcolor} 
\usepackage{colortbl}
\usepackage{soul}
\usepackage{csquotes}
\usepackage{listings}
\newcommand{\blue}[1]{\textcolor{blue}{#1}}
\usepackage{seqsplit}

\usepackage[export]{adjustbox}
\usepackage[most]{tcolorbox}
\usepackage{listings}
\usepackage{xcolor}
\usepackage{enumitem}
\usepackage{marvosym}
\definecolor{promptbg}{RGB}{248,250,255}
\definecolor{promptframe}{RGB}{120,145,190}

\definecolor{answerbg}{RGB}{248,253,248}
\definecolor{answerframe}{RGB}{120,170,125}

\definecolor{incpromptbg}{RGB}{255,251,245}
\definecolor{incpromptframe}{RGB}{215,165,105}

\definecolor{rulebg}{RGB}{253,253,253}
\definecolor{rulegray}{RGB}{180,180,180}

\definecolor{boxtitlegray}{RGB}{45,45,45}

\definecolor{jsonbg}{RGB}{252,252,252}
\definecolor{jsonkey}{RGB}{130,45,45}
\definecolor{jsonstr}{RGB}{35,85,135}

\lstdefinestyle{jsonstyle}{
  basicstyle=\ttfamily\scriptsize,
  backgroundcolor=\color{jsonbg},
  keywordstyle=\color{jsonkey},
  stringstyle=\color{jsonstr},
  showstringspaces=false,
  breaklines=true,
  breakatwhitespace=false,
  columns=fullflexible,
  keepspaces=true,
  upquote=true,
  frame=none,
  xleftmargin=2pt,
  xrightmargin=2pt,
  aboveskip=2pt,
  belowskip=2pt,
  literate=
    *{:}{{{\color{black}{:}}}}{1}
     {,}{{{\color{black}{,}}}}{1}
     {\{}{{{\color{black}{\{}}}}{1}
     {\}}{{{\color{black}{\}}}}}{1}
     {[}{{{\color{black}{[}}}}{1}
     {]}{{{\color{black}{]}}}}{1},
}

\tcbset{
  promptlikebox/.style={
    enhanced,
    breakable,
    boxrule=0.45pt,
    arc=2pt,
    left=4pt,
    right=4pt,
    top=2pt,
    bottom=2pt,
    toptitle=2pt,
    bottomtitle=2pt,
    before skip=3pt,
    after skip=3pt,
    fonttitle=\bfseries\sffamily\footnotesize,
    coltitle=black,
    title filled=false,
  }
}

\newtcolorbox{promptbox}[1][]{%
  promptlikebox,
  colback=promptbg,
  colframe=promptframe,
  title={#1},
}

\newtcolorbox{answerbox}[1][]{%
  promptlikebox,
  colback=answerbg,
  colframe=answerframe,
  title={#1},
}

\newtcolorbox{incpromptbox}[1][]{%
  promptlikebox,
  colback=incpromptbg,
  colframe=incpromptframe,
  title={#1},
}

\newtcolorbox{rulebox}{%
  enhanced,
  breakable,
  colback=rulebg,
  colframe=rulegray,
  boxrule=0.35pt,
  arc=1.5pt,
  left=4pt,
  right=4pt,
  top=2pt,
  bottom=2pt,
  before skip=2pt,
  after skip=2pt,
}

\usepackage{ragged2e}
\usepackage{xstring}
\usepackage{microtype}

\usepackage{etoolbox}

\AtBeginEnvironment{displayquote}{%
  \setlength{\topsep}{0pt}%
  \setlength{\partopsep}{0pt}%
  \setlength{\parsep}{0pt}%
  \setlength{\itemsep}{0pt}%
}
\BeforeBeginEnvironment{displayquote}{\vspace{-3.5pt}}
\AfterEndEnvironment{displayquote}{\vspace{-2.5pt}}

\usepackage{xparse}
\usepackage{amsthm}
\newtheorem{proposition}{Proposition}
\NewDocumentEnvironment{spacedequation}{O{2pt} O{3pt}}{%
  \begingroup
  \setlength{\abovedisplayskip}{#1}%
  \setlength{\abovedisplayshortskip}{#1}%
  \setlength{\belowdisplayskip}{#2}%
  \setlength{\belowdisplayshortskip}{#2}%
  \begin{equation}
}{%
  \end{equation}
  \endgroup
  \ignorespacesafterend
}
\usepackage{tabularx}
\usepackage{array}
\usepackage{caption}
\usepackage{subcaption}
\usepackage{colortbl}
\usepackage{nicematrix}

\newcommand{\vz}{\mathbf{z}}

\usepackage{wrapfig}
\definecolor{Gray}{gray}{0.85}
\definecolor{Redo}{rgb}{0.95,0.69,0.51}
\definecolor{LightCyan}{rgb}{0.88,1,1}
\usepackage{afterpage}
\usepackage{pifont}
\usepackage{csquotes}
\usepackage{lipsum}

\usepackage{xspace}
\usepackage{subcaption}
\usepackage{multirow}
\usepackage{amsmath,amssymb}
\usepackage{mathtools}
\usepackage{amsfonts}
\usepackage{bm}
\usepackage{xfrac}
\usepackage[english]{babel}
\usepackage{hyphenat}
\usepackage{tabularx}
\usepackage{xcolor}
\usepackage{colortbl}
\usepackage{soul}
\usepackage{csquotes}
\usepackage{listings}
\usepackage{seqsplit}
\usepackage{arydshln}
\usepackage{wrapfig}
\usepackage{enumitem}
\usepackage{algorithm}
\usepackage{algorithmic}
\usepackage{graphicx}
\usepackage{afterpage}
\usepackage{pifont}
\usepackage{lipsum}

\newcommand{\eg}{\textit{e.g.}}
\newcommand{\ie}{\textit{i.e.}}
\usepackage{amsmath}
\usepackage{amssymb}
\usepackage{cleveref}
\usepackage[dvipsnames]{xcolor}

%% file: sec/0_abstract.tex
\begin{abstract}
Class-Incremental Learning (CIL) enables models to learn new classes continually while preserving past knowledge. 
Recently, vision-language models like CLIP offer transferable features via multimodal pre-training, making them well-suited for CIL. 
However, real-world visual and linguistic concepts are inherently hierarchical: a textual concept like ``dog'' subsumes fine-grained categories such as ``Labrador'' and ``Golden Retriever,'' and each category is associated with corresponding images. 
Nevertheless, existing CLIP-based CIL methods fail to explicitly capture this inherent hierarchy, leading to feature drift of fine-grained classes during incremental updates and ultimately to catastrophic forgetting. 
To address this challenge, we propose \name (Hierarchical Semantic Tree Anchoring), a hierarchy-aware framework that uses semantic structure to stabilize CLIP-based CIL. 
Rather than treating classes as isolated labels, \name leverages external hierarchical knowledge as structured supervision to organize visual and textual features in hyperbolic space, helping maintain parent--child relations as new tasks arrive and mitigating feature drift. 
Since the shared hyperbolic mapper is updated across tasks, we further stabilize it by constraining its updates to a null space induced by prior-task features, reducing interference with previous mappings while retaining adaptability to new classes. 
Extensive experiments on nine benchmarks show that \name consistently outperforms existing methods while reducing feature drift and catastrophic forgetting. Code is available at: \url{https://github.com/LAMDA-CL/NeurIPS2026-HASTEN}.

\end{abstract}

%% file: sec/1_inroduction.tex
\section{Introduction}
\label{sec:intro}
Recent advancements in deep learning~\cite{Hinton06, deng2009imagenet, lecun2015deep, he2015residual, goodfellow2016deep} have driven progress in vision and language tasks, but real-world applications face challenges with non-stationary streaming data, requiring adaptation to new data while retaining prior knowledge~\cite{aggarwal2018survey, li2024configure}. Class-Incremental Learning (CIL)~\cite{rebuffi2017icarl,zhao2020maintaining, masana2022class} enables incremental learning of new classes without forgetting earlier knowledge, yet suffers from catastrophic forgetting~\cite{french1999catastrophic,serra2018overcoming, ramasesh2021effect, shi2021overcoming}, where new tasks interfere with old ones. Recent pre-trained models (PTMs) like CLIP~\cite{radford2021learning} offer generalizable, transferable features from large-scale pre-training~\cite{zhou2022learning,zhou2022conditional,huang2024class,d2023multimodal}, providing a strong foundation for continual learning and fast adaptation to new tasks via embedded prior knowledge.\looseness -1

Recent studies have attempted to address catastrophic forgetting~\cite{yu2024boosting, dohare2024loss, wang2024comprehensive, lai2025order} by freezing the CLIP backbone to preserve general knowledge while adapting lightweight components (\eg, adapters~\cite{huang2024class, zhou2023revisiting} or prompts~\cite{wang2022learning, zhou2023learning}). Although these approaches mitigate forgetting to some extent, natural classes are inherently {\em hierarchical}. For example, ``dog'' subsumes ``Labrador'' and ``Golden Retriever,'' and each fine-grained class has its own image set. However, current CLIP-based CIL methods often overlook this hierarchical signal because CLIP’s Euclidean feature space is ill-suited to modeling hierarchy~\cite{desai2023hyperbolic, gao2022pyramidclip}. Its uniform metric treats all points alike, so placing a generic concept near many related items compresses distances and inadvertently pulls in unrelated specific concepts. As a result, broad concepts become too close to many specifics, blurring boundaries between hierarchical levels and weakening the representation of structure.\looseness -1

\begin{wrapfigure}{l}{0.45\columnwidth}
    \centering
    \includegraphics[width=\linewidth]{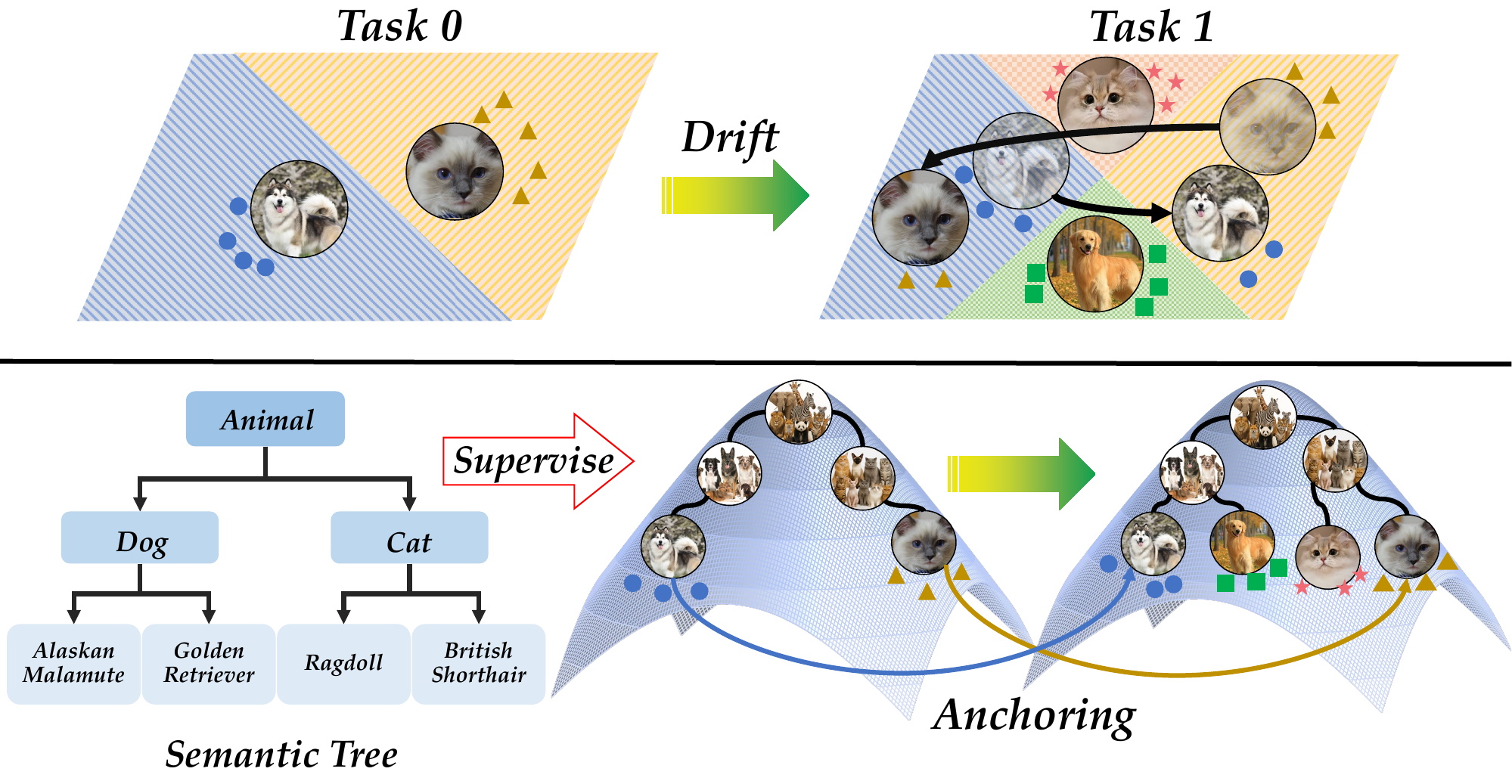}
\caption{\small Effect of feature hierarchy. \textbf{Top:} Without hierarchy, fine-grained features tend to drift as new classes emerge. \textbf{Bottom:} With a hierarchical constraint, features are better anchored and relations are more stable during incremental learning.
}
\label{fig:drift_hyperbolic}
\end{wrapfigure}

This hierarchy is especially useful in incremental learning, where preserving relations among old and new classes can help mitigate catastrophic forgetting. As illustrated in Figure~\ref{fig:drift_hyperbolic} (Top), without hierarchy, fine-grained class features (\eg, specific dog and cat breeds) gradually drift~\cite{yu2020semantic} across incremental tasks. As new classes emerge, previous class features will shift from their original positions and even incorporate features of unrelated classes.
In contrast, Figure~\ref{fig:drift_hyperbolic} (Bottom) shows that with explicit hierarchical constraints, features can be anchored in a structured hyperbolic space, thereby suppressing drift during continual updates.\looseness -1

These observations motivate a stronger hierarchy-aware inductive bias. Without hierarchical constraints, early-class features lack effective protection and are gradually eroded by training on new classes. An ideal incremental learner should provide stable anchors for latent parent--child relations, keeping siblings compact yet separable and preventing broad concepts from being crowded by fine-grained ones. Meanwhile, its updates should respect the existing hierarchical organization while remaining plastic to new classes. Such a design can stabilize accumulated knowledge and mitigate catastrophic forgetting.\looseness=-1

Based on this insight, we propose \name (HierArchical Semantic TreE Anchoring), a hierarchy-aware framework that incorporates hierarchical semantic structure into CLIP-based continual learning. Concretely, \name first uses an external hierarchical semantic tree as structured supervision to model text--text and image--text relations through two hierarchy-aware modules. The enhanced features are then mapped by a shared hyperbolic mapper into hyperbolic space, where hierarchical relations can be more naturally encoded and maintained across tasks. However, because this mapper bridges all tasks and modalities while being continuously updated, unconstrained updates may disturb previously learned mappings. We therefore constrain its updates to the null space induced by prior-task features, reducing interference with previous mappings while retaining adaptability to new classes. In this way, \name maintains hierarchical structure while stabilizing the shared mapper, jointly mitigating feature drift and catastrophic forgetting throughout incremental learning.\looseness -1

%% file: sec/2_related_work.tex
\section{Related Work}
\label{sec:related_work}
\noindent\textbf{Class-Incremental Learning (CIL)}:
CIL enables models to incrementally learn new classes without forgetting previously acquired knowledge~\cite{masana2022class,de2021survey}, and traditional approaches can be broadly categorized into several groups. Replay-based methods preserve past knowledge by storing and revisiting samples from previous tasks~\cite{rebuffi2017icarl,castro2018end,chaudhry2018efficient,luo2023class}, though this may come at the cost of memory usage and data privacy. Regularization-based methods estimate parameter importance to constrain updates~\cite{kirkpatrick2017overcoming,zenke2017continual,aljundi2018memory,aljundi2019task} or use knowledge distillation to preserve information at the logit~\cite{rebuffi2017icarl,li2017learning}, feature~\cite{hou2019learning,lu2022augmented,park2021class}, or group level~\cite{douillard2020podnet,gao2022rdfcil,dong2021few}. Architecture-based methods mitigate forgetting through feature or module expansion~\cite{yan2021dynamically,wang2022foster,wang2023beef,zheng2025task} and dynamically introduced task-specific tokens~\cite{douillard2022dytox}, balancing knowledge retention and adaptability at the cost of additional model complexity. Bias rectification methods address prediction and classifier biases~\cite{shi2022mimicking,zhao2020maintaining,wu2019large}.

\textbf{CIL with Pre-Trained Models}: The paradigm of CIL has been revitalized by powerful pre-trained models (PTMs) like Vision Transformers~\cite{dosovitskiy2020image} and CLIP~\cite{radford2021learning}, which offer strong, generalizable representations~\cite{wang2022learning}. A prevalent strategy is to freeze the PTM backbone and introduce lightweight, learnable modules for adaptation. For vision-centric PTMs, research has explored visual prompt learning~\cite{wang2022learning,smith2023coda,jung2023generating,li2025addressing}, adapter-based tuning~\cite{rebuffi2017learning,chen2022adaptformer,yu2024boosting,qi2025adaptive}, and prototype-based classifiers that match features to class prototypes in embedding space~\cite{zhou2023revisiting,mcdonnell2023ranpac,snell2017prototypical}. Building on this, a significant trend leverages vision-language models like CLIP to exploit cross-modal alignment~\cite{li2026bofa}. This is achieved through various means, including learning multimodal or textual prompts~\cite{wang2022dualprompt,wang2023attriclip,zhou2022learning}, adapting the architecture with new projection heads as in PROOF~\cite{zhou2023learning}, or using parameter-efficient adapters as in RAPF~\cite{huang2024class} to balance stability and plasticity. Recent studies have also explored leveraging external models to extract and incorporate fine-grained representations. ENGINE~\cite{zhou2025external} extracts and aligns fine-grained features through a dual-branch architecture, while AREA~\citep{xie2026area} learns to aggregate external features without forgetting.

\textbf{Hyperbolic Geometry for Hierarchical Representation}:
Hyperbolic geometry provides a natural inductive bias for hierarchical modeling, as its negative curvature and exponential volume growth enable embeddings of tree-like or partial-order structures with substantially lower distortion than in Euclidean spaces~\cite{gromov1987hyperbolic,cannon1997hyperbolic}. 
In representation learning, a line of work has exploited hyperbolic manifolds (e.g., Poincar\'e ball and Lorentz model) to learn continuous hierarchies and taxonomies, showing strong advantages on structured data such as graphs and knowledge bases~\cite{nickel2017poincare,nickel2018learning,ganea2018hyperbolic}. 
Beyond distance-based embeddings, order- and entailment-aware formulations further connect hierarchy to geometry by enforcing partial-order relations via specialized losses or cone-based constraints, enabling more explicit supervision of parent--child semantics~\cite{vendrov2015order,ganea2018hyperbolic}.
These ideas have been generalized to deep models, including hyperbolic neural networks and hyperbolic graph representation learning, and have been systematically reviewed in recent surveys~\cite{peng2021hyperbolic}. 
More recently, hyperbolic representations have also been explored in vision-language modeling to better organize multimodal semantics under hierarchical or safety-aware structures~\cite{poppi2025hyperbolic}.

%% file: sec/3_preliminaries.tex
\section{Preliminaries}
\label{sec:prelim}
\noindent{\bf Class-Incremental Learning (CIL):}
CIL aims to build a classifier that learns new classes from sequential tasks without forgetting previously learned ones~\cite{rebuffi2017icarl}. We consider a sequence of tasks $\{\mathcal{D}^1, \mathcal{D}^2, \dots, \mathcal{D}^B\}$, where each task $\mathcal{D}^b = \{(\mathbf{x}_i, y_i)\}_{i=1}^{n_b}$ contains $n_b$ instances. Here, $\mathbf{x}_i \in \mathbb{R}^D$ represents the feature vector, and $y_i \in Y_b$ is the class label for task $b$. The label sets across tasks are disjoint, \ie, $Y_b \cap Y_{b'} = \varnothing$ for any $b \neq b'$. 
The exemplar-free CIL setting~\cite{zhu2021prototype, wang2022learning} prohibits storing or replaying data from previous tasks, which means the model can only access the data from the current task $\mathcal{D}^b$ during training. The goal is to learn a unified model $f$ that generalizes to all seen classes $\mathcal{Y}_b = Y_1 \cup \dots \cup Y_b$ and minimizes the empirical risk:
\begin{equation} \label{eq:totalrisk} 
    f^* = \underset{f \in \mathcal{H}}{\operatorname{argmin}} \; \mathbb{E}_{(\mathbf{x}, y) \in \mathcal{D}^1 \cup \dots \cup \mathcal{D}^b} \left[ \mathbb{I}(y \neq f(\mathbf{x})) \right] \,,
\end{equation}
where $\mathcal{H}$ is the hypothesis space, and $\mathbb{I}(\cdot)$ is the indicator function. In this paper, following recent work~\cite{zhou2023learning,yu2024boosting,huang2024class}, we build our model $f(\mathbf{I})$ upon a pre-trained CLIP model~\cite{radford2021learning}. CLIP consists of a visual encoder $g_v$ and a text encoder $g_t$. For a given image $\mathbf{I}$, the visual embedding is $\mathbf{e}_v = g_v(\mathbf{I})$. For each class $c$, we construct a class-specific prompt $\mathbf{t}_c$ (e.g., ``a photo of a [CLASS]$_c$''), and the textual embedding is $\mathbf{e}_t^c = {g}_t(\mathbf{t}_c)$. The image $\mathbf{I}$ is classified by comparing the cosine similarity between $\mathbf{e}_v$ and $\mathbf{e}_t^c$ of all seen classes $\mathcal{Y}_b$:
\begingroup
\setlength{\abovedisplayskip}{-3pt}
\begin{equation} \label{eq:clip_pred}
    P_c(\mathbf{I}) = \frac{\exp \left( \text{cos}(\mathbf{e}_v, \mathbf{e}_t^c) / \tau \right)}{\sum_{c^\prime \in \mathcal{Y}_b} \exp \left( \text{cos}(\mathbf{e}_v, \mathbf{e}_t^{c^\prime}) / \tau \right)}, 
\end{equation}
\endgroup
where $\tau$ is a temperature parameter that controls the softness of the distribution.
\\{\bf Adapting CLIP for CIL:} 
\label{subsec:clip_cil}
A common approach in CLIP-based CIL is to freeze the pre-trained visual and text encoders ($g_v$ and $g_t$) and insert a lightweight adaptation module $h_v$ after the visual encoder~\cite{huang2024class,zhou2023learning}. In this setup, the adapted image feature is represented as:
\begin{equation}\label{eq:rapf}
\mathbf{z}_v = h_v(\mathbf{e}_v) \,,
\end{equation}
where $\mathbf{e}_v = g_v(\mathbf{I})$ is the frozen visual embedding. This adapted feature is then substituted for $\mathbf{e}_v$ in~Eq.~\ref{eq:clip_pred} to compute classification probabilities, which are subsequently used to calculate the contrastive loss during training. This strategy allows the model to learn downstream knowledge while mitigating catastrophic forgetting effects by restricting the number of trainable parameters.
\\{\bf Discussion:} Although Eq.~\ref{eq:rapf} mitigates forgetting via lightweight tuning, the adapter-only tuning approach is intrinsically fragile under sequential updates. Continual revisions across tasks will induce class-feature drift~\cite{yu2020semantic}, which progressively erodes the text–image correspondence and breaks the cross-modal alignment required by Eq.~\ref{eq:clip_pred}. This accumulation of unreliable logits and visual–textual mismatches ultimately results in catastrophic forgetting. To counteract this, an effective CLIP adaptation must explicitly encode an ordered hierarchical structure to organize and anchor visual and textual features across all tasks, thus curbing drift and preserving crucial cross-modal alignment.

%% file: sec/4_method.tex
\begin{figure*}[t]
	\centering
    {\includegraphics[width=1\columnwidth]{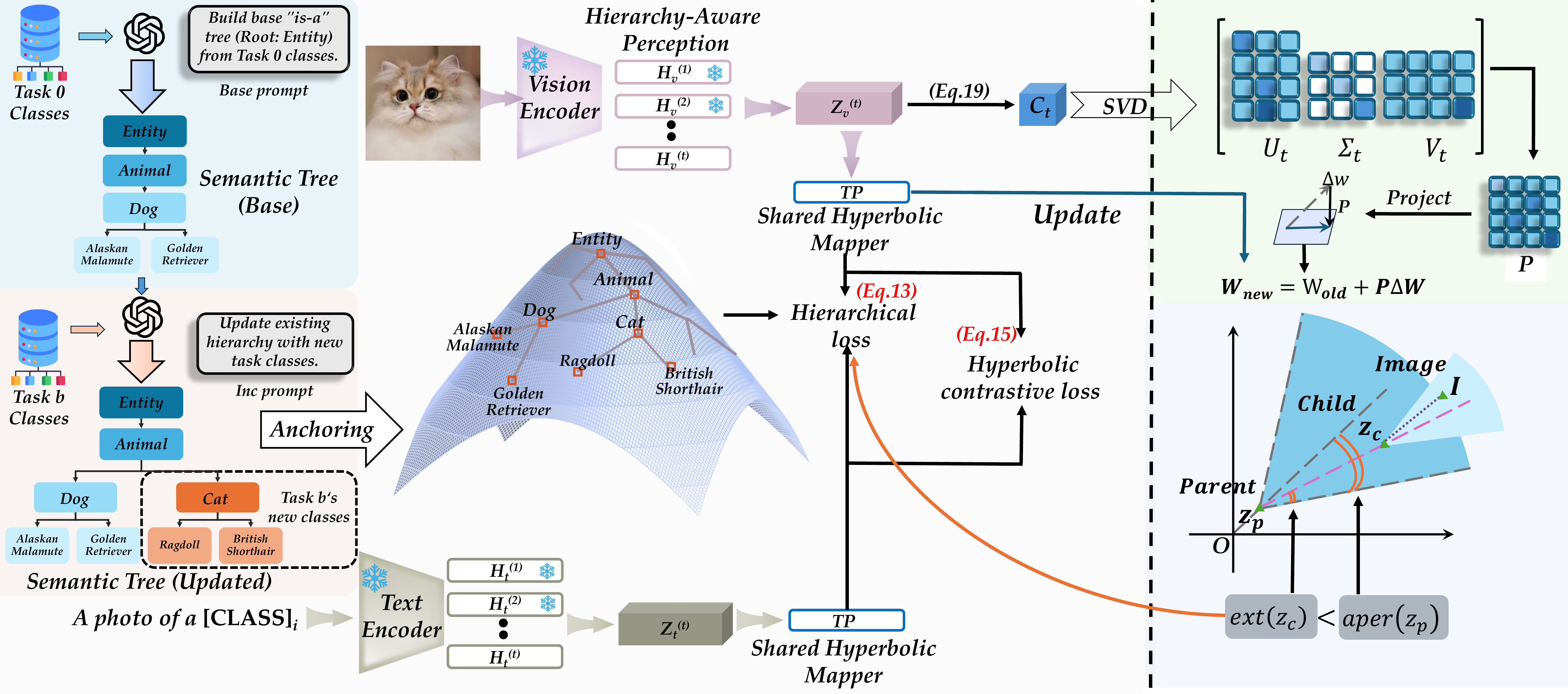}}
	\vspace{-3mm}
	\caption{Illustration of \mame. {\bf Left}: Hierarchical semantic tree building and hyperbolic projection. We use GPT-5 to generate a tree-structured semantic hierarchy and design task-specific hierarchical perception modules to meet downstream knowledge requirements. Euclidean features are projected into hyperbolic space via a global hyperbolic mapping layer. {\bf Top-Right}: Null space projection of the TP layer, ensuring TP does not interfere with the outputs of old tasks during updates. {\bf Bottom-Right}: Illustration of the entailment loss in $\mathcal{B}^2$. This loss pushes the embedding $\vz_c$ within an imaginary cone projected by its paired parent embedding $\vz_p$.}
	\vspace{-3mm}
	\label{figure:teaser}
\end{figure*}
\section{{\textsc{Hasten}}: Hierarchical Semantic Tree Anchoring}

Motivated by feature drift in CLIP-based CIL, we propose \name, a hierarchy-aware framework that uses semantic structure to organize visual and textual features for mitigating catastrophic forgetting. 
\name first builds an externally generated hierarchical semantic tree as structured supervision, allowing classes to be modeled through parent-child relations rather than isolated labels. 
The hierarchy-aware perception modules and shared hyperbolic mapper then embed visual and textual features into hyperbolic space to maintain hierarchical relations across tasks. 
To stabilize this shared mapper, we project its gradients onto an approximate null space induced by prior-task features, reducing interference with previous mappings while preserving adaptability to new classes.

\subsection{Hierarchical Semantic Tree Construction}
\label{subsec:semantic_tree}
In the real world, visual and linguistic concepts naturally form hierarchies, where broad concepts subsume fine-grained subclasses and images correspond to leaf-level instances. To model this structure under the CIL protocol, we construct a dataset-conditioned semantic tree in an incremental manner, using only the class names available at each stage and without accessing future classes.
\\\textbf{Base Tree Initialization ($t=0$):} 
At the initial stage, we use GPT-5 to build an \emph{is-a} hierarchy from the base classes, with a virtual root ``entity''. GPT-5 may introduce abstract virtual classes to connect dataset-native real classes into a coherent tree. For example:
\begin{displayquote}
{\bf Q}: Build a complete \emph{is-a} hierarchy with a single root ``entity'' from the base class list. You may introduce abstract parent classes if needed. Input list: [``ragdoll'', ``british shorthair'']\\
{\bf A}: 
$\begin{aligned}[t]
\{ &\text{``\sethlcolor{Apricot!100}\hl{entity}'': [``\sethlcolor{Apricot!60}\hl{animal}'']},\\
   &\text{``\sethlcolor{Apricot!60}\hl{animal}'': [``\sethlcolor{Apricot!40}\hl{cat}'']},\\
   &\text{``\sethlcolor{Apricot!40}\hl{cat}'': [``\sethlcolor{Apricot!20}\hl{ragdoll}'', ``\sethlcolor{Apricot!20}\hl{british shorthair}'']} \}
\end{aligned}$
\end{displayquote}

\textbf{Incremental Tree Expansion ($t>0$):} 
For later stages, we expand the existing hierarchy by appending only the newly observed classes. Existing parent-child edges are kept fixed, and new virtual parents are added only when necessary to connect new classes to existing branches.
\begin{displayquote}
{\bf Q}: Given an existing taxonomy tree, integrate the new classes without removing or altering existing nodes. Existing tree: \{``entity''\!: [``animal''], ``animal''\!: [``cat''], ``cat''\!: [``ragdoll'', ``british shorthair'']\}. New classes: [``golden retriever'', ``alaskan malamute'']\\
{\bf A}: 
$\begin{aligned}[t]
\{ &\text{``\sethlcolor{Apricot!100}\hl{entity}'': [``\sethlcolor{Apricot!60}\hl{animal}'']},\\
   &\text{``\sethlcolor{Apricot!60}\hl{animal}'': [``\sethlcolor{Apricot!40}\hl{cat}'', ``\sethlcolor{Apricot!40}\hl{dog}'']},\\
   &\text{``\sethlcolor{Apricot!40}\hl{cat}'': [``\sethlcolor{Apricot!20}\hl{ragdoll}'', ``\sethlcolor{Apricot!20}\hl{british shorthair}'']},\\
   &\text{``\sethlcolor{Apricot!40}\hl{dog}'': [``\sethlcolor{Apricot!20}\hl{golden retriever}'', ``\sethlcolor{Apricot!20}\hl{alaskan malamute}'']} \}
\end{aligned}$
\end{displayquote}

Through this process, we obtain a model-agnostic hierarchical supervision signal at each learning stage. We denote the current semantic tree as $\mathcal{K}=(\mathcal{C},\mathcal{E},r)$, where $\mathcal{C}=\mathcal{C}_{\mathrm{real}}\cup\mathcal{C}_{\mathrm{virt}}$ contains real and virtual classes, $r=\text{``entity''}$ is the root, and $\mathcal{E}\subseteq\mathcal{C}\times\mathcal{C}$ contains directed parent$\to$child edges. This tree defines the text-to-text hierarchy used for subsequent hierarchy-aware feature learning.

\subsection{Hierarchy-Aware Anchoring}
With the GPT-generated hierarchical semantic tree in hand, we leverage it during training so that classes explicitly inherit parent–child structure. This choice aligns with hyperbolic geometry. Theoretically, hyperbolic space can embed tree-structured data with minimal distortion~\cite{sarkar2011low,desai2023hyperbolic}, making it a natural fit for representing and preserving hierarchical semantics throughout learning.
\\{\bf Hyperbolic Geometry Setup}: Concretely, we adopt the Lorentz model $\mathcal{B}^d$ (embedded in $\mathbb{R}^{d+1}$) to instantiate our hyperbolic representation, where $d$ matches the Euclidean embedding dimension. A point $\mathbf{p} \in \mathcal{B}^d$ is parameterized as $[\mathbf{p}_{s}, p_{t}]$, where $p_{t} = \sqrt{\tfrac{1}{c}+\|\mathbf{p}_{s}\|^2}$. 
The resulting geodesic distance $d_{\mathcal{B}}(\mathbf{p},\mathbf{q})$ in $\mathcal{B}^d$ is:
\begin{spacedequation}[-3mm]\label{eq:lorenz_dist}
    d_{\mathcal{B}}(\mathbf{p},\mathbf{q}) = \frac{1}{\sqrt{c}}\cosh^{-1}\!\bigl(-c (\langle \mathbf{p}_{s},\mathbf{q}_{s}\rangle - p_{t}q_{t})\bigr).
\end{spacedequation}
Here, $c>0$ denotes the curvature, which is set to $1$ for stability. Negative curvature provides exponential volume growth and a depth--radius correspondence, naturally matching tree-structured data; a detailed justification is given in Appendix~\ref{app:hyperbolic_hierarchy}. In contrast, Euclidean volume grows only polynomially and may cause crowding near generic concepts. In our model, $d_{\mathcal{B}}$ serves as the task metric for similarity, margins, and hierarchy constraints. We use the spatial component of the exponential map at the origin $\mathbf{o}=[\mathbf{0},1]$ to map a Euclidean tangent vector $\mathbf{v}$:
\begin{spacedequation}
    \mathrm{expm}_{\mathbf{o}}(\mathbf{v})=\frac{\sinh(\|\mathbf{v}\|)}{\|\mathbf{v}\|}\mathbf{v}.
\end{spacedequation}
These components provide the necessary hyperbolic similarities and geodesics that respect hierarchical geometry, which we utilize for subsequent alignment and drift control.
\\\textbf{Hierarchy-Aware Perception and Aggregation}: To encode hierarchical information while retaining CLIP's generalization power, we introduce two \emph{task-specific} linear hierarchy-aware modules, $H_v^b$ (for visual features) and $H_t^b$ (for textual features), immediately following the frozen CLIP encoders ${g}_v$ and ${g}_t$ for each incremental task $b$. The modules map from $\mathbb{R}^d$ to $\mathbb{R}^d$. Given a text $\mathbf{t}$ and an image $\mathbf{I}$, the features in Euclidean space are $\mathbf{e}_t$ and $\mathbf{e}_v$. During task $b$, only the new modules $\{H_v^b, H_t^b\}$ are trained, while all earlier modules $\{H_v^1,\dots,H_v^{b-1}\}$ and $\{H_t^1,\dots,H_t^{b-1}\}$ are frozen to preserve prior knowledge.
We cumulatively aggregate the outputs of all modules to obtain the task-aware Euclidean features before mapping to the hyperbolic space:
\begin{spacedequation} 
\tilde{\mathbf{z}}_v = \sum_{i=1}^{b} H_v^i\!\big(\mathbf{e}_v\big),\quad
\tilde{\mathbf{z}}_t = \sum_{i=1}^{b} H_t^i\!\big(\mathbf{e}_t\big),\quad 
\end{spacedequation}
This placement and aggregation preserves old-task information through frozen components while enabling the new module to specialize for the current task $b$.
\\\textbf{Global Hyperbolic Mapping Layer}: To map all aggregated visual and textual features from  Euclidean space onto a shared hyperbolic manifold, we utilize a task-shared linear layer, $\mathrm{TP}(\cdot):\mathbb{R}^d\to\mathbb{R}^d$. It maps an aggregated Euclidean feature $\tilde{\mathbf{z}}$ (\eg, $\tilde{\mathbf{z}}_v$ or $\tilde{\mathbf{z}}_t$) to the spatial component $\mathbf{z}_s$ of the Lorentz model $\mathcal{B}^d$ via $\mathbf{z}_s = \mathrm{TP}(\tilde{\mathbf{z}})$. The resulting hyperbolic representations used for training are:\looseness -1
\begin{equation} \label{eq:agg_hyp}
    \mathbf{z}_v = \mathrm{TP}\left(\tilde{\mathbf{z}}_v\right),\quad \mathbf{z}_t =  \mathrm{TP}\!\left(\tilde{\mathbf{z}}_t\right). 
\end{equation}
For notational simplicity, $\mathbf{z}_v$ and $\mathbf{z}_t$ also denote the corresponding full Lorentz points. To balance projection fidelity and adaptation flexibility, we regularize $\mathrm{TP}$ toward the exponential map at the origin:
\begin{equation}\label{eq:tp_loss}
\mathcal{L}_{\mathrm{tp}}=\big\|\mathrm{TP}(\tilde{\mathbf{z}})-{\mathrm{expm}_{\mathbf{o}}(\tilde{\mathbf{z}})}\big\|_2^2.
\end{equation}
Overall, task-shared $\mathrm{TP}$ maps aggregated module outputs into a common hyperbolic space, producing text ($\mathbf{z}_t$) and image ($\mathbf{z}_v$) features for hierarchy-aware alignment and stable incremental training.
\\\textbf{Entailment Loss for Partial-Order Supervision}: We leverage the semantic tree $\mathcal{K}=(\mathcal{C},\mathcal{E},r)$ to encode its partial order in hyperbolic space: for any edge $(p,c)\in\mathcal{E}$, the child embedding must lie inside a parent-centered cone of the parent. We use the hyperbolic text embedding $\mathbf{z}_t^u$ for any node $u \in \mathcal{C}$ and the hyperbolic visual embedding $\mathbf{z}_v$ for an image $\mathbf{I}$.
The cone half-aperture for a hyperbolic point $\mathbf{p}$ is defined as: $\mathrm{aper}(\mathbf{p})=\sin^{-1}\!\left(\frac{2\kappa}{\sqrt{c}\,\|\mathbf{p}\|}\right)$, 
where $\kappa=0.1$ unless noted. Deeper parents (larger $\|\mathbf{p}\|$) induce tighter cones, which aligns with increasing semantic specificity. The exterior angle $\mathrm{ext}(\mathbf{p},\mathbf{q})$ measures the child $\mathbf{q}$’s deviation from the cone boundary centered at parent $\mathbf{p}$:\looseness -1
\begin{equation}
        \mathrm{ext}(\mathbf{p},\mathbf{q})=\cos^{-1}\!\left(\frac{q_t+p_t\,(c\langle \mathbf{p},\mathbf{q}\rangle_{\mathcal{B}})}{\|\mathbf{p}_s\|\sqrt{(c\langle \mathbf{p},\mathbf{q}\rangle_{\mathcal{B}})^2-1}}\right),
\end{equation}
where $\mathbf{p},\mathbf{q}\in\mathcal{B}^d$. To ensure that the learned hyperbolic space represents the semantic hierarchy, we impose the geometric constraint that each child node must lie within an aperture cone centered at its parent, formalized as $\mathrm{ext}(\mathbf{p},\mathbf{q})\le \mathrm{aper}(\mathbf{p})$. 
As visualized in Figure~\ref{figure:teaser} (Bottom-Right), this constraint encourages child nodes to lie in parent-centered descendant regions, while the separation term in Eq.~\ref{eq:text_hier_loss} further promotes a radial hierarchy between parents and children.
We penalize violations using the entailment loss:
\begin{equation}
    \mathcal{L}_{\mathrm{entail}}(\mathbf{p},\mathbf{q})=\operatorname{SP}\!\big(\mathrm{ext}(\mathbf{p},\mathbf{q})-\mathrm{aper}(\mathbf{p})\big),
\end{equation}
with the soft-plus operation $\operatorname{SP}(z)=\log(1+\mathrm{e}^z)$. In practice, we apply $\mathcal{L}_{\mathrm{entail}}$ to: (1) Text–Text pairs $(\mathbf{z}_t^p, \mathbf{z}_t^c)$ for $(p,c)\in\mathcal{E}$ and (2) Text–Image pairs $(\mathbf{z}_t^c, \mathbf{z}_v)$, effectively pulling all children into their respective parents’ cones and instantiating $\mathcal{K}$'s hierarchical relation in $\mathcal{B}^d$.
\\\textbf{Hierarchical Supervision}: 
Building on the entailment cones, we supervise both the text--text hierarchy and text--image anchoring with two complementary terms. 
For a parent-child edge $(p,c)\in\mathcal{E}$, the text--text hierarchy term is:
\begin{equation}\label{eq:text_hier_loss}
    \mathcal{L}_{\mathrm{hier}}^{\mathrm{txt}\to\mathrm{txt}}
= \mathcal{L}_{\mathrm{entail}}(\mathbf{z}_t^p,\mathbf{z}_t^c)
+ \operatorname{SP}\big(\delta - d_{\mathcal{B}}(\mathbf{z}_t^p,\mathbf{z}_t^c)\big). 
\end{equation}
In Eq.~\ref{eq:text_hier_loss}, the second term enforces a minimum parent--child separation $\delta$ to avoid embedding collapse. 
The text--image hierarchy term aligns the visual embedding $\mathbf{z}_v$ with its corresponding class text node $\mathbf{z}_t^c$:
\begin{equation}\label{eq:img_hier_loss}
    \mathcal{L}_{\mathrm{hier}}^{\mathrm{txt}\to\mathrm{img}}
=\mathcal{L}_{\mathrm{entail}}\big(\mathbf{z}_t^c,\, \mathbf{z}_v\big). 
\end{equation}
We combine the two terms in Eqs.~\ref{eq:text_hier_loss} and~\ref{eq:img_hier_loss} to form the hierarchical supervision objective:
\begin{equation} \label{eq:hier}
    \mathcal{L}_{\mathrm{hier}}
= \lambda_1\mathcal{L}_{\mathrm{hier}}^{\mathrm{txt}\to\mathrm{txt}}
+ \lambda_2\mathcal{L}_{\mathrm{hier}}^{\mathrm{txt}\to\mathrm{img}} \,.
\end{equation}
Eq.~\ref{eq:hier} stabilizes features by anchoring children within parent-centered cones, ensuring a hierarchy-respecting geometry across tasks with sufficient separation.
\\{\bf Hyperbolic Contrastive Loss}: Complementing the cone-based hierarchical supervision, we further tighten cross-modal alignment with a hyperbolic contrastive loss. Using the negative hyperbolic distance from ~Eq.~\ref{eq:lorenz_dist} as the similarity metric, we adopt the standard symmetric contrastive loss framework (similar to the multi-class N-pair loss~\cite{sohn2016improved,radford2021learning}).\\
Specifically, for the $j$-th positive image-text pair $(\mathbf{I}_j, \mathbf{t}_j)$ in the current batch, with hyperbolic embeddings $\mathbf{z}_{v,j}$ and $\mathbf{z}_{t,j}$, the match probabilities used in the contrastive loss are:
\begin{equation}
P_{j}^\text{I} = \frac{\exp(-d_\mathcal{B}(\mathbf{z}_{v,j}, \mathbf{z}_{t,j})/\tau)}{\sum_{i=1}^N \exp(-d_\mathcal{B}(\mathbf{z}_{v,j}, \mathbf{z}_{t,i})/\tau)},\quad 
P_{j}^\text{T} = \frac{\exp(-d_\mathcal{B}(\mathbf{z}_{t,j}, \mathbf{z}_{v,j})/\tau)}{\sum_{i=1}^N \exp(-d_\mathcal{B}(\mathbf{z}_{t,j}, \mathbf{z}_{v,i})/\tau)}. 
\end{equation}
The hyperbolic contrastive loss $\mathcal{L}_{\mathrm{contrast}}$ is defined as:
\begin{equation}
\label{eq:contrast-loss}
\mathcal{L}_{\mathrm{contrast}}=-\frac{1}{2N}\sum_{j=1}^{N}\left(\log P_j^I+\log P_j^T\right).
\end{equation}
Eq.~\ref{eq:contrast-loss} pulls matched pairs together and separates mismatches in $\mathcal{B}^d$, thereby stabilizing cross-task features and reinforcing cross-modal clusters.
\\\textbf{Drift Mitigation across Tasks}: 
The hierarchical supervision in Eq.~\ref{eq:hier} serves as the anchoring objective that preserves the semantic tree structure in the hyperbolic space and helps resist feature drift. To keep these hierarchical anchors stable across incremental stages, after each task we cache and freeze the hyperbolic text embeddings of all virtual nodes. In the next task, we only instantiate missing ancestors, reuse the cached virtual nodes, and compute the textual hierarchy loss over the resulting union, so that new classes are attached to existing parents without shifting previous anchors. Meanwhile, to maintain old class regions without storing exemplars, we follow~\cite{zhu2021prototype} to draw class-level features from $\mathcal{N}(\mu_k,\Sigma_k)$, where $\mu_k$ and $\Sigma_k$ are derived from the frozen encoder $g_v$. These surrogates provide old-class references in the hyperbolic space and help reduce cross-modal feature drift during training.

\subsection{Stabilizing the Shared TP}
Since $\mathrm{TP}$ is shared across tasks and modalities, its updates may shift previously learned visual--textual mappings in the hyperbolic space. To mitigate forgetting, we restrict them to directions with limited response on past pre-TP features. At the beginning of task \(b\), we compute the projection subspace from the accumulated covariance \(C^{(b-1)}\) of previous-task pre-TP features:
\begin{equation}
C^{(b-1)} = U\,\Sigma\,V^{\top},
\quad
\Sigma=\mathrm{diag}(\sigma_1\ge\cdots\ge\sigma_d).
\end{equation}
Let \(r\) be the smallest index whose cumulative energy \(\sum_{k=1}^{r}\sigma_k^{2}/\sum_{k=1}^{d}\sigma_k^{2}\) exceeds a preset threshold; this yields a data-adaptive approximation to the null space rather than an exact one. We define the orthogonal complement as $V_{\perp}=[\mathbf{v}_{r+1},\ldots,\mathbf{v}_d]$ and project the raw TP gradient $\Delta W$ onto this space:\looseness -1
\begin{equation}
\Delta W_{\mathrm{proj}}=V_{\perp}V_{\perp}^{\top}\Delta W.
\end{equation}
Since $C^{(b-1)}\propto X^{\top}X$ for past features $X$, the columns of $V_{\perp}$ span directions of near-zero variance. Thus $X^{o}V_{\perp}\approx 0$ for old-task features $X^{o}$, leading to:
\begin{equation}
X^{o}\Delta W_{\mathrm{proj}}
= X^{o}V_{\perp}V_{\perp}^{\top}\Delta W
= (X^{o}V_{\perp})(V_{\perp}^{\top}\Delta W)
\approx 0,
\end{equation}
which suggests that the projected update has limited response on old-task features and reduces the drift of previous TP mappings while still enabling learning in new directions.

After completing task \(b\), we update the covariance used for subsequent tasks:
\begin{equation}
C^{(b)}=\frac{\alpha_{b-1}\,C^{(b-1)}+N_b\,C_{\text{new}}^{(b)}}{\alpha_b},
\quad
\alpha_i=\sum_{j=1}^{i}N_j,
\end{equation}
where $N_j$ is the number of samples in task $j$, and $C_{\mathrm{new}}^{(b)}$ denotes the uncentered second-moment matrix of the current-task pre-TP features.
\\\textbf{Discussion}: 
The purpose of this step is to stabilize the shared $\mathrm{TP}$ across tasks and modalities. Projecting gradients onto the approximate null space induced by past pre-TP features reduces interference from previous tasks while keeping $\mathrm{TP}$ plastic for new ones. Specifically, under an idealized gradient descent step, the parameter update becomes $\theta_{\mathrm{TP}} \leftarrow \theta_{\mathrm{TP}} - \eta(V_{\perp}V_{\perp}^\top \Delta W)$. The energy threshold that selects $r$ tunes the stability--plasticity trade-off, where larger $r$ protects more high-variance old-task directions while smaller $r$ allows faster adaptation. We update $C^{(b)}$ from pre-TP Euclidean features after each task and refresh $V_{\perp}$ for subsequent tasks to control cost. In this way, the projection limits disturbance to previously mapped regions while allowing $\mathrm{TP}$ to learn new mappings.\looseness -1

\noindent{\bf Summary of \textsc{Hasten}}:
In \mame, we leverage hierarchical knowledge for continual learning via a semantic tree. We design hierarchy-aware perception with hyperbolic mapping and stabilize the shared mapper via projected updates. The training objective is defined by combining Eqs.~\ref{eq:hier}, \ref{eq:contrast-loss}, and~\ref{eq:tp_loss}:\looseness -1
\begin{equation} 
\label{eq:total-loss}
\mathcal{L} = 
\mathcal{L}_{\mathrm{hier}}
+\mathcal{L}_{\text{contrast}}
+\beta\mathcal{L}_{\mathrm{tp}}\,.
\end{equation}
During inference, we enhance robustness by ensembling predictions from two distinct feature spaces. The first component, the {original space probability} $P_{\text{orig}, k}(\mathbf{x})$, is computed via cosine similarity in the original Euclidean space between the image feature $\mathbf{e}_v$ and class centers $\mu_k$. The second, the {hyperbolic space probability} $P_{\text{hyp}, k}(\mathbf{x})$, leverages the learned hierarchy, using the negative hyperbolic distance $-d_{\mathcal{B}}(\mathbf{z}_v, \mathbf{z}_{t}^c)$ between the hierarchy-aware image feature $\mathbf{z}_v$ and the hyperbolic text prototype $\mathbf{z}_{t}^c$ as the logit. These two distributions are then fused to form the final class probability:
\begin{equation} \label{eq:pred}
P_k(\mathbf{x}) = P_{\text{orig}, k}(\mathbf{x}) + P_{\text{hyp}, k}(\mathbf{x})\,.
\end{equation}
The final predicted class $\hat{y}$ is determined by selecting the class with the maximum fused probability: $\hat{y} = {\operatorname{arg}\max}_k \, P_k(\mathbf{x})$.

%% file: sec/5_experiments.tex
\section{Experiments}
\label{sec:exp}
In this section, we evaluate \name on nine benchmarks against state-of-the-art methods. Furthermore, we provide comprehensive analyses, including ablation studies on hierarchical supervision, key hyperparameters, and t-SNE visualizations comparing embeddings with and without preserving hierarchical structure. Finally, we visualize image-to-root geodesic traversals to explicitly illustrate how the hyperbolic representation preserves hierarchical semantics from specific to generic concepts.

\input{table/tab1}

\subsection{Implementation Details}
\noindent {\bf Dataset}: We adopt the evaluation protocol from \cite{zhou2025external,zhou2023learning,zhou2022learning} and evaluate on nine benchmarks that exhibit domain shifts from CLIP’s pre-training data: CIFAR100 \cite{krizhevsky2009learning}, CUB200 \cite{WahCUB2002011}, ObjectNet \cite{barbu2019objectnet}, ImageNet-R \cite{hendrycks2021many}, FGVCAircraft \cite{maji2013fine}, StanfordCars \cite{krause20133d}, Food101 \cite{bossard2014food}, SUN397 \cite{xiao2010sun}, and UCF101 \cite{soomro2012ucf101}.
Following \cite{zhou2023learning}, we use sampled subsets for practical partitioning: 100 classes each from CIFAR100, FGVCAircraft, StanfordCars, Food101, and UCF101; 200 classes each from CUB200, ObjectNet, and ImageNet-R; and 300 classes from SUN397. 

{\bf Dataset split:} Following \cite{rebuffi2017icarl,zhou2025external, zhou2023learning}, we use `B-$m$ Inc-$n$' for CIL class splitting: $m$ denotes the number of classes in the first stage, and $n$ the number of classes in each subsequent stage. Consistent with \cite{rebuffi2017icarl}, we randomly shuffle the class order with seed 1993, which is kept consistent across all methods.\looseness -1

{\bf Comparison methods:} We compare our method with SOTA pre-trained model-based CIL algorithms, such as L2P~\cite{wang2022learning}, DualPrompt~\cite{wang2022dualprompt}, CODA-Prompt~\cite{smith2023coda}, and SimpleCIL~\cite{zhou2023revisiting}.
We also include SOTA CLIP-based CIL approaches, including PROOF~\cite{zhou2023learning}, RAPF~\cite{huang2024class}, and CLG-CBM~\cite{yu2025language}. The baseline of finetuning CLIP for incremental tasks is denoted as Finetune. All methods are initialized with the same CLIP model for a fair comparison.\looseness -1

{\bf Training details:}
All experiments are run on NVIDIA A100 GPUs with PyTorch~\cite{paszke2019pytorch}. Following~\cite{zhou2023learning,huang2024class}, we evaluate CLIP ViT-B/16 pre-trained on LAION-400M~\cite{ilharco2021openclip} across all methods to ensure a fair comparison. For vision-only methods that cannot use textual prompts, we initialize them with CLIP’s visual encoder. In \mame, we employ the AdamW~\cite{loshchilov2017decoupled} optimizer with a batch size of $64$ to train the model for $10$ epochs. The learning rate starts at $0.001$ and follows cosine annealing. We set $\lambda_1=0.5$ and $\lambda_2=0.1$ for the text–text and text–image hierarchical losses, $\beta=0.1$ for the TP regularizer, and $\delta=0.2$. For the fixed class order determined by seed 1993, the stage-wise hierarchical semantic trees are incrementally generated using GPT-5~\cite{OpenAI2025GPT5} before training, and the full prompts are provided in the supplementary.

{\bf Evaluation metric:} Following~\cite{rebuffi2017icarl,zhou2023learning}, we denote the model’s top-1 accuracy after the $b$-th stage by $\mathcal{A}_b$. We report $\mathcal{A}_B$ as the final-stage accuracy and $\bar{\mathcal{A}}=\frac{1}{B}\sum_{b=1}^{B}\mathcal{A}_b$ as the mean accuracy across incremental stages.

\begin{figure*}[t]
	\centering
	\begin{subfigure}{0.325\linewidth}
		\includegraphics[width=1\columnwidth]{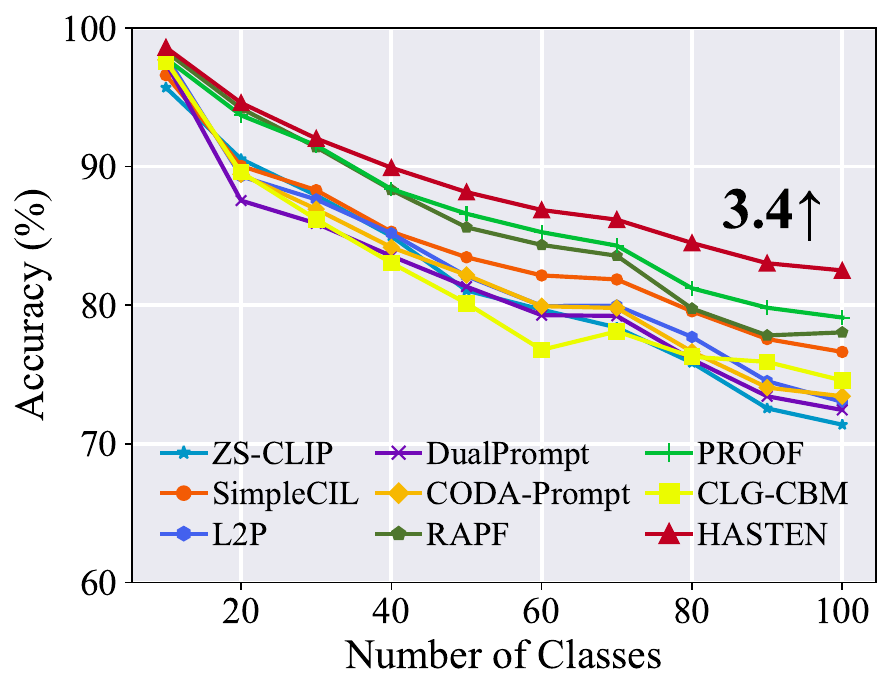}
		\caption{CIFAR100 Base0 Inc10}
		\label{fig:benchmark-cifar}
	\end{subfigure}
	\hfill
	\begin{subfigure}{0.325\linewidth}
		\includegraphics[width=1\linewidth]{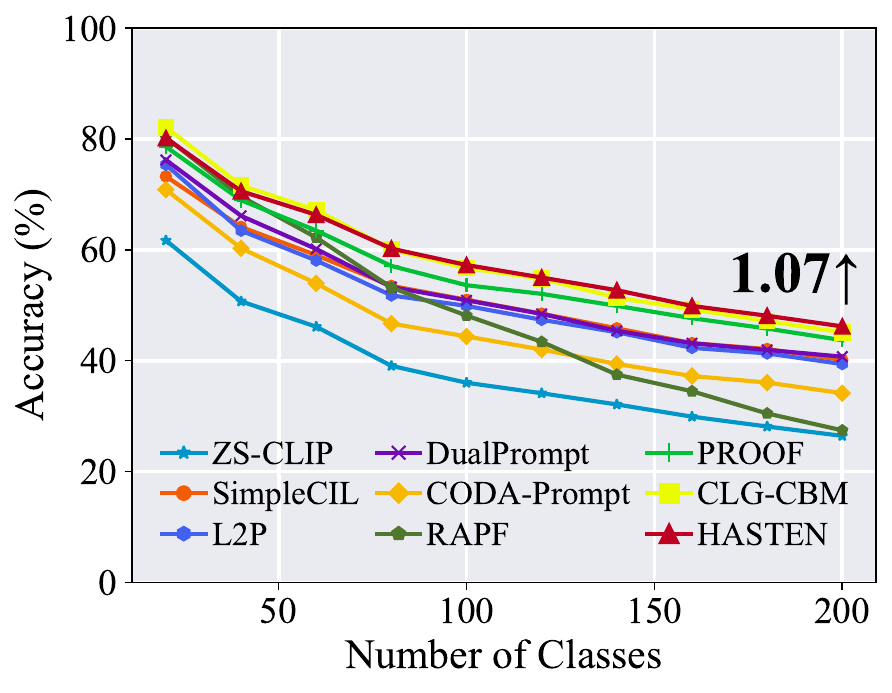}
		\caption{ObjectNet Base0 Inc20}
		\label{fig:benchmark-cars}
	\end{subfigure}
	\hfill
	\begin{subfigure}{0.325\linewidth}
		\includegraphics[width=1\columnwidth]{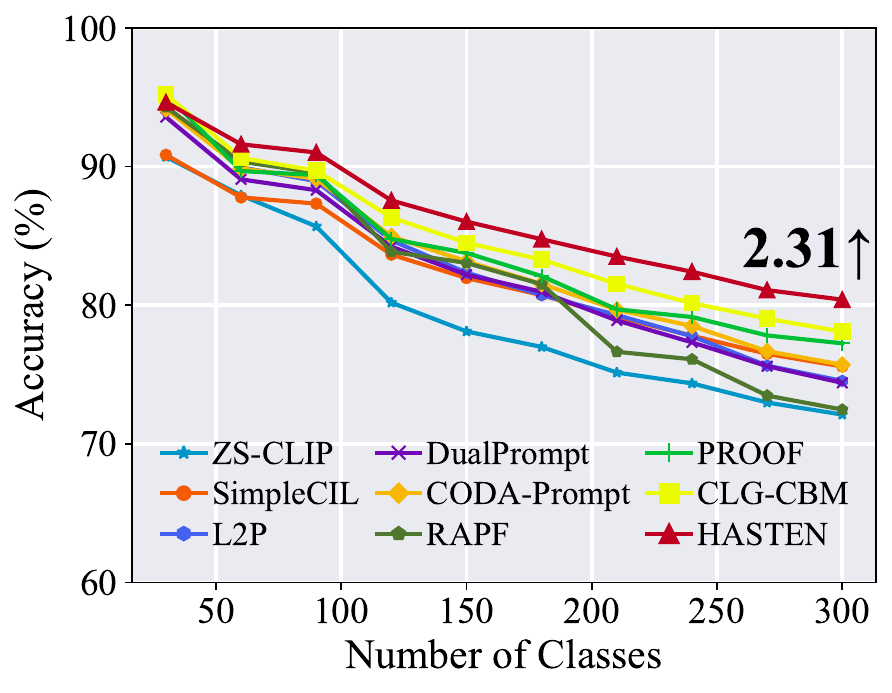}
		\caption{SUN Base0 Inc30}
		\label{fig:benchmark-ucf}
	\end{subfigure}
	\caption{\small Incremental performance of different methods. We report the performance gap after the last incremental stage of \name and the runner-up method at the end of the line. All methods use the same pre-trained CLIP weights. More results are in the supplementary.}
	\vspace{-5mm}
	\label{fig:benchmark}
\end{figure*}

\subsection{Benchmark Comparison}
We compare \name with state-of-the-art methods on benchmark datasets, with representative results reported in Table~\ref{tab:benchmark} and Figure~\ref{fig:benchmark}, and complete results on all nine datasets provided in  Appendix~\ref{sec:app_full_results}. As shown, \name consistently achieves state-of-the-art performance across the evaluated settings, demonstrating robust generalization and broad applicability. Finetuning performs the worst, indicating severe catastrophic forgetting. Visual prompt methods fall behind as they fail to leverage textual information. Compared to strong CLIP-based methods such as RAPF and PROOF, \mame's gains highlight its anti-forgetting capability while retaining the advantages of cross-modal representations.\looseness -1

\begin{figure*}[t]
    \centering
    \captionsetup[subfigure]{font=scriptsize, skip=1pt}
    \begin{subfigure}[t]{0.245\textwidth}
        \centering
        \includegraphics[width=\linewidth]{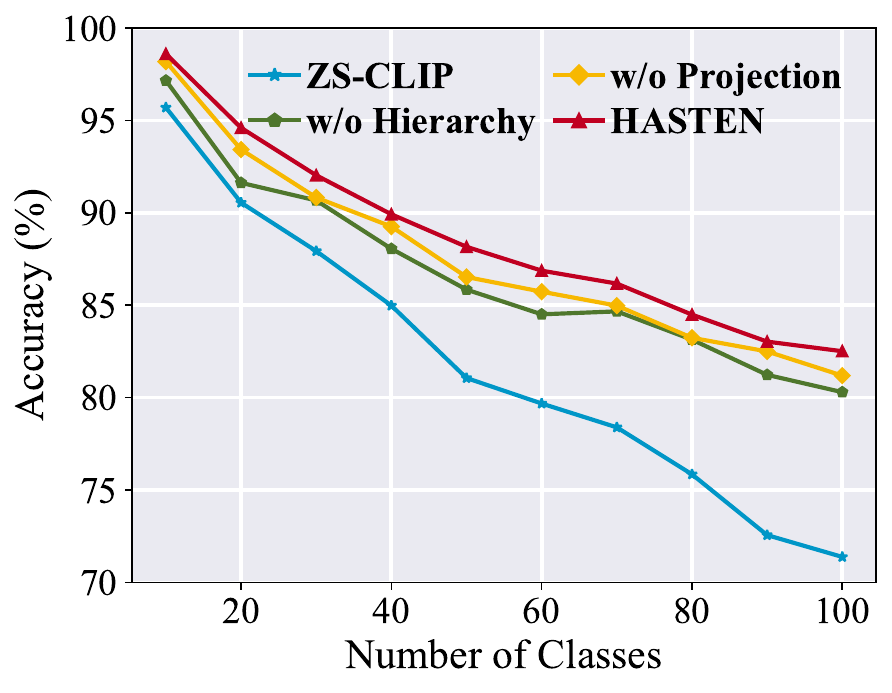}
        \caption{Ablation study}
        \label{fig:ablation}
    \end{subfigure}
    \hfill
    \begin{subfigure}[t]{0.245\textwidth}
        \centering
        \includegraphics[width=\linewidth]{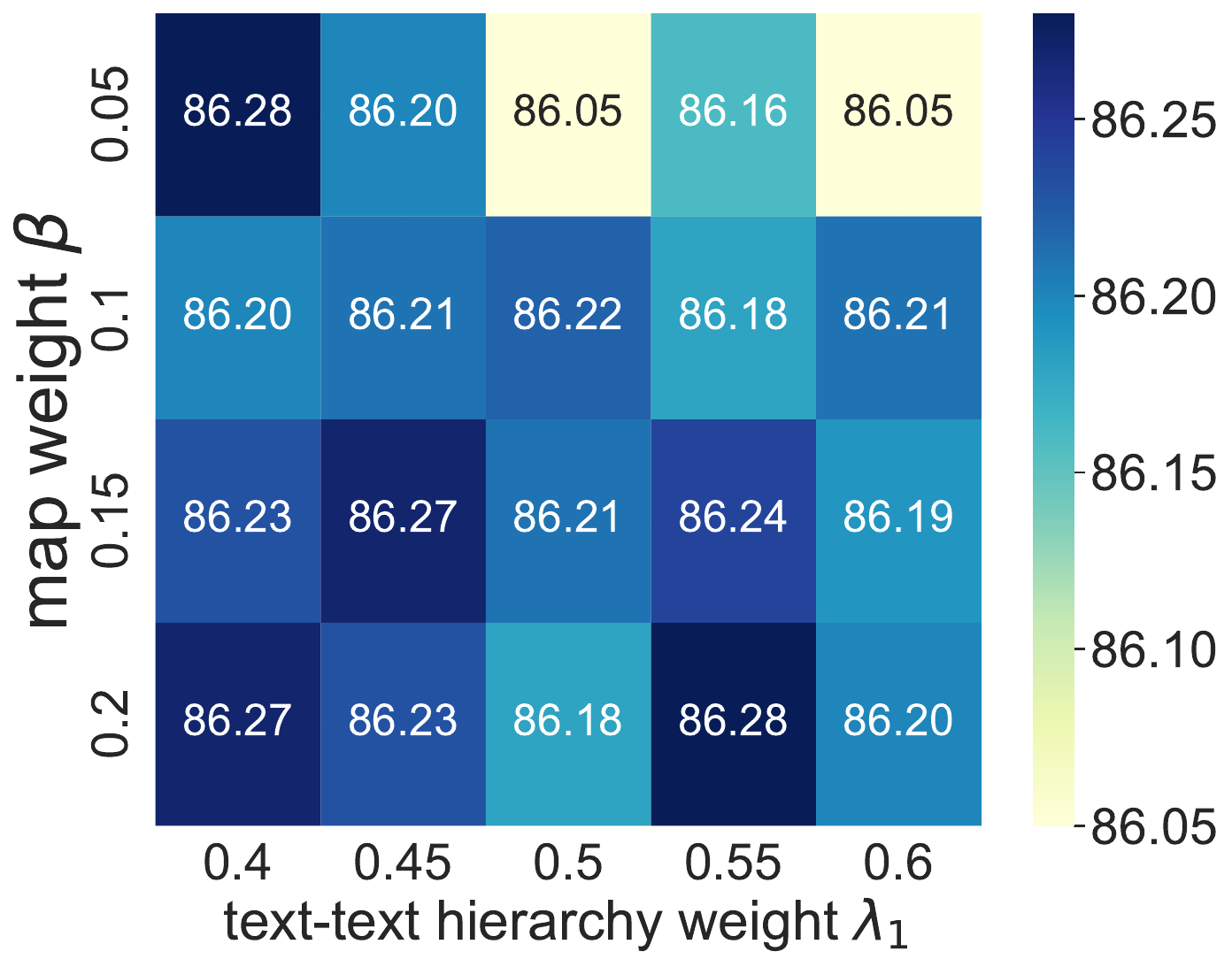}
        \caption{Parameter sensitivity}
        \label{fig:sensitivity}
    \end{subfigure}
    \hfill
    \begin{subfigure}[t]{0.245\textwidth}
        \centering
        \includegraphics[width=\linewidth]{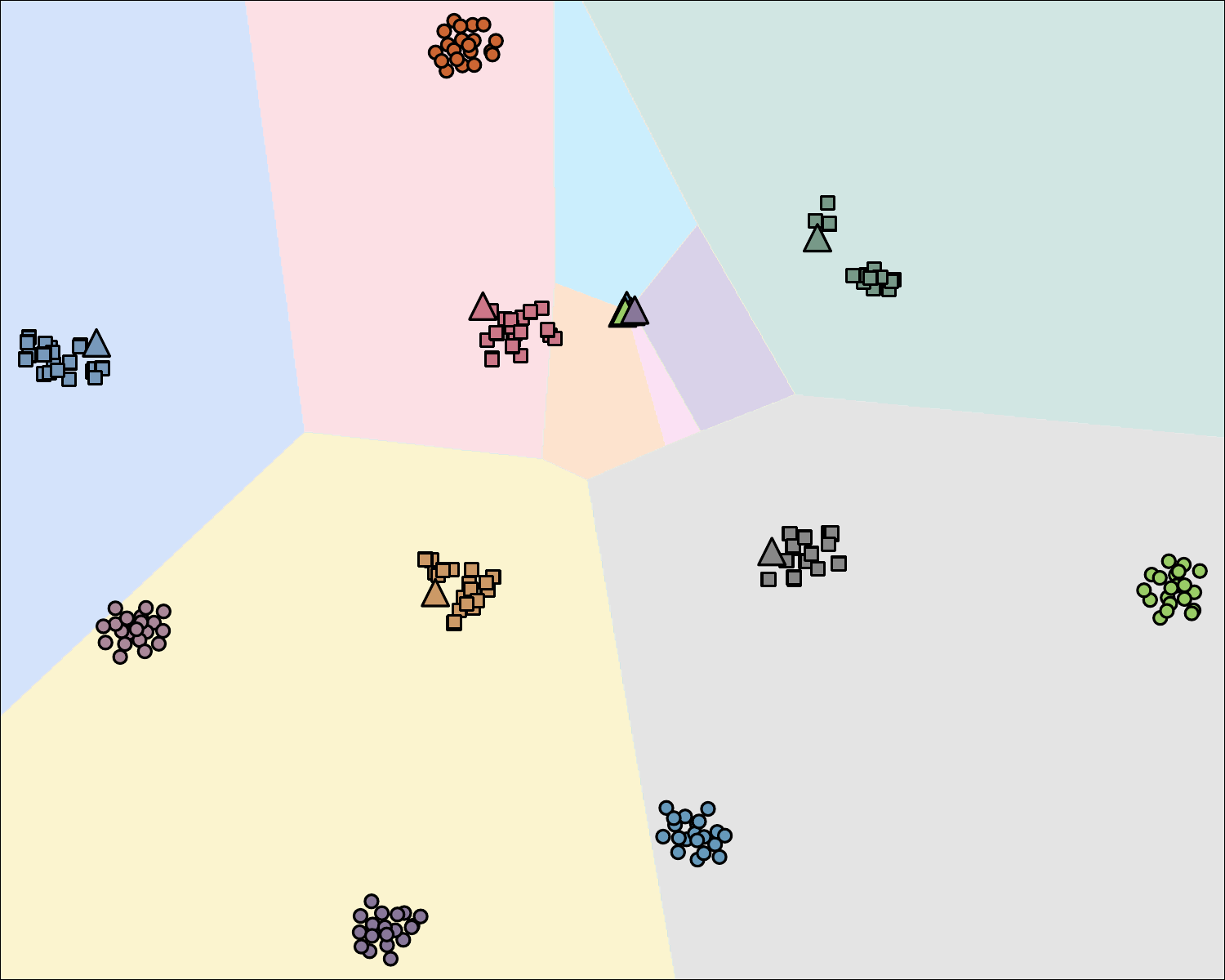}
        \caption{w/o preserving hierarchy}
        \label{fig:drift}
    \end{subfigure}
    \hfill
    \begin{subfigure}[t]{0.245\textwidth}
        \centering
        \includegraphics[width=\linewidth]{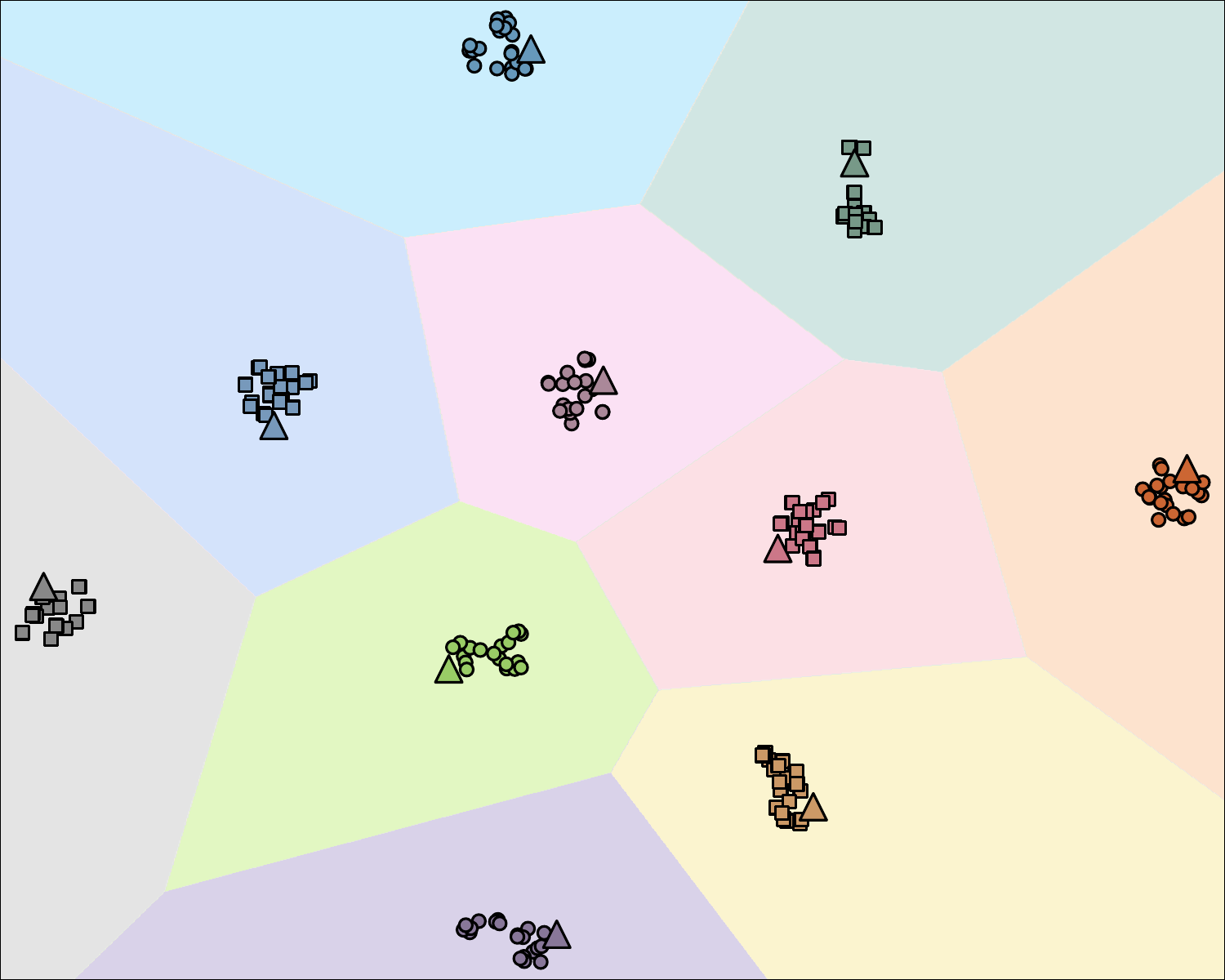}
        \caption{w/ preserving hierarchy}
        \label{fig:hier}
    \end{subfigure}
    \caption{\small
    \textbf{(a)} Ablation study and \textbf{(b)} parameter sensitivity analysis.
    \textbf{(c)(d)} t-SNE~\cite{van2008visualizing} visualizations of visual and textual features on CIFAR100 B0 Inc5.
    We show the feature distributions of old classes and new classes without (c) and with (d) hierarchy.
    }
    \label{fig:combined_analysis}
\end{figure*}

\begin{figure*}[t]
    \centering
    \begingroup
    \tiny
    \setlength{\tabcolsep}{1pt}
    \setlength{\extrarowheight}{0pt}
    \begin{minipage}[t]{0.49\textwidth}
        \centering
        \begin{minipage}[b][3.1cm][b]{0.51\linewidth}
            \centering
            \includegraphics[width=\linewidth,height=3.1cm,keepaspectratio]{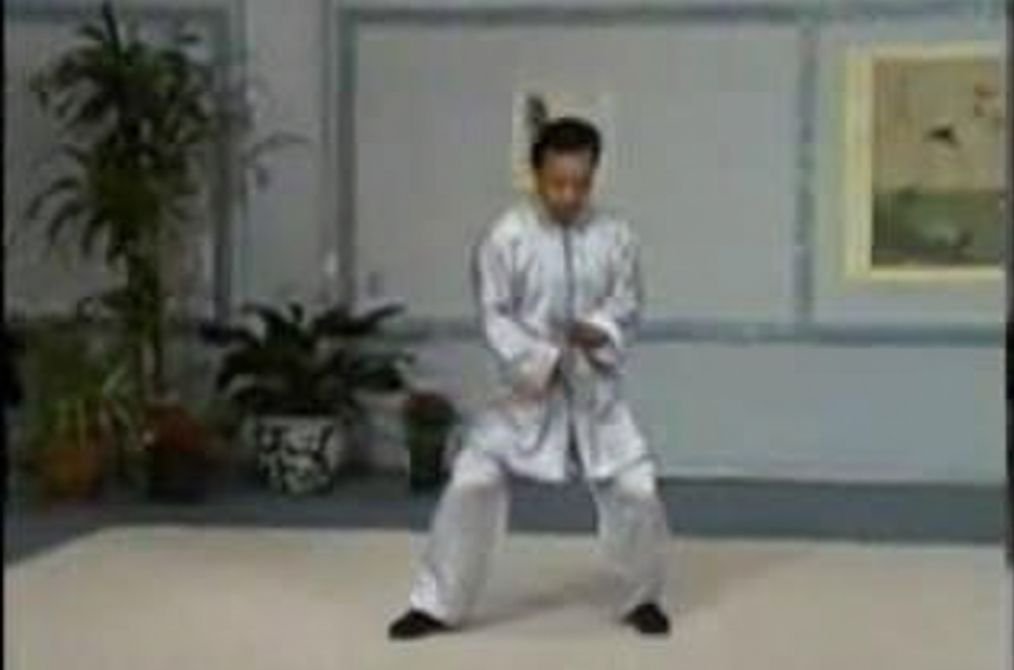}
        \end{minipage}
        \hfill
        \begin{minipage}[b][3.1cm][b]{0.48\linewidth}
            \centering
            \renewcommand{\arraystretch}{0.85}
            \begin{tabularx}{\linewidth}[b]{>{\raggedright\arraybackslash}X >{\centering\arraybackslash}m{0.22\linewidth}}
                \rowcolor{Gray!40}
                \textbf{Ours} & \textbf{CLIP} \\
                \hline
                \rowcolor{Apricot!100}
                \emph{Tai Chi} & \emph{Tai Chi} \\
                \hline
                \rowcolor{Apricot!80}
                \emph{martial\_art} & $\downarrow$ \\
                \hline
                \rowcolor{Apricot!60}
                \emph{a practitioner flowing through slow, mindful tai chi forms in a park.} & $\downarrow$ \\
                \hline
                \rowcolor{Apricot!40}
                \emph{a martial arts drill emphasizing timing, guard, and footwork.} & $\downarrow$ \\
                \hline
                \textbf{``entity''} & \texttt{\textbf{[ROOT]}} \\
                \hline
            \end{tabularx}
        \end{minipage}
    \end{minipage}
    \hfill
    \begin{minipage}[t]{0.5\textwidth}
        \centering
        \begin{minipage}[b][3.1cm][b]{0.51\linewidth}
            \centering
            \includegraphics[width=\linewidth]{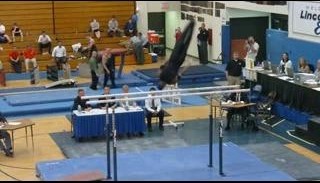}
        \end{minipage}
        \hfill
        \begin{minipage}[b][3.1cm][b]{0.48\linewidth}
            \centering
            \renewcommand{\arraystretch}{0.8}
            \begin{tabularx}{\linewidth}[b]{>{\raggedright\arraybackslash}X >{\centering\arraybackslash}m{0.22\linewidth}}
                \rowcolor{Gray!40}
                \textbf{Ours} & \textbf{CLIP} \\
                \hline
                \rowcolor{Apricot!100}
                \emph{Parallel Bars} & \emph{Parallel Bars} \\
                \hline
                \rowcolor{Apricot!80}
                \emph{a gymnast supporting the body and swinging between the parallel bars.} & $\downarrow$ \\
                \hline
                \rowcolor{Apricot!20}
                \emph{a gymnast balancing inverted and walking on hands across a mat.} & $\downarrow$ \\
                \hline
                \textbf{``entity''} & \texttt{\textbf{[ROOT]}} \\
                \hline
            \end{tabularx}
        \end{minipage}
    \end{minipage}
    \endgroup
    \caption{\small
    \textbf{Image traversals with \textsc{Hasten} and CLIP.}
    \mame's path through hyperbolic space reveals a smooth transition from specific to generic text descriptions, demonstrating a learned hierarchy.
    In contrast, CLIP's flat representation typically yields fewer descriptions along the path.
    }
    \label{fig:travel}
\end{figure*}

\subsection{Further Analysis}
\noindent\textbf{Ablation study}: We conduct ablations on CIFAR100 B0 Inc10 in Figure~\ref{fig:ablation}. \textbf{HASTEN} denotes the full model. \textbf{w/o Hierarchy} removes the hierarchical supervision. \textbf{w/o Projection} removes the TP null-space projection. \textbf{ZS-CLIP} is the zero-shot baseline and performs worse due to the downstream pre-training distribution gap. The curves show that \name stays highest across all stages; the gap over baselines widens as the number of classes grows. Removing hierarchy yields a larger and persistent drop, indicating increased drift without hierarchical supervision. Removing projection incurs a smaller early drop but hurts later stages as interference accumulates. These trends confirm that hierarchical supervision and TP stabilization provide complementary benefits for strong incremental performance.\looseness -1

\textbf{Parameter robustness}: 
We evaluate the sensitivity to two hyperparameters, including the text--text hierarchical weight $\lambda_1$ in Eq.~\ref{eq:hier} and the hyperbolic mapping weight $\beta$ in~Eq.~\ref{eq:total-loss}. 
On the CUB B0 Inc20 setting, we sweep $\lambda_1$ among $\{0.40,0.45,0.50,0.55,0.60\}$ and $\beta$ among $\{0.05,0.10,0.15,0.20\}$. 
Figure~\ref{fig:sensitivity} reports the average validation accuracy, showing that \name maintains stable performance across a broad range of hyperparameter choices. 
Additional sensitivity results and details are provided in Appendix~\ref{sec:app_sensitivity}.

\textbf{Visualizations}: 
We use t-SNE~\cite{van2008visualizing} to visualize cross-modal features on CIFAR100 B0 Inc5 in Figure~\ref{fig:drift} and Figure~\ref{fig:hier}, comparing models with and without preserving hierarchical structure across tasks. First-task visual features are dots ($\bigcirc$) and second-task ones are squares ($\square$). Text embeddings are triangles ($\triangle$), and shaded regions denote decision areas induced by the text points. As shown in Figure~\ref{fig:drift}, without preserving hierarchical structure, visual clusters drift away from their text counterparts and even enter other classes' regions. As shown in Figure~\ref{fig:hier}, preserving hierarchical structure suppresses drift and keeps image--text features co-located for old and new tasks. These results indicate that \name uses hierarchical supervision to preserve the learned structure and maintain cross-modal alignment during incremental training.\looseness -1

\textbf{Image traversals}: 
Following~\cite{desai2023hyperbolic}, we visualize hierarchical consistency on UCF101 by traversing from each image
embedding to the root (``entity'') along the geodesic in our hyperbolic space, and along the image-to-\texttt{[ROOT]}
path for CLIP. We sample 100 equally spaced points on each path; ours retrieves the nearest ancestor node, while CLIP retrieves the highest-similarity text from a text set that includes the root. 
As shown in Figure~\ref{fig:travel}, our method yields descriptions that become increasingly generic as the path approaches the root, indicating that the hyperbolic representation captures the semantic tree. Conversely, CLIP's flat feature space typically produces a single description. More details and examples are in  Appendix~\ref{sec:app_hierarchy_details}.\looseness -1

%% file: table/tab1.tex
\begin{table*}[!t]
	\caption{Average and last performance comparison of different methods.  
		The best performance is shown in bold.  
		{\bf All methods are initialized with the same pre-trained CLIP for a fair comparison.} }\label{tab:benchmark}
	\centering
    \setlength{\tabcolsep}{10pt}
    
	\resizebox{1\textwidth}{!}{%
		\begin{NiceTabular}{@{} l *{12}{c} }
			\toprule
			\multicolumn{1}{c}{\multirow{3}{*}{Method}}
			&
			\multicolumn{4}{c}{Aircraft }   & 
			\multicolumn{4}{c}{CIFAR100 }	&	
			\multicolumn{4}{c}{Cars }   
			\\ 
			& 
			\multicolumn{2}{c}{B0 Inc10}   & 
			\multicolumn{2}{c}{B50 Inc10}	&		
			\multicolumn{2}{c}{B0 Inc10}   & 
			\multicolumn{2}{c}{B50 Inc10}	& 
			\multicolumn{2}{c}{B0 Inc10}   & 
			\multicolumn{2}{c}{B50 Inc10}	
			\\  
			& 
			{$\bar{\mathcal{A}}$} & ${\mathcal{A}_B}$  
			& {$\bar{\mathcal{A}}$} & ${\mathcal{A}_B}$
			& {$\bar{\mathcal{A}}$} & ${\mathcal{A}_B}$ 
			& {$\bar{\mathcal{A}}$} & ${\mathcal{A}_B}$  
			& {$\bar{\mathcal{A}}$} & ${\mathcal{A}_B}$
			& {$\bar{\mathcal{A}}$} & ${\mathcal{A}_B}$ 
			\\
			\midrule
			Finetune  &  3.16 & 0.96 & 1.72 & 1.05 & 7.84 & 4.44& 5.30 & 2.46& 3.14 & 1.10 & 1.54 & 1.13\\
			SimpleCIL~\cite{zhou2023revisiting} &59.24 & 48.09 & 53.05 & 48.09 & 84.15 & 76.63& 80.20 & 76.63& 92.04 & 86.85 & 88.96 & 86.85\\
			ZS-CLIP~\cite{radford2021learning} &26.66 & 17.22 & 21.70 & 17.22& 81.81 & 71.38& 76.49 & 71.38& 82.60 & 76.37& 78.32 & 76.37\\
			L2P~\cite{wang2022learning}  &47.19 & 28.29 &44.07&32.13& 82.74 & 73.03& 81.14 & 73.61& 76.63 & 61.82& 76.37 & 65.64 \\
			DualPrompt~\cite{wang2022dualprompt}  & 44.30& 25.83 &46.07&33.57 & 81.63 & 72.44& 80.12 & 72.57& 76.26 & 62.94& 76.88 & 67.55 \\
			CODA-Prompt~\cite{smith2023coda}  & 45.98 & 27.69 & 45.14 & 32.28& 82.43 & 73.43& 78.69 & 71.58& 80.21 & 66.47& 75.06 & 64.19 \\
            PROOF~\cite{zhou2023learning} &  63.81& 	56.14&	59.47&	57.10&	86.77&	79.11&	83.32&	79.73&	90.74&	86.51&	88.00&	85.58\\
			RAPF~\cite{huang2024class}   &  50.38  & 23.61 &  40.47 &  25.44 & 86.14 & 78.04 & 82.17 &  77.93  & 82.89 & 62.85 &  75.87 & 63.19\\
            CLG-CBM~\cite{yu2025language} & 66.05 & 55.93 & 59.26 & 55.39 & 81.82 & 74.57 & 68.61 & 65.55 & 78.94 & 74.26 & 80.11 & 73.62 \\
            \rowcolor{LightCyan}\name   & \bf67.23&\bf56.32&\bf61.47&\bf57.19&\bf88.64&\bf82.51&\bf85.45&\bf82.38&\bf93.06&\bf88.93&\bf90.23&\bf88.46\\
		\end{NiceTabular}
	}	
	
	\resizebox{1\textwidth}{!}{%
		\begin{NiceTabular}{@{} l *{12}{c} }
			\toprule
			\multicolumn{1}{c}{\multirow{3}{*}{Method}}
			& 
			\multicolumn{4}{c}{ImageNet-R }   & 
			\multicolumn{4}{c}{CUB }	&	
            \multicolumn{4}{c}{UCF }   
			\\ 
			& 
			\multicolumn{2}{c}{B0 Inc20}   & 
			\multicolumn{2}{c}{B100 Inc20}	&	
            \multicolumn{2}{c}{B0 Inc20}   & 
			\multicolumn{2}{c}{B100 Inc20}	& 
			\multicolumn{2}{c}{B0 Inc10}   & 
			\multicolumn{2}{c}{B50 Inc10}	
			\\  
			& 
			{$\bar{\mathcal{A}}$} & ${\mathcal{A}_B}$  
			& {$\bar{\mathcal{A}}$} & ${\mathcal{A}_B}$
			& {$\bar{\mathcal{A}}$} & ${\mathcal{A}_B}$ 
			& {$\bar{\mathcal{A}}$} & ${\mathcal{A}_B}$  
			& {$\bar{\mathcal{A}}$} & ${\mathcal{A}_B}$
			& {$\bar{\mathcal{A}}$} & ${\mathcal{A}_B}$ 
			\\
			\midrule
			Finetune  & 1.37 & 0.43& 1.01 & 0.88& 2.06 & 0.64& 0.56 & 0.47& 4.51 & 1.59& 1.21 & 0.80\\
			SimpleCIL~\cite{zhou2023revisiting} & 81.06 & 74.48& 76.84 & 74.48& 83.81 & 77.52& 79.75 & 77.52& 90.44 & 85.68& 88.12 & 85.68\\
			ZS-CLIP~\cite{radford2021learning} &83.37 & 77.17& 79.57 & 77.17 & 74.38 & 63.06& 67.96 & 63.06& 75.50 & 67.64& 71.44 & 67.64\\
			L2P~\cite{wang2022learning}  &75.97 & 66.52 & 72.82 & 66.77&   70.87&57.93 & 75.64 &66.12 & 86.34 & 76.43& 83.95 & 76.62 \\
			DualPrompt~\cite{wang2022dualprompt}  &76.21 & 66.65 & 73.22 & 67.58&69.89 &57.46 & 74.40 &64.84 & 85.21 & 75.82& 84.31 & 76.35 \\
			CODA-Prompt~\cite{smith2023coda}  & 77.69 & 68.95 & 73.71 & 68.05& 73.12&62.98 &73.95&62.21 & 87.76 & 80.14&83.04 & 75.03 \\
            PROOF~\cite{zhou2023learning} &  83.84&	78.40&	81.20&	78.92&	82.31&	76.64&	79.20&	76.37&	94.58&	91.10&	93.58&	90.91\\
			RAPF~\cite{huang2024class}  & 81.26  & 70.48 & 76.10 & 70.23 &  79.09 & 62.77& 72.82 & 62.93 & 92.28 & 80.33&90.31 & 81.55\\
            CLG-CBM~\cite{yu2025language} & 79.19 & 72.03 & 75.20 & 70.50 & 78.11 & 72.14 & 78.17 & 73.03 & 95.04 & 91.36 & 94.17 & 91.85 \\
			\rowcolor{LightCyan}\name    & \bf85.73&\bf80.70&\bf82.58&\bf80.53&\bf86.22&\bf80.96&\bf82.86&\bf80.20&\bf96.49&\bf92.69&\bf95.55&\bf92.91\\
		\end{NiceTabular}
	    }
	\resizebox{\textwidth}{!}{%
		\begin{NiceTabular}{@{} l *{12}{c}}
            \toprule
			\multicolumn{1}{c}{\multirow{3}{*}{Method}}
			&
			\multicolumn{4}{c}{SUN }   & 
			\multicolumn{4}{c}{Food }	&	\multicolumn{4}{c}{ObjectNet }   
			\\ 
			& 
			\multicolumn{2}{c}{B0 Inc30}   & 
			\multicolumn{2}{c}{B150 Inc30}	&		\multicolumn{2}{c}{B0 Inc10}   & 
			\multicolumn{2}{c}{B50 Inc10}	& 
			\multicolumn{2}{c}{B0 Inc20}   & 
			\multicolumn{2}{c}{B100 Inc20}	 
			\\  
			& 
			{$\bar{\mathcal{A}}$} & ${\mathcal{A}_B}$  
			& {$\bar{\mathcal{A}}$} & ${\mathcal{A}_B}$
			& {$\bar{\mathcal{A}}$} & ${\mathcal{A}_B}$ 
			&  {$\bar{\mathcal{A}}$} & ${\mathcal{A}_B}$  
			& {$\bar{\mathcal{A}}$} & ${\mathcal{A}_B}$
			& {$\bar{\mathcal{A}}$} & ${\mathcal{A}_B}$ 
			\\
			\midrule
			Finetune  &4.51 & 1.59& 0.78 & 0.72& 3.49 & 1.71& 2.14 & 1.52& 1.34 & 0.47 & 0.69 & 0.54\\
			SimpleCIL~\cite{zhou2023revisiting} & 82.13 & 75.58& 78.62 & 75.58& 87.89 & 81.65& 84.73 & 81.65& 52.06 & 40.13& 45.11 & 40.13\\
			ZS-CLIP~\cite{radford2021learning} &79.42 & 72.11& 74.95 & 72.11& 87.86 & 81.92& 84.75 & 81.92& 38.43 & 26.43 & 31.12 & 26.43\\	
			L2P~\cite{wang2022learning} &82.82 & 74.54 & 79.57 & 73.10 & 85.66 & 77.33& 80.42 & 73.13& 51.40 & 39.39& 48.91 & 42.83 \\
			DualPrompt~\cite{wang2022dualprompt} & 82.46 & 74.40 & 79.37 & 73.02& 84.92 &77.29& 80.00 & 72.75& 52.62 & 40.72& 49.08 & 42.92 \\
			CODA-Prompt~\cite{smith2023coda} &   83.34 & 75.71 & 80.38 & 74.17& 86.18 & 78.78& 80.98 & 74.13& 46.49 & 34.13& 40.57 & 34.13 \\
            PROOF~\cite{zhou2023learning} &  83.89&	77.25&	80.15&	76.54&	90.04&	84.73&	87.52&	84.74&	56.07&	43.69&	48.90&	43.62\\
			RAPF~\cite{huang2024class}  & 82.13 & 72.47 & 78.04 & 73.10 & 88.57 & 81.15&\ 85.53 & 81.17&  48.67 & 27.43 & 39.28 &  28.73 \\
            CLG-CBM~\cite{yu2025language} & 84.85 & 78.09 & 81.58 & 77.79 & 89.12 & 83.05 & 86.76 & 83.85 & 58.52 & 45.11 & 49.80 & 43.56 \\
			\rowcolor{LightCyan}\name  & \bf86.30&\bf80.40&\bf82.99&\bf80.25&\bf90.67&\bf85.70&\bf88.18&\bf85.75&\bf58.63&\bf46.18&\bf52.03&\bf45.88 \\
			\bottomrule
		\end{NiceTabular}
        }
\end{table*}

%% file: sec/6_conclusion.tex
\section{Conclusion} 
\label{sec:conclusion}
This paper presents \mame, a hierarchy-aware framework for CLIP-based CIL in hyperbolic space. \name leverages external hierarchical knowledge as structured supervision to organize visual and textual features into a hierarchically structured representation, instead of isolated class labels. This representation helps maintain semantic relations and suppress feature drift during incremental learning. To further reduce forgetting, null-space projection stabilizes the shared hyperbolic mapper by reducing interference with prior mappings. Extensive experiments show consistent gains across diverse benchmarks, improved resistance to feature drift, and reduced catastrophic forgetting.
\\\textbf{Limitations}: 
This paper uses a GPT-generated semantic tree for hierarchical supervision, which may be imperfect in specialized domains with ambiguous or fine-grained class relations. Future work will explore curated taxonomies, data-driven hierarchy induction, and light human verification for more reliable domain-specific deployment.

\section*{Acknowledgments}

This work was supported in part by the Basic Research Program of Jiangsu (BK20251251), NSFC (62506160),  JSTJ-2025-147, Fundamental and Interdisciplinary Disciplines Breakthrough Plan of the Ministry of Education of China (No. JYB2025XDXM118), the 111 Center (No. B26023), and the Collaborative Innovation Center of Novel Software Technology and Industrialization. The authors declare no competing interests.

%% file: sec/X_suppl.tex
\clearpage
\appendix
\section*{\centering Appendix}
\renewcommand{\thesection}{\Alph{section}}
\setcounter{section}{0}

In this appendix, we provide supplementary materials for \mame, including additional method details, semantic-tree construction protocols, comprehensive experimental results, robustness analyses, and efficiency studies.

\noindent\textbf{Section}~\ref{sec:more} provides more details of \mame, including an overview of the training algorithm and a geometric justification for using Lorentz hyperbolic space to model hierarchical structures.

\noindent\textbf{Section}~\ref{sec:app_prompt} presents the prompts and protocols used to dynamically construct hierarchical semantic trees with GPT-5.

\noindent\textbf{Section}~\ref{sec:app_more_results} provides supplementary experimental results and analyses, including component analysis, forgetting analysis, generalization studies, robustness evaluations, and descriptions of the compared methods.

\noindent\textbf{Section}~\ref{sec:app_sensitivity} supplements the parameter robustness study with sensitivity analyses of additional hyperparameters, including $\lambda_2$ for text-to-image hierarchy, $\delta$ for parent--child separation, the curvature $c$, and the cone half-aperture parameter $\kappa$.

\noindent\textbf{Section}~\ref{sec:app_params} analyzes computational efficiency, including trainable parameters, running time, and additional overhead.

\noindent\textbf{Section}~\ref{sec:app_full_results} reports the full incremental learning curves under different benchmark settings.

\noindent\textbf{Section}~\ref{sec:app_hierarchy_details} provides more image traversal examples and details the construction of the text set used for hierarchical consistency evaluation.

\noindent\textbf{Section}~\ref{sec:app_hierarchy_verification} quantitatively verifies the learned hierarchical structure across multiple datasets, showing that the hyperbolic representations preserve radial expansion and semantic topology after the final incremental stage.

\noindent\textbf{Section}~\ref{sec:app_broader_impacts} discusses the broader impacts and practical considerations.

\section{More Details of \textsc{Hasten}}
\label{sec:more}
\subsection{Overview of Training Algorithm}
\label{sec:alg}
The training process of \name is summarized in Algorithm~\ref{alg:hasten}. At the beginning of the $b$-th incremental task, \name first constructs or expands the hierarchical semantic tree using only the currently observed classes, without accessing future classes (Lines~\ref{line:tree_start}--\ref{line:tree_end}). Then, all previously learned hierarchy-aware modules and cached virtual-node embeddings are frozen, and the current hierarchy-aware modules are initialized for the new task (Lines~\ref{line:freeze_old}--\ref{line:init_new}). During training, frozen CLIP encoders extract visual and textual features, which are processed by the accumulated hierarchy-aware modules and mapped into the Lorentz hyperbolic space through the shared TP layer (Lines~\ref{line:clip_feat}--\ref{line:hyper_feat}). The model is optimized with the hierarchical loss, hyperbolic contrastive loss, and TP regularization (Lines~\ref{line:loss_start}--\ref{line:loss_end}). To stabilize the shared TP, its gradient is projected onto the null space induced by prior-task features before the parameter update (Lines~\ref{line:proj_start}--\ref{line:proj_end}). After finishing the task, \name updates the class statistics, covariance matrix, projection subspace, and cached virtual-node embeddings for subsequent tasks (Lines~\ref{line:update_start}--\ref{line:update_end}).

\begin{algorithm}[t]
\caption{Training \name for CLIP-based CIL}
\label{alg:hasten}
\raggedright
\textbf{Input:} Incremental task dataset $\mathcal{D}^b$; class set $\mathcal{Y}_b$; frozen CLIP encoders $g_v,g_t$; previous semantic tree $\mathcal{K}^{(b-1)}$; previous modules $\{H_v^i,H_t^i\}_{i=1}^{b-1}$; shared hyperbolic mapper $\mathrm{TP}$; previous covariance $C^{(b-1)}$; cached virtual-node embeddings.\\
\textbf{Output:} Updated model, semantic tree, covariance, projection subspace, and cached virtual-node embeddings.
\begin{algorithmic}[1]
    \IF{$b=1$}
        \STATE Construct the base semantic tree $\mathcal{K}^{(b)}$ from $\mathcal{Y}_b$ using GPT-5; \label{line:tree_start}
    \ELSE
        \STATE Expand $\mathcal{K}^{(b-1)}$ with new classes $\mathcal{Y}_b$ without modifying existing edges;
    \ENDIF
    \STATE Obtain the current semantic tree $\mathcal{K}^{(b)}=(\mathcal{C},\mathcal{E},r)$; \label{line:tree_end}

    \STATE Freeze previous hierarchy-aware modules $\{H_v^i,H_t^i\}_{i=1}^{b-1}$ and cached virtual-node embeddings; \label{line:freeze_old}
    \STATE Initialize current hierarchy-aware modules $H_v^b$ and $H_t^b$; \label{line:init_new}

    \IF{$b>1$}
        \STATE Compute the projection basis $V_{\perp}^{(b-1)}$ from $C^{(b-1)}$ via SVD;
    \ELSE
        \STATE Set $V_{\perp}^{(b-1)}$ to the identity projection;
    \ENDIF

    \FOR{each training iteration on $\mathcal{D}^b$}
        \STATE Sample a mini-batch $(\mathbf{I},y)$ from $\mathcal{D}^b$;
        \STATE Extract frozen CLIP features $\mathbf{e}_v=g_v(\mathbf{I})$ and $\mathbf{e}_t=g_t(\mathbf{t}_y)$; \label{line:clip_feat}

        \STATE Aggregate hierarchy-aware features:
        $\tilde{\mathbf{z}}_v=\sum_{i=1}^{b}H_v^i(\mathbf{e}_v)$,
        $\tilde{\mathbf{z}}_t=\sum_{i=1}^{b}H_t^i(\mathbf{e}_t)$;

        \STATE Map features into hyperbolic space:
        $\mathbf{z}_v=\mathrm{TP}(\tilde{\mathbf{z}}_v)$,
        $\mathbf{z}_t=\mathrm{TP}(\tilde{\mathbf{z}}_t)$; \label{line:hyper_feat}

        \STATE Sample old-class surrogates from $\mathcal{N}(\mu_k,\Sigma_k)$ for previous classes when needed;

        \STATE Compute the text--text and text--image hierarchical loss $\mathcal{L}_{\mathrm{hier}}$ using $\mathcal{K}^{(b)}$ and cached virtual nodes; \label{line:loss_start}
        \STATE Compute the hyperbolic contrastive loss $\mathcal{L}_{\mathrm{contrast}}$;
        \STATE Compute the TP regularization loss $\mathcal{L}_{\mathrm{tp}}$;
        \STATE Compute the total loss
        $\mathcal{L}=\mathcal{L}_{\mathrm{hier}}+\mathcal{L}_{\mathrm{contrast}}+\beta\mathcal{L}_{\mathrm{tp}}$; \label{line:loss_end}

        \STATE Back-propagate $\mathcal{L}$ to obtain gradients for $H_v^b$, $H_t^b$, and $\mathrm{TP}$;
        \STATE Project the TP gradient:
        $\Delta w_{\mathrm{proj}}=V_{\perp}^{(b-1)}(V_{\perp}^{(b-1)})^{\top}\Delta w$; \label{line:proj_start}
        \STATE Update $H_v^b$, $H_t^b$, and update $\mathrm{TP}$ with $\Delta w_{\mathrm{proj}}$; \label{line:proj_end}
    \ENDFOR

    \STATE Estimate class statistics $\{\mu_k,\Sigma_k\}$ for classes in $\mathcal{Y}_b$ using frozen visual features; \label{line:update_start}
    \STATE Compute $C_{\mathrm{new}}^{(b)}$ from pre-TP features and update $C^{(b)}$;
    \STATE Recompute the SVD of $C^{(b)}$ and update the projection subspace for future tasks;
    \STATE Cache and freeze hyperbolic text embeddings of virtual nodes in $\mathcal{K}^{(b)}$; \label{line:update_end}
    \RETURN Updated \mame.
\end{algorithmic}
\end{algorithm}

\subsection{Geometric Justification of Hyperbolic Hierarchy Modeling}
\label{app:hyperbolic_hierarchy}

We provide a geometric justification for why the Lorentz hyperbolic space used in \name is suitable for representing hierarchical structures.

\begin{proposition}
The Lorentz hyperbolic space $\mathcal{B}^d$ with curvature parameter $c>0$ provides exponentially increasing surface capacity with respect to the geodesic radius. Specifically, the surface area of a geodesic sphere with radius $\rho$ grows as
\begin{equation}
    S_{\mathcal{B}^d}(\rho)
    =
    \Theta\!\left(e^{(d-1)\sqrt{c}\rho}\right),
    \quad \rho\rightarrow\infty .
\end{equation}
Thus, it naturally matches the exponential branching pattern of tree-structured data and supports a depth-radius representation of hierarchical concepts.
\end{proposition}

\noindent\textbf{Proof:}
Following the notation in the main paper, we use the Lorentz model $\mathcal{B}^d$ with curvature parameter $c>0$. For a point $\mathbf{p}=[\mathbf{p}_s,p_t]\in\mathcal{B}^d$, the Lorentz constraint is
\begin{equation}
    \langle \mathbf{p},\mathbf{p}\rangle_{\mathcal{B}}
    =
    \|\mathbf{p}_s\|^2-p_t^2
    =
    -\frac{1}{c},
    \quad p_t>0 .
\end{equation}
The geodesic distance used in the main paper is
\begin{equation}
    d_{\mathcal{B}}(\mathbf{p},\mathbf{q})
    =
    \frac{1}{\sqrt{c}}
    \cosh^{-1}\!\left(
    -c\langle \mathbf{p},\mathbf{q}\rangle_{\mathcal{B}}
    \right).
\end{equation}
Let $\rho=d_{\mathcal{B}}(\mathbf{o},\mathbf{p})$ denote the geodesic radius from the origin. For general $c$, the origin is $\mathbf{o}=[\mathbf{0},1/\sqrt{c}]$, which reduces to $[\mathbf{0},1]$ when $c=1$. Any point at radius $\rho$ can be parameterized by a direction $\mathbf{u}\in\mathbb{S}^{d-1}$ as
\begin{equation}
    \Phi(\rho,\mathbf{u})
    =
    \left[
    \frac{\sinh(\sqrt{c}\rho)}{\sqrt{c}}\mathbf{u},
    \frac{\cosh(\sqrt{c}\rho)}{\sqrt{c}}
    \right].
\end{equation}
This parameterization lies on $\mathcal{B}^d$ because
\begin{equation}
    \left\|
    \frac{\sinh(\sqrt{c}\rho)}{\sqrt{c}}\mathbf{u}
    \right\|^2
    -
    \left(
    \frac{\cosh(\sqrt{c}\rho)}{\sqrt{c}}
    \right)^2
    =
    \frac{\sinh^2(\sqrt{c}\rho)-\cosh^2(\sqrt{c}\rho)}{c}
    =
    -\frac{1}{c}.
\end{equation}
Moreover, $\Phi(\rho,\mathbf{u})$ is indeed at geodesic distance $\rho$ from the origin, since
\begin{equation}
    -c\langle \mathbf{o},\Phi(\rho,\mathbf{u})\rangle_{\mathcal{B}}
    =
    \cosh(\sqrt{c}\rho),
\end{equation}
and therefore
\begin{equation}
    d_{\mathcal{B}}(\mathbf{o},\Phi(\rho,\mathbf{u}))
    =
    \frac{1}{\sqrt{c}}\cosh^{-1}\!\big(\cosh(\sqrt{c}\rho)\big)
    =
    \rho .
\end{equation}

We next derive the volume-growth property of $\mathcal{B}^d$. Differentiating the above parameterization gives
\begin{equation}
    d\mathbf{p}_s
    =
    \cosh(\sqrt{c}\rho)\mathbf{u}\,d\rho
    +
    \frac{\sinh(\sqrt{c}\rho)}{\sqrt{c}}d\mathbf{u},
    \quad
    dp_t
    =
    \sinh(\sqrt{c}\rho)d\rho .
\end{equation}
Since $\|\mathbf{u}\|=1$, we have $\mathbf{u}^{\top}d\mathbf{u}=0$. Thus, the induced metric becomes
\begin{align}
    ds^2
    &=
    \langle d\mathbf{p},d\mathbf{p}\rangle_{\mathcal{B}} \notag\\
    &=
    \|d\mathbf{p}_s\|^2-dp_t^2 \notag\\
    &=
    \bigl(\cosh^2(\sqrt{c}\rho)-\sinh^2(\sqrt{c}\rho)\bigr)d\rho^2
    +
    \left(
    \frac{\sinh(\sqrt{c}\rho)}{\sqrt{c}}
    \right)^2 d\Omega^2 \notag\\
    &=
    d\rho^2
    +
    \left(
    \frac{\sinh(\sqrt{c}\rho)}{\sqrt{c}}
    \right)^2 d\Omega^2 ,
\end{align}
where $d\Omega^2$ is the metric on the unit sphere $\mathbb{S}^{d-1}$. Hence, the surface area of a geodesic sphere with radius $\rho$ is
\begin{equation}
    S_{\mathcal{B}^d}(\rho)
    =
    \omega_{d-1}
    \left(
    \frac{\sinh(\sqrt{c}\rho)}{\sqrt{c}}
    \right)^{d-1},
\end{equation}
where $\omega_{d-1}$ is the surface area of the unit $(d-1)$-sphere. Since
\begin{equation}
    \sinh(\sqrt{c}\rho)
    =
    \frac{e^{\sqrt{c}\rho}-e^{-\sqrt{c}\rho}}{2},
\end{equation}
we obtain
\begin{equation}
    S_{\mathcal{B}^d}(\rho)
    =
    \Theta\!\left(
    e^{(d-1)\sqrt{c}\rho}
    \right),
    \quad \rho\rightarrow\infty .
\end{equation}
Therefore, the available surface capacity at radius $\rho$ grows exponentially with $\rho$. This is fundamentally different from Euclidean space, where the surface area grows only polynomially:
\begin{equation}
    S_{\mathbb{R}^d}(\rho)=\omega_{d-1}\rho^{d-1}.
\end{equation}

This exponential growth matches the structure of a tree. For a balanced $b$-ary tree, the number of nodes at depth $\ell$ is $b^\ell$, which also grows exponentially with depth. If tree depth is represented by geodesic radius and nodes at depth $\ell$ are placed around radius $\rho_\ell=\gamma\ell$, then the available hyperbolic surface capacity grows as
\begin{equation}
    S_{\mathcal{B}^d}(\rho_\ell)
    =
    \Theta\!\left(
    e^{(d-1)\sqrt{c}\gamma\ell}
    \right).
\end{equation}
Choosing
\begin{equation}
    \gamma=\frac{\log b}{(d-1)\sqrt{c}}
\end{equation}
makes this growth proportional to $b^\ell$. Thus, hyperbolic space can allocate capacity for exponentially many nodes using only a linearly increasing geodesic radius. In contrast, since the surface area of a Euclidean sphere grows as $R^{d-1}$, matching $b^\ell$ nodes at depth $\ell$ requires
\begin{equation}
    R_\ell
    =
    \Omega\!\left(
    b^{\ell/(d-1)}
    \right),
\end{equation}
which grows exponentially with $\ell$. This mismatch explains the crowding issue when many fine-grained concepts must be represented around broad parent concepts in Euclidean space.

The Lorentz distance also naturally reflects both hierarchical depth and branch separation. Consider two points
\(\mathbf{p}=\Phi(\rho,\mathbf{u})\) and
\(\mathbf{q}=\Phi(\eta,\mathbf{v})\), where
\(\mathbf{u}^{\top}\mathbf{v}=\cos\theta\). Substituting the parameterization into the Lorentz distance formula gives
\begin{equation}
    \cosh\!\left(
    \sqrt{c}\,d_{\mathcal{B}}(\mathbf{p},\mathbf{q})
    \right)
    =
    \cosh(\sqrt{c}\rho)\cosh(\sqrt{c}\eta)
    -
    \sinh(\sqrt{c}\rho)\sinh(\sqrt{c}\eta)\cos\theta .
\end{equation}
When $\theta=0$, the two points lie on the same radial direction, and the above equation reduces to
\begin{equation}
    \cosh\!\left(
    \sqrt{c}\,d_{\mathcal{B}}(\mathbf{p},\mathbf{q})
    \right)
    =
    \cosh\!\left(\sqrt{c}(\rho-\eta)\right),
\end{equation}
which implies
\begin{equation}
    d_{\mathcal{B}}(\mathbf{p},\mathbf{q})=|\rho-\eta|.
\end{equation}
Thus, movement along the same radial direction corresponds directly to changes in semantic depth, making it natural to place more generic concepts closer to the origin and more fine-grained concepts farther away. When $\theta>0$, the distance increases with angular separation because
\begin{equation}
    \frac{\partial}{\partial \theta}
    \left[
    \cosh(\sqrt{c}\rho)\cosh(\sqrt{c}\eta)
    -
    \sinh(\sqrt{c}\rho)\sinh(\sqrt{c}\eta)\cos\theta
    \right]
    =
    \sinh(\sqrt{c}\rho)\sinh(\sqrt{c}\eta)\sin\theta
    \ge 0 .
\end{equation}
Since $\cosh^{-1}(\cdot)$ is monotonic, larger angular separation yields larger geodesic distance. Therefore, radial coordinates can encode hierarchical depth, while angular directions can separate different semantic branches.

This geometry is also consistent with the entailment cones used in our hierarchical objective. Recall that the cone aperture is defined as
\begin{equation}
    \mathrm{aper}(\mathbf{p})
    =
    \sin^{-1}\!\left(
    \frac{2\kappa}{\sqrt{c}\|\mathbf{p}\|}
    \right).
\end{equation}
The norm appearing in the aperture increases as a point moves outward in geodesic radius. For example, under the Lorentz parameterization, the spatial norm satisfies
\begin{equation}
    \|\mathbf{p}_s\|
    =
    \frac{\sinh(\sqrt{c}\rho)}{\sqrt{c}},
\end{equation}
which is monotonically increasing in $\rho$. The full Euclidean norm of $\mathbf{p}$ is also monotonically increasing in $\rho$. Therefore, larger-radius points induce smaller cone apertures. To see this explicitly, for fixed \(c\) and \(\kappa\), let \(s=\|\mathbf{p}\|\). When the aperture is well-defined, \(s>\frac{2\kappa}{\sqrt{c}}\), and
\begin{equation}
    \frac{\partial \mathrm{aper}(\mathbf{p})}{\partial s}
    =
    -
    \frac{2\kappa}{\sqrt{c}s^2}
    \cdot
    \frac{1}{
    \sqrt{
    1-\left(
    \frac{2\kappa}{\sqrt{c}s}
    \right)^2
    }}
    <0 .
\end{equation}
Hence, higher-level concepts can have broader descendant regions, while deeper and more specific concepts induce tighter cones. The constraint
\begin{equation}
    \mathrm{ext}(\mathbf{p},\mathbf{q})
    \le
    \mathrm{aper}(\mathbf{p})
\end{equation}
therefore turns a parent--child relation into a geometric inclusion relation in \(\mathcal{B}^d\).

Overall, the Lorentz hyperbolic space is suitable for hierarchical representation because it provides exponential capacity for tree-like branching, a natural depth-radius correspondence, and a distance geometry that separates different branches while supporting ancestor-descendant organization. These properties support our use of hyperbolic space for modeling hierarchical semantic structures in CLIP-based class-incremental learning.

\section{Generating Hierarchical Semantic Tree}
\label{sec:app_prompt}
\begin{figure}
	\centering
     \vspace{-4mm}
    \includegraphics[width=0.9\linewidth]{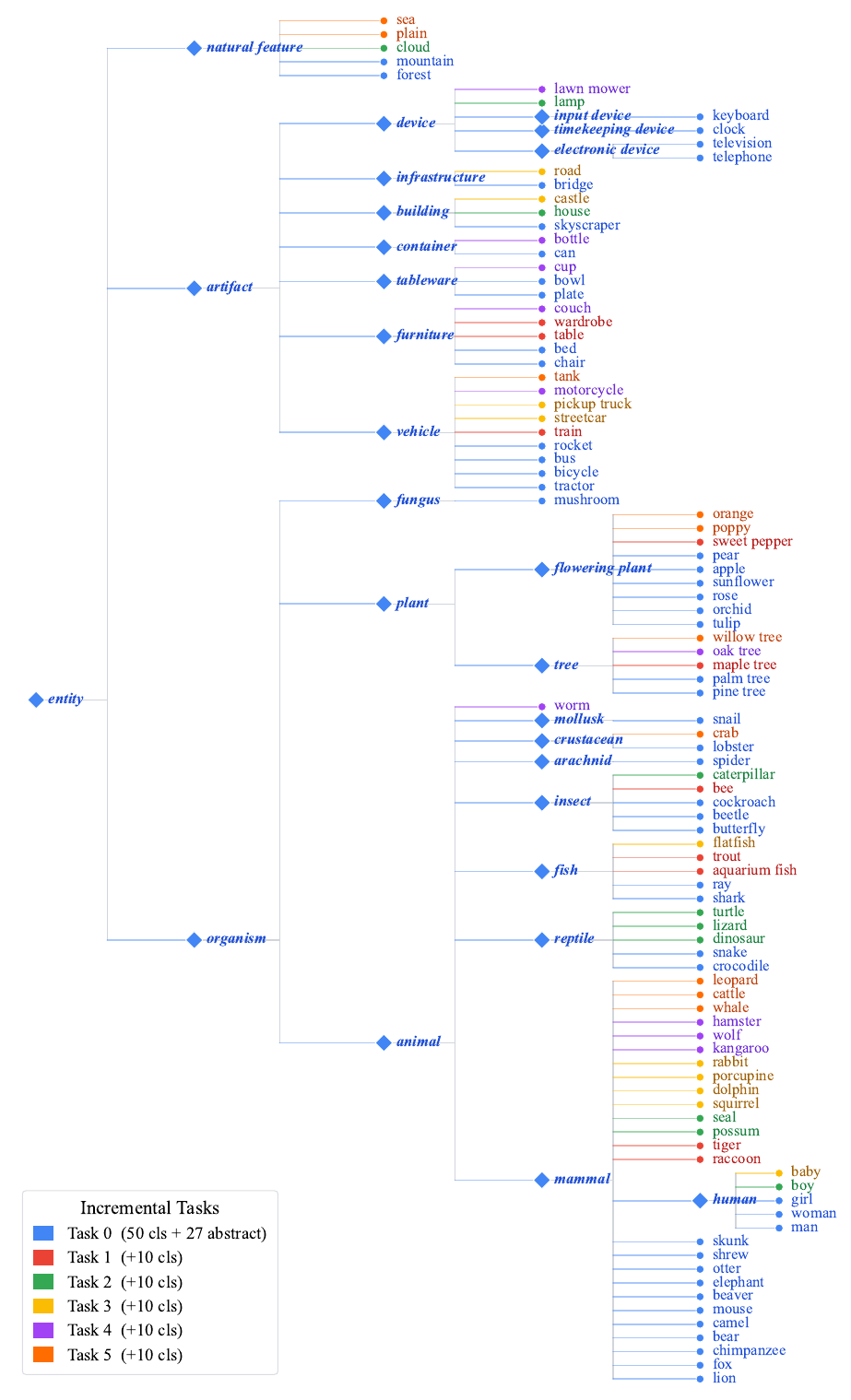}
    \vspace{-3mm}
    \caption{\small Visualization of the semantic tree structure derived from the output JSON on CIFAR100. The root node ``entity'' branches into abstract categories (e.g., ``organism''), which further subdivide until reaching the specific leaf classes from the input list.}
	\label{fig:tree_incremental}
\end{figure}
In this section, we present the full prompts used for hierarchical semantic tree generation and illustrate them with a CIFAR-100 example under the B50 Inc10 incremental setting. To strictly adhere to the class-incremental learning paradigm and prevent any leakage of future classes, our method dynamically constructs the hierarchy \emph{incrementally} in two stages:
\begin{enumerate}[leftmargin=1.5em,itemsep=2pt]
    \item \textbf{Base construction}: build an initial taxonomy from the first task's classes only;
    \item \textbf{Incremental update}: progressively integrate each subsequent task's new classes while \emph{strictly preserving} all existing parent--child edges.
\end{enumerate}

\subsection{Stage 1: Base Taxonomy Construction}
The base prompt takes only the class names available at Task~0 and produces a complete ``is-a'' hierarchy rooted at
\texttt{entity}.

\begin{promptbox}[\textsf{Prompt} \normalfont\itshape (Base Construction)]

You are a meticulous and highly accurate taxonomist and ontologist.
Your goal is to construct an initial hierarchical ``is-a'' taxonomy from a flat list of class names.

\smallskip
\begin{rulebox}
\textbf{Non-negotiable Rules}
\begin{enumerate}[leftmargin=1.2em, itemsep=1pt, label=\textbf{\arabic*)}]
\item \textbf{Total Inclusion} -- The final JSON must contain ALL class names from the input list. Each class
appears exactly once. No omissions, no duplicates.
\item \textbf{Name Preservation} {\small\textcolor{red}{(Critical)}} -- Keep every input class name \emph{exactly}
as given (spelling, spacing, case). Abstract parent nodes you create MAY use \texttt{snake\_case}.
\item \textbf{Single Root} -- All branches connect to one root node: \texttt{"entity"}. No disconnected subtrees.
\item \textbf{True ``is-a'' Relationships} -- Every parent $\to$ child must be a genuine ``is-a-type-of'' relation.
Avoid vague placeholders (\texttt{"object"}, \texttt{"thing"}, \texttt{"item"}) unless logically necessary.
\item \textbf{Minimal but Coherent Abstraction} -- Introduce abstract parents only when necessary. Prefer
semantically meaningful intermediate categories. No unnecessary multi-layer chains.
\item \textbf{Structure Constraints} -- Single valid JSON object. Keys = parent nodes, values = lists of direct
children. Leaf nodes appear only in child lists. No empty lists, no cycles, no self-loops.
\end{enumerate}
\end{rulebox}

\smallskip
\textbf{Self-check before output:}
\checkmark~Every input class appears exactly once \quad
\checkmark~All nodes reachable from \texttt{"entity"} \quad
\checkmark~All edges are valid ``is-a'' relations \quad
\checkmark~No input class renamed

\medskip
\textbf{CLASS LIST (Task 0, 50 classes):}\\[3pt]
{\small\ttfamily
\RaggedRight
\StrSubstitute{["lion", "tulip", "butterfly", "pear", "tractor", "chair", "fox", "rocket", "chimpanzee", "bear",
"snail", "forest", "orchid", "telephone", "mountain", "spider", "woman", "camel", "can", "plate", "man", "pine tree",
"keyboard", "bridge", "palm tree", "mouse", "crocodile", "television", "lobster", "snake", "cockroach", "apple",
"beetle", "mushroom", "beaver", "shark", "bicycle", "girl", "bowl", "bus", "rose", "shrew", "elephant", "otter",
"sunflower", "bed", "ray", "skunk", "clock", "skyscraper"]}{, }{,\allowbreak\ }}

\end{promptbox}

\begin{answerbox}[\textsf{Response} \normalfont\itshape (Base Taxonomy -- Task 0)]
\begin{lstlisting}[style=jsonstyle]
{
"entity": ["organism", "artifact", "natural_feature"],
"organism": ["animal", "plant", "fungus"],
"animal": ["mammal", "reptile", "fish", "insect",
           "arachnid", "crustacean", "mollusk"],
"mammal": ["lion", "fox", "chimpanzee", "bear", "camel",
           "mouse", "beaver", "elephant", "otter", "shrew",
           "skunk", "human"],
"human": ["man", "woman", "girl"],
"reptile": ["crocodile", "snake"],
"fish": ["shark", "ray"],
"insect": ["butterfly", "beetle", "cockroach"],
"arachnid": ["spider"],
"crustacean": ["lobster"],
"mollusk": ["snail"],
"plant": ["tree", "flowering_plant"],
"tree": ["pine tree", "palm tree"],
"flowering_plant": ["tulip", "orchid", "rose", "sunflower",
                    "apple", "pear"],
"fungus": ["mushroom"],
"natural_feature": ["forest", "mountain"],
"artifact": ["vehicle", "furniture", "tableware", "container",
             "building", "infrastructure", "device"],
"vehicle": ["tractor", "bicycle", "bus", "rocket"],
"furniture": ["chair", "bed"],
"tableware": ["plate", "bowl"],
"container": ["can"],
"building": ["skyscraper"],
"infrastructure": ["bridge"],
"device": ["electronic_device", "timekeeping_device",
           "input_device"],
"electronic_device": ["telephone", "television"],
"timekeeping_device": ["clock"],
"input_device": ["keyboard"]
}
\end{lstlisting}
\end{answerbox}

\subsection{Stage 2: Incremental Taxonomy Update}

When a new task arrives, we feed the \emph{current} hierarchy together with the new class list into the incremental
prompt. The key constraint is \textbf{no rewiring}: all existing edges are frozen, and the model may only
\emph{append} new children or introduce new parent keys.

\begin{incpromptbox}[\textsf{Prompt} \normalfont\itshape (Incremental Update)]

You are a meticulous and highly accurate taxonomist and ontologist.
Your goal is to update an existing hierarchical ``is-a'' structure by integrating a list of NEW class names into it.

\smallskip
\begin{rulebox}
\textbf{Non-negotiable Rules}
\begin{enumerate}[leftmargin=1.2em, itemsep=1pt, label=\textbf{\arabic*)}]
\item \textbf{Total Inclusion} -- The final JSON must retain ALL strings from the existing hierarchy AND include ALL
new classes. Do not delete anything.
\item \textbf{Name Preservation} {\small\textcolor{red}{(Critical)}} -- Keep every new class name exactly as given.
New abstract nodes you invent MAY use \texttt{snake\_case}.
\item \textbf{Stability / No Rewiring} {\small\textcolor{red}{(Critical)}} -- Do \textbf{not} remove or change any
existing parent$\to$child edge. Do \textbf{not} move an existing child between parents. You may \textbf{only}:
(a)~append new children to existing parent lists; (b)~add new parent keys; (c)~add new high-level children under
\texttt{"entity"}.
\item \textbf{Minimal New Abstract Nodes} -- Create the minimum number of new grouping nodes required.
\item \textbf{Fallback Bucket} -- If a new class cannot be confidently placed, put it under \texttt{"unclassified"}
(reachable from \texttt{"entity"}).
\end{enumerate}
\end{rulebox}

\smallskip
\textbf{Self-check:}
\checkmark~Every new class appears exactly once \quad
\checkmark~Every existing string still present \quad
\checkmark~No existing edge removed

\medskip
\textbf{EXISTING HIERARCHY:}\\[2pt]
{\small\ttfamily \textcolor{gray}{\{~\textnormal{\itshape ... the Task~0 JSON shown above ...}~\}}}

\medskip
\textbf{NEW CLASSES (Task 1, 10 classes):}\\[3pt]
{\small\ttfamily
\RaggedRight
\StrSubstitute{["maple tree", "sweet pepper", "train", "aquarium fish", "bee", "table", "wardrobe", "raccoon",
"tiger", "trout"]}{, }{,\allowbreak\ }}

\end{incpromptbox}

\begin{answerbox}[\textsf{Response} \normalfont\itshape (Updated Taxonomy -- after Task 1)]
\begin{lstlisting}[style=jsonstyle]
{
"entity": ["organism", "artifact", "natural_feature"],
"organism": ["animal", "plant", "fungus"],
"animal": ["mammal", "reptile", "fish", "insect",
           "arachnid", "crustacean", "mollusk"],
"mammal": ["lion", "fox", "chimpanzee", "bear", "camel",
           "mouse", "beaver", "elephant", "otter", "shrew",
           "skunk", "human", "raccoon", "tiger"],
"human": ["man", "woman", "girl"],
"reptile": ["crocodile", "snake"],
"fish": ["shark", "ray", "aquarium fish", "trout"],
"insect": ["butterfly", "beetle", "cockroach", "bee"],
"arachnid": ["spider"],
"crustacean": ["lobster"],
"mollusk": ["snail"],
"plant": ["tree", "flowering_plant"],
"tree": ["pine tree", "palm tree", "maple tree"],
"flowering_plant": ["tulip", "orchid", "rose", "sunflower",
                    "apple", "pear", "sweet pepper"],
"fungus": ["mushroom"],
"natural_feature": ["forest", "mountain"],
"artifact": ["vehicle", "furniture", "tableware", "container",
             "building", "infrastructure", "device"],
"vehicle": ["tractor", "bicycle", "bus", "rocket", "train"],
"furniture": ["chair", "bed", "table", "wardrobe"],
"tableware": ["plate", "bowl"],
"container": ["can"],
"building": ["skyscraper"],
"infrastructure": ["bridge"],
"device": ["electronic_device", "timekeeping_device",
           "input_device"],
"electronic_device": ["telephone", "television"],
"timekeeping_device": ["clock"],
"input_device": ["keyboard"]
}
\end{lstlisting}

\smallskip
\noindent\textbf{Observations:}
\begin{itemize}[leftmargin=1.2em, itemsep=1pt]
\item All 10 new classes are correctly inserted into existing parent lists.
\item No existing parent$\to$child edge has been removed or rewired.
\item New classes are appended to semantically appropriate parents: \texttt{"raccoon"} and \texttt{"tiger"} $\to$
\texttt{mammal}; \texttt{"train"} $\to$ \texttt{vehicle}; \texttt{"maple tree"} $\to$ \texttt{tree}; etc.
\item Every path still connects to \texttt{"entity"}.
\item This process repeats identically for Tasks~2--5, each time feeding the latest accumulated hierarchy and the
new batch of 10~classes.
\end{itemize}
\end{answerbox}

Figure~\ref{fig:tree_incremental} visualizes the final hierarchical semantic tree for CIFAR-100 under the B50 Inc10
setting, constructed through the above two-stage prompting process. Nodes are color-coded by the incremental task
that introduced them: \textcolor[HTML]{4285F4}{\textbf{blue}} denotes Task~0 (the 50 base classes and all abstract
grouping nodes), while subsequent tasks are shown in \textcolor[HTML]{EA4335}{\textbf{red}} (Task~1),
\textcolor[HTML]{34A853}{\textbf{green}} (Task~2), \textcolor[HTML]{FBBC04}{\textbf{yellow}} (Task~3),
\textcolor[HTML]{A142F4}{\textbf{purple}} (Task~4), and \textcolor[HTML]{FF6D01}{\textbf{orange}} (Task~5). Within
each parent node, children added by later tasks are appended below existing ones, visually reflecting the incremental
growth of the taxonomy. Notably, the no-rewiring constraint ensures that all blue edges established at Task~0 remain
unchanged throughout, while colored branches extend outward as new classes arrive.

\noindent\textbf{Validation and fallback}. 
After generation, we check whether all seen classes appear exactly once, all nodes are reachable from the root, and no previous edge is removed or rewired. If the output violates these constraints, we re-query the model; unresolved classes are placed under the ``unclassified'' node.

\section{{Supplementary Results and Methods}}
\label{sec:app_more_results}

\subsection{Evaluation Across Multiple Random Seeds}
\begin{figure*}[ht]
    \centering
    \begin{subfigure}[t]{0.32\textwidth}
        \centering
        \includegraphics[width=1\linewidth]{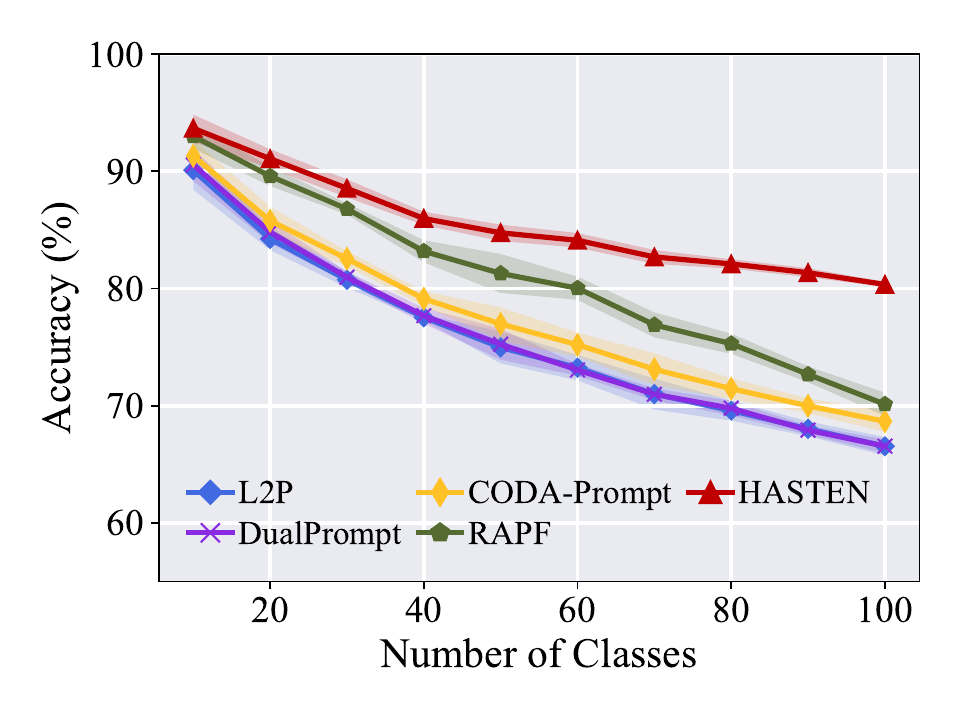}
        \caption{Results on ImageNet-R B0 Inc20 with multiple runs. \name consistently outperforms other methods by a substantial margin.}
        \label{fig:app_seed}
    \end{subfigure}%
    \hfill
    \begin{subfigure}[t]{0.32\textwidth}
        \centering
        \includegraphics[width=1\linewidth]{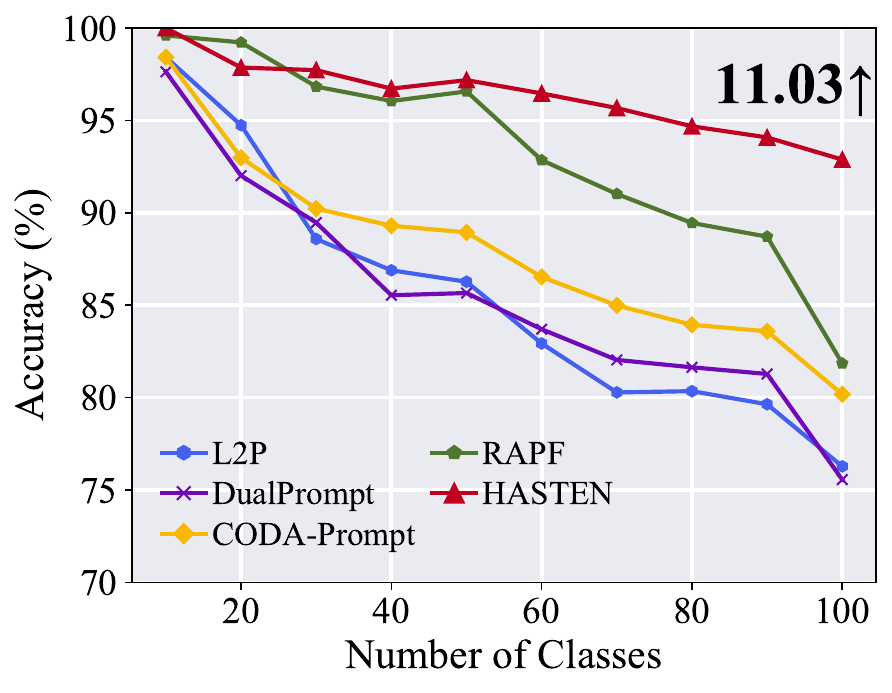}
        \caption{Experiments when using OpenAI weights on UCF B0 Inc10. \name consistently outperforms other methods across different backbone weights.}
        \label{fig:app_backbone}
    \end{subfigure}%
    \hfill
    \begin{subfigure}[t]{0.32\textwidth}
        \centering
        \includegraphics[width=1\linewidth]{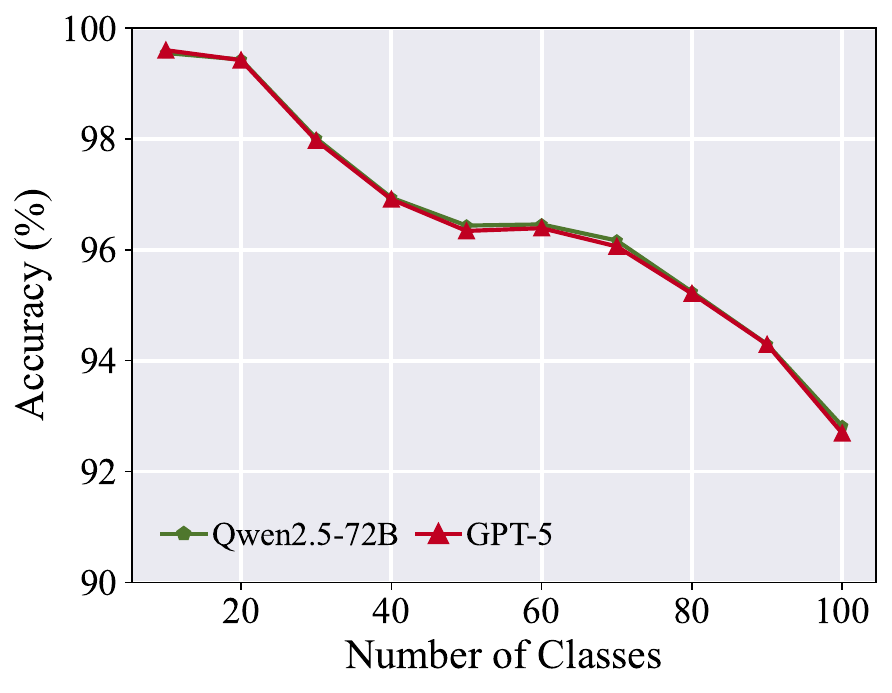}
        \caption{Results on UCF B0 Inc10 with different LLMs for hierarchical semantic tree. \name is robust and compatible with various LLMs.}
        \label{fig:app_llm}
    \end{subfigure}
\caption{Comprehensive analysis of \name on different benchmarks. Our method demonstrates consistent superiority and robustness across various random seeds, pre-trained checkpoints, and LLMs.}
    \label{fig:overall_analysis}
\end{figure*}

In the main paper, all experimental results follow the standard CIL protocol with a fixed class order. To further assess statistical robustness, we additionally repeat the experiments under multiple random class-order splits. Specifically, we conduct five runs with different class-order seeds and report the mean accuracy together with the standard deviation. Figure~\ref{fig:app_seed} shows the results on ImageNet-R B0 Inc20, where the solid lines denote the mean performance and the shaded bands indicate one standard deviation across runs. The results show that \name consistently outperforms the baselines under different class orders, demonstrating strong robustness to randomness in class splitting.

\subsection{Results with Different Backbones}
In the main paper, our experiments mainly use CLIP ViT-B/16 with LAION-400M pre-trained weights~\cite{ilharco2021openclip}. To further examine the generality of our approach, we additionally report results of our method using the OpenAI CLIP pre-trained weights on UCF B0 Inc10, as shown in Figure~\ref{fig:app_backbone}. Across different backbone initializations, \name consistently outperforms competing methods, demonstrating strong robustness to variations in pre-trained models.

\begin{figure*}[t]
    \centering
    \captionsetup{skip=2pt}
    \captionsetup[subfigure]{font=scriptsize, skip=1pt}

    \begin{subfigure}[t]{0.32\textwidth}
        \centering
        \includegraphics[width=\linewidth]{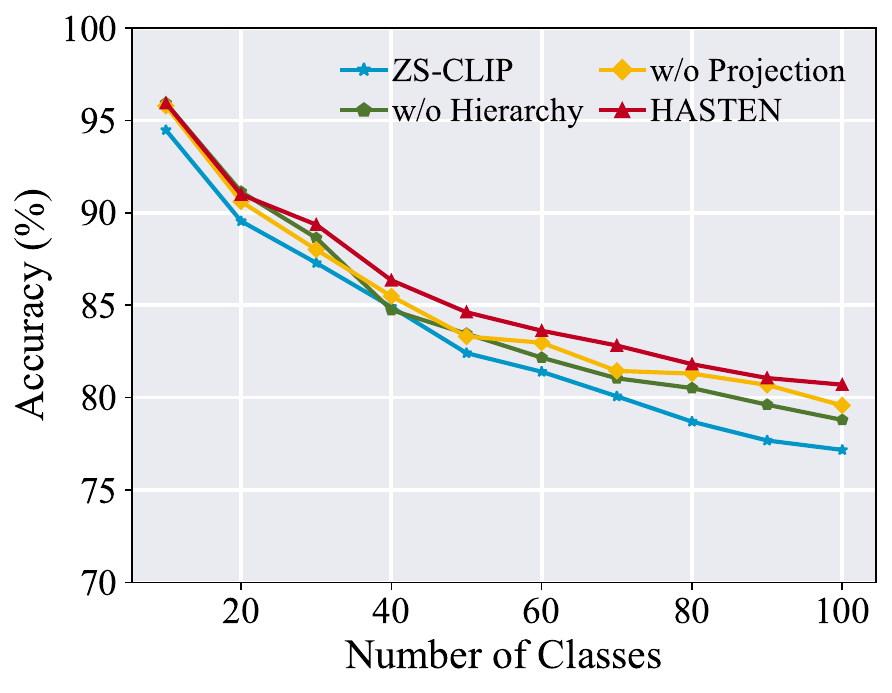}
        \caption{ImageNet-R B0 Inc20}
        \label{fig:app_ablation_imagenet}
    \end{subfigure}
    \hfill
    \begin{subfigure}[t]{0.32\textwidth}
        \centering
        \includegraphics[width=\linewidth]{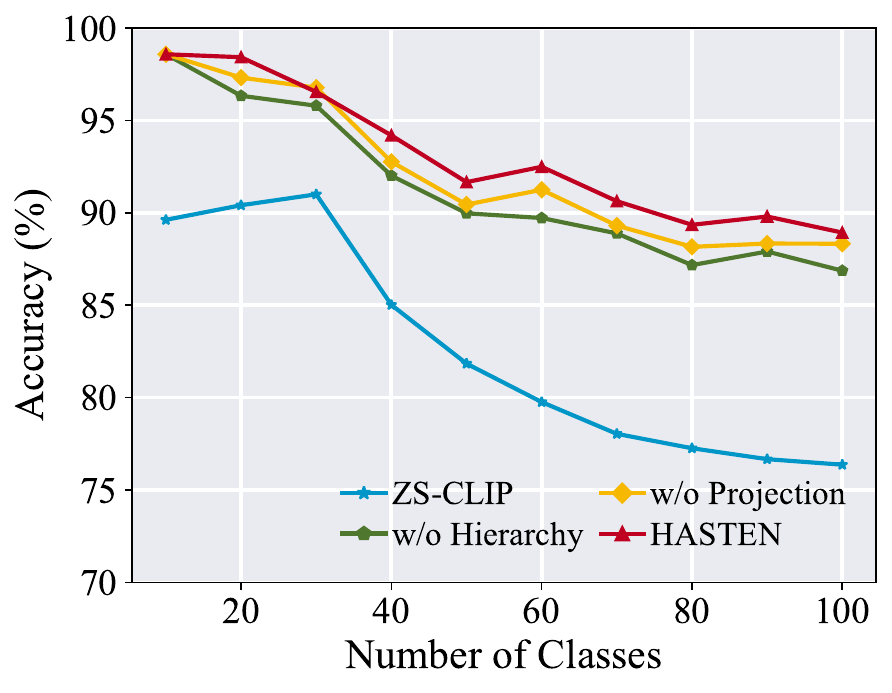}
        \caption{Cars B0 Inc10}
        \label{fig:app_ablation_cars}
    \end{subfigure}
    \hfill
    \begin{subfigure}[t]{0.32\textwidth}
        \centering
        \includegraphics[width=\linewidth]{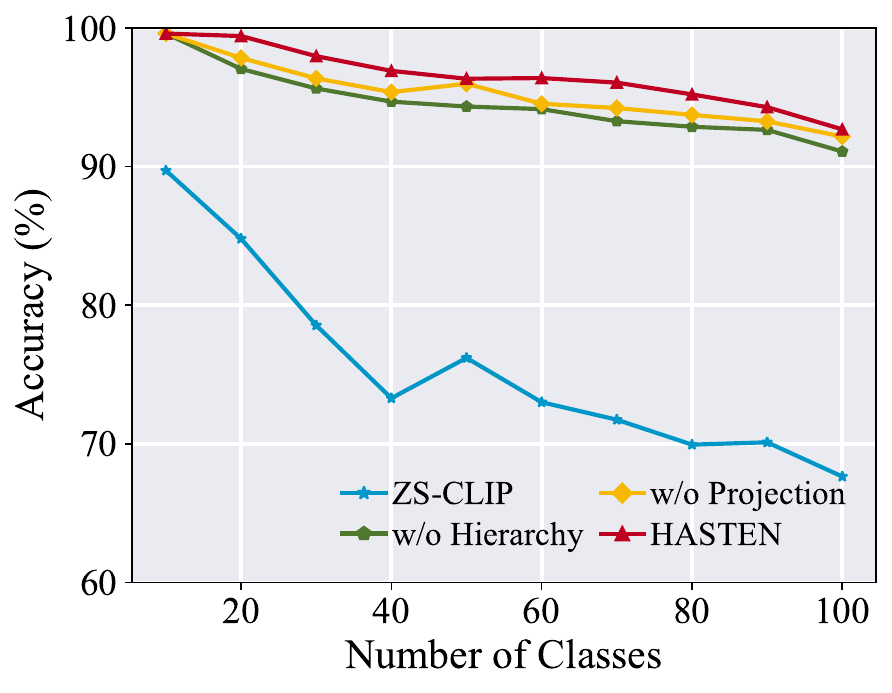}
        \caption{UCF B0 Inc10}
        \label{fig:app_ablation_ucf}
    \end{subfigure}

\caption{\small
Additional ablation studies on ImageNet-R, Cars, and UCF. 
The full \name model consistently outperforms variants without hierarchical supervision or TP projection, confirming the complementary roles of hierarchy preservation and shared-mapper stabilization.
}
    \label{fig:app_ablation_more}
\end{figure*}

\subsection{Component Analysis}
\label{sec:app_component_analysis}
To further investigate the contribution of individual components in \mame, we conduct additional ablation studies on the hierarchy-aware objectives, the parent--child separation term, and the TP regularizer. For individual loss ablations, all other components remain unchanged. We also construct a Euclidean counterpart by replacing the Lorentz hyperbolic representation with a Euclidean projection, the geodesic distances in the hyperbolic contrastive loss with Euclidean distances, and the entailment-cone constraints with Euclidean distance-based parent--child margins, while keeping the remaining experimental settings unchanged. All experiments are conducted under the B0 Inc10 setting, and we report both average accuracy $\bar{\mathcal{A}}$ and final-stage accuracy $\mathcal{A}_B$.

\begin{table}[t]
    \caption{\small
    Component analysis of \name under the B0 Inc10 setting.
    Each entry reports average accuracy $\bar{\mathcal{A}}$ and final-stage
    accuracy $\mathcal{A}_B$.
    }
    \label{tab:component-analysis}
    \centering
    \footnotesize
    \setlength{\tabcolsep}{18pt}
    \renewcommand{\arraystretch}{0.95}
    \resizebox{\columnwidth}{!}{%
    \begin{NiceTabular}{@{}p{0.38\columnwidth}cccccc}
        \toprule
        \multirow{2}{*}{Variant}
        & \multicolumn{2}{c}{Cars}
        & \multicolumn{2}{c}{CIFAR100}
        & \multicolumn{2}{c}{UCF101}
        \\
        \cmidrule(lr){2-3}
        \cmidrule(lr){4-5}
        \cmidrule(lr){6-7}
        & $\bar{\mathcal{A}}$ & $\mathcal{A}_B$
        & $\bar{\mathcal{A}}$ & $\mathcal{A}_B$
        & $\bar{\mathcal{A}}$ & $\mathcal{A}_B$
        \\
        \midrule

        \rowcolor{LightCyan}
        \name (Full)
        & 93.06 & 88.93
        & 88.64 & 82.51
        & 96.49 & 92.69
        \\

        w/o $\mathcal{L}_{\mathrm{hier}}^{\mathrm{txt}\rightarrow\mathrm{txt}}$
        & 91.99 & 87.56
        & 86.75 & 80.48
        & 94.21 & 91.80
        \\

        w/o $\mathcal{L}_{\mathrm{hier}}^{\mathrm{txt}\rightarrow\mathrm{img}}$
        & 92.38 & 88.25
        & 87.37 & 81.33
        & 95.84 & 92.54
        \\

        w/o parent--child separation term
        & 92.93 & 88.73
        & 88.68 & 82.62
        & 96.16 & 92.57
        \\

        w/o $\mathcal{L}_{\mathrm{tp}}$
        & 90.98 & 86.25
        & 85.70 & 80.55
        & 94.06 & 90.46
        \\

        Euclidean counterpart
        & 90.14 & 85.86
        & 84.58 & 78.41
        & 93.25 & 88.57
        \\

        \bottomrule
    \end{NiceTabular}}
\end{table}

As shown in Table~\ref{tab:component-analysis}, removing either $\mathcal{L}_{\mathrm{hier}}^{\mathrm{txt}\rightarrow\mathrm{txt}}$ or $\mathcal{L}_{\mathrm{hier}}^{\mathrm{txt}\rightarrow\mathrm{img}}$ consistently degrades performance across all three datasets, indicating the complementary benefits of hierarchical semantic supervision and visual-textual alignment. Removing the parent-child separation term has a relatively minor impact, with slight performance decreases on Cars and UCF101 and marginal improvements on CIFAR100, suggesting that its contribution varies across datasets. Meanwhile, removing $\mathcal{L}_{\mathrm{tp}}$ leads to substantial performance drops, supporting its role in stabilizing the shared hyperbolic mapper during continual updates. Finally, the Euclidean counterpart consistently underperforms the full model across all datasets, suggesting that the combined hyperbolic representation and geometry-aware objectives are beneficial for modeling hierarchical semantic relations. Overall, these results highlight the importance of hierarchy-aware supervision, TP regularization, and hyperbolic modeling in \mame.

\subsection{Additional Ablation Studies}
\label{sec:app_ablation}
To further validate the contribution of each component beyond CIFAR100, we conduct additional ablation studies on three representative benchmarks: ImageNet-R B0 Inc20, Cars B0 Inc10, and UCF B0 Inc10. The results are shown in Figure~\ref{fig:app_ablation_more}. Across all datasets, the full \name model consistently achieves the best performance throughout incremental learning. Removing the hierarchical supervision leads to clear and persistent degradation, especially in later stages, indicating that the learned hierarchy is important for mitigating feature drift and preserving cross-modal structure. Removing the TP projection also reduces performance, showing that stabilizing the shared hyperbolic mapper helps suppress accumulated interference from new tasks. Compared with ZS-CLIP, both ablated variants still benefit from adaptation, but neither matches the full model. These results demonstrate that hierarchical structure preservation and TP stabilization provide complementary benefits across datasets with different domains and granularities.

\begin{table*}[t]
    \caption{\small
    Forgetting measure $F_B$ of different methods under different incremental settings. 
    Lower $F_B$ indicates less forgetting, and the best result in each column is highlighted in bold.
    }
    \label{tab:forgetting}
    \vspace{-2mm}
    \centering
    \footnotesize
    \setlength{\tabcolsep}{8pt}
    \renewcommand{\arraystretch}{0.95}
    \resizebox{1\textwidth}{!}{%
    \begin{NiceTabular}{@{}lcccccccc}
        \toprule
        \multirow{2}{*}{Method} 
        & \multicolumn{2}{c}{Food} 
        & \multicolumn{2}{c}{Cars} 
        & \multicolumn{2}{c}{UCF}
        & \multicolumn{2}{c}{SUN} \\
        \cmidrule(lr){2-3}
        \cmidrule(lr){4-5}
        \cmidrule(lr){6-7}
        \cmidrule(l){8-9}
        & B0 Inc10 & B50 Inc10 
        & B0 Inc10 & B50 Inc10 
        & B0 Inc10 & B50 Inc10
        & B0 Inc30 & B150 Inc30 \\
        \midrule
        ZS-CLIP~\cite{radford2021learning} 
        & 5.22 & \textbf{2.84} & 6.41 & 3.14 & 10.77 & 4.74 & 10.16 & 4.97 \\
        SimpleCIL~\cite{zhou2023revisiting} 
        & 6.22 & 3.76 & 5.11 & 2.51 & 4.86 & \textbf{2.31} & 8.12 & \textbf{3.51} \\
        L2P~\cite{wang2022learning} 
        & 24.69 & 20.26 & 7.45 & 4.22 & 8.15 & 6.22 & 22.87 & 17.57 \\
        DualPrompt~\cite{wang2022dualprompt} 
        & 24.28 & 20.77 & 7.35 & 4.71 & 9.26 & 7.95 & 23.54 & 17.63 \\
        CODA-Prompt~\cite{smith2023coda} 
        & 25.82 & 22.41 & 6.14 & 4.59 & 9.80 & 9.21 & 22.71 & 17.69 \\
        RAPF~\cite{huang2024class}  
        & 8.16 & 7.83 & 28.27 & 29.30 & 15.89 & 20.50 & 14.58 & 9.36 \\
        CLG-CBM~\cite{yu2025language} 
        & 11.79 & 9.38 & 6.86 & 4.86 & 8.25 & 6.87 & 14.62 & 12.26 \\
        PROOF~\cite{zhou2023learning} 
        & 16.58 & 16.94 & 4.48 & \textbf{2.30} & 4.22 & 4.85 & 18.33 & 15.71 \\
        \midrule
        \rowcolor{LightCyan}
        \name (Ours) 
        & \textbf{4.86} & 2.86 
        & \textbf{4.30} & 2.68 
        & \textbf{4.12} & 3.22
        & \textbf{7.32} & 3.66 \\
        \bottomrule
    \end{NiceTabular}}
    \vspace{-3mm}
\end{table*}

\subsection{Forgetting Analysis}
\label{sec:app_forgetting}

We further evaluate the forgetting behavior of different methods in Table~\ref{tab:forgetting}. 
Following the standard CIL protocol~\citep{rebuffi2017icarl,zhou2023learning}, we compute the forgetting measure after the final incremental stage as
\begin{equation}
      F_{B} = \frac{1}{B-1}\sum_{b=1}^{B-1}
      \max_{l \in \{b, \dots, B-1\}} 
      \big(\mathcal{A}_{l,b}- \mathcal{A}_{B,b}\big),
\end{equation}
where \(\mathcal{A}_{l,b}\) denotes the accuracy on task \(b\) after learning stage \(l\), and \(\mathcal{A}_{B,b}\) denotes the accuracy on task \(b\) after the final stage \(B\). 
A lower \(F_B\) indicates that the method better preserves previously learned task knowledge.

As shown in Table~\ref{tab:forgetting}, \name achieves the lowest forgetting in all B0 settings on Food, Cars, UCF, and SUN, demonstrating its ability to preserve earlier knowledge when learning starts without base classes. 
In the half-base settings, \name remains competitive and obtains forgetting scores close to the best methods, while also achieving stronger overall accuracy as reported in the main benchmark results. 
Compared with prompt-based methods such as L2P, DualPrompt, and CODA-Prompt, \name consistently shows much smaller forgetting across datasets, indicating that hierarchical supervision and TP stabilization effectively reduce accumulated interference during incremental learning. 
Although ZS-CLIP and SimpleCIL show low forgetting in some settings, they perform limited or no task-specific adaptation and mainly serve as reference baselines. 
Overall, these results show that \name improves incremental performance while maintaining strong resistance to forgetting across diverse datasets and class splits.

\subsection{Robustness to Different Tree Constructors}
In the main paper, we use GPT-5 to generate the hierarchical semantic tree. To examine whether \name depends on a specific LLM, we replace GPT-5 with Qwen2.5-72B~\cite{qwen} for tree construction and evaluate the resulting hierarchy on UCF B0 Inc10. As shown in Figure~\ref{fig:app_llm}, the two tree constructors yield highly consistent performance across incremental stages. This indicates that \name does not rely on GPT-5 itself, but on the hierarchical structure used as supervision. Therefore, the semantic tree can also be generated by other LLMs or replaced by curated taxonomies and domain ontologies when available.
\begin{figure}[t]
    \centering
    \includegraphics[width=0.95\columnwidth]{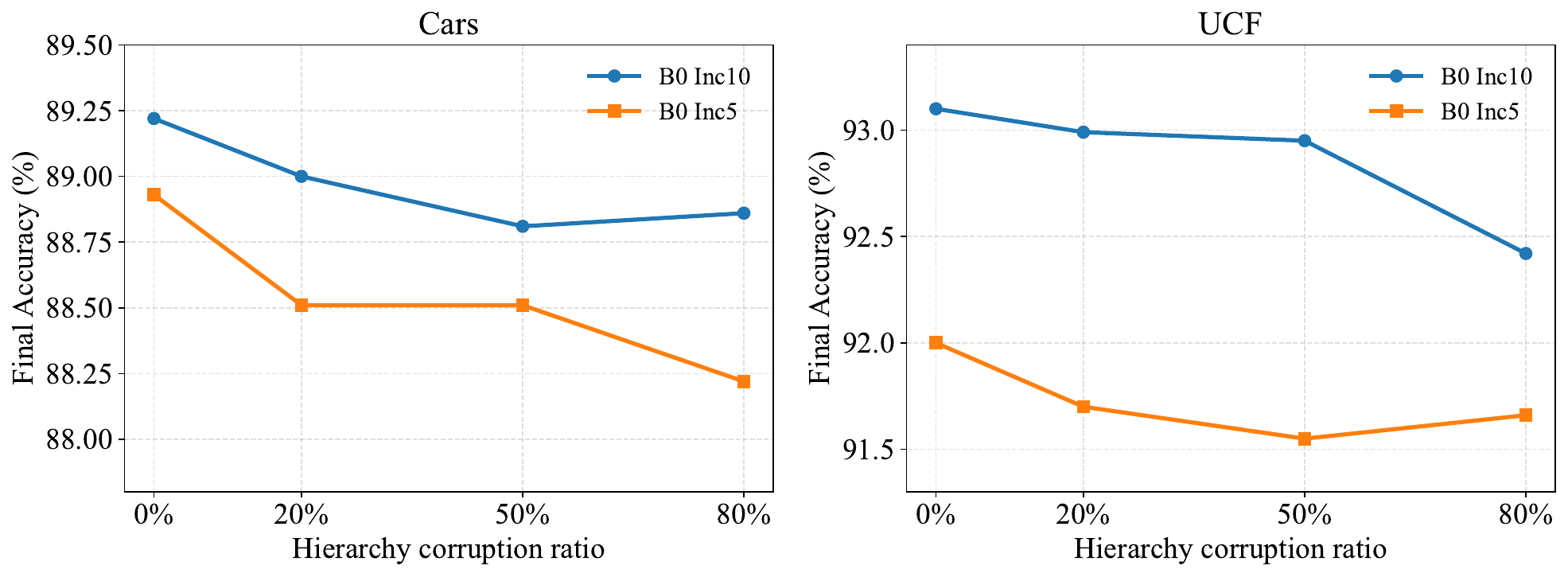}
    \vspace{-2mm}
\caption{\small Robustness to noisy hierarchical supervision on Cars and UCF. 
We corrupt parent-child relations in the semantic tree and report final-stage accuracy. 
Clean hierarchies consistently perform best, showing the benefit of accurate hierarchical supervision.}
    \label{fig:app_hierarchy_noise}
    \vspace{-4mm}
\end{figure}
\\\textbf{Robustness to noisy hierarchies.}
To further examine the dependence of \name on the quality of the semantic tree, we intentionally corrupt the parent-child relations and report the final-stage accuracy in Figure~\ref{fig:app_hierarchy_noise}. 
For each task-wise cumulative tree, we randomly reassign 20\%, 50\%, or 80\% of existing parent-child edges to other valid internal parent nodes while preserving a valid DAG structure and keeping the numbers of nodes and edges unchanged. 
As shown in the figure, the clean hierarchy consistently achieves the best final accuracy across both datasets and both incremental settings. 
As the corruption ratio increases, the final accuracy generally decreases, indicating that correct parent-child relations contribute meaningful supervision for stabilizing incremental learning. 
This experiment is different from the `w/o Hierarchy' ablation: the noisy-tree setting still keeps hierarchical supervision active but injects incorrect structural relations, whereas `w/o Hierarchy' removes the semantic-tree supervision entirely. 
Therefore, the results show that hierarchical supervision itself is useful, and its effectiveness is further strengthened when the semantic tree provides more accurate structural relations.\looseness -1

\subsection{Generalization of the Frozen CLIP Backbone}
\label{sec:app_generalization}
\name builds its hierarchy-aware adaptation on top of frozen CLIP visual and textual encoders. 
The original CLIP backbone is not updated during continual training, so its pre-trained representation and zero-shot matching ability remain directly accessible. 
The hierarchy-aware modules and the shared hyperbolic mapper serve as additional adaptation layers for organizing seen incremental classes with semantic structure, rather than replacing the original CLIP feature space. 
Therefore, when evaluating unseen datasets or open-vocabulary categories, one can still directly use the original frozen CLIP pathway. 
This design allows \name to improve class-incremental learning with hierarchy-aware adaptation while keeping CLIP's original generalization ability available.

\subsection{Descriptions of Compared Methods}
\label{app:sub_method}
In this section, we provide details of the methods compared in the main paper. To ensure a fair comparison, all methods are evaluated using the same pre-trained model. The methods listed in Table~\ref{tab:benchmark} are described as follows:

\begin{itemize}
\item {\bf Finetune}: initializes with a pre-trained CLIP model and finetunes it for each new task. Consequently, it experiences significant catastrophic forgetting on previous tasks.
\item \textbf{SimpleCIL~\cite{zhou2023revisiting}:} utilizes only the pre-trained image encoder, excluding the text encoder. As a result, we remove the text branch in the pre-trained CLIP and evaluate the model using just the visual branch. The frozen image encoder is used to generate class prototypes for each new class, with a cosine classifier employed for classification. Since the model isn't updated with backpropagation, this method highlights the generalizability of the pre-trained vision encoder for downstream tasks.
\item \textbf{ZS-CLIP~\cite{radford2021learning}:} freezes the pre-trained CLIP model and predicts the logits of incoming classes based on cosine similarity. It serves as a benchmark for evaluating the performance of the pre-trained CLIP on downstream tasks.
\item {\bf L2P~\cite{wang2022learning}}: utilizes only the visual branch of CLIP. During the update process, it freezes the pre-trained weights and applies visual prompt tuning~\cite{jia2022visual} to adapt to new task features. It generates instance-specific prompts through a prompt pool, constructed via key-value mapping.
\item  {\bf DualPrompt~\cite{wang2022dualprompt}}: extends L2P by introducing two types of prompts, namely general and expert prompts. The rest of the procedure remains identical to L2P, including the use of a prompt pool to create instance-specific prompts. This approach also relies solely on the visual branch of CLIP.
\item {\bf CODA-Prompt~\cite{smith2023coda}}: addresses the limitations of instance-specific prompt selection;  this method replaces the prompt selection process with prompt reweighting. It uses attention-based recombination of prompts instead, while still involving only the visual branch of CLIP.
\item {\bf RAPF~\cite{huang2024class}}: enhances the continual learning capacity of CLIP by combining a hard class separation loss with decomposed parameter fusion to integrate new knowledge into the CLIP model.
\item {\bf PROOF~\cite{zhou2023learning}}: aims to improve CLIP's continual learning capabilities by learning expandable projection layers and a cross-modal fusion module. The historical prototypes of visual and textual features are passed through the cross-modal fusion module for further alignment.
\item \textbf{CLG-CBM~\cite{yu2025language}:} utilizes a concept bottleneck layer to align semantics with CLIP, enabling the learning of human-understandable and cross-task generalizable concepts.
\end{itemize}

\section{Hyperparameter Sensitivity Analysis}
\label{sec:app_sensitivity}

In the main paper, we presented a robustness analysis for the text-to-text hierarchical weight $\lambda_1$ and the hyperbolic mapping regularizer weight $\beta$. 
This section provides complementary sensitivity analyses for additional key hyperparameters. 
For the sensitivity analyses, we use a validation split from the original training data. 
Specifically, the training data are split into training and validation subsets with a ratio of $4{:}1$, and all sensitivity results are reported as average accuracies across incremental stages on the validation subset.

\begin{figure*}[t]
    \centering
    \captionsetup[subfigure]{
        font=scriptsize,
        skip=1pt,
        justification=centering
    }

    \makebox[\textwidth][c]{%
        \adjustbox{valign=t}{%
            \subcaptionbox{
                Sensitivity to $\lambda_2$ and $\delta$
                \label{fig:app_sensitivity_a}
            }[0.31\textwidth][c]{%
                \includegraphics[width=\linewidth]{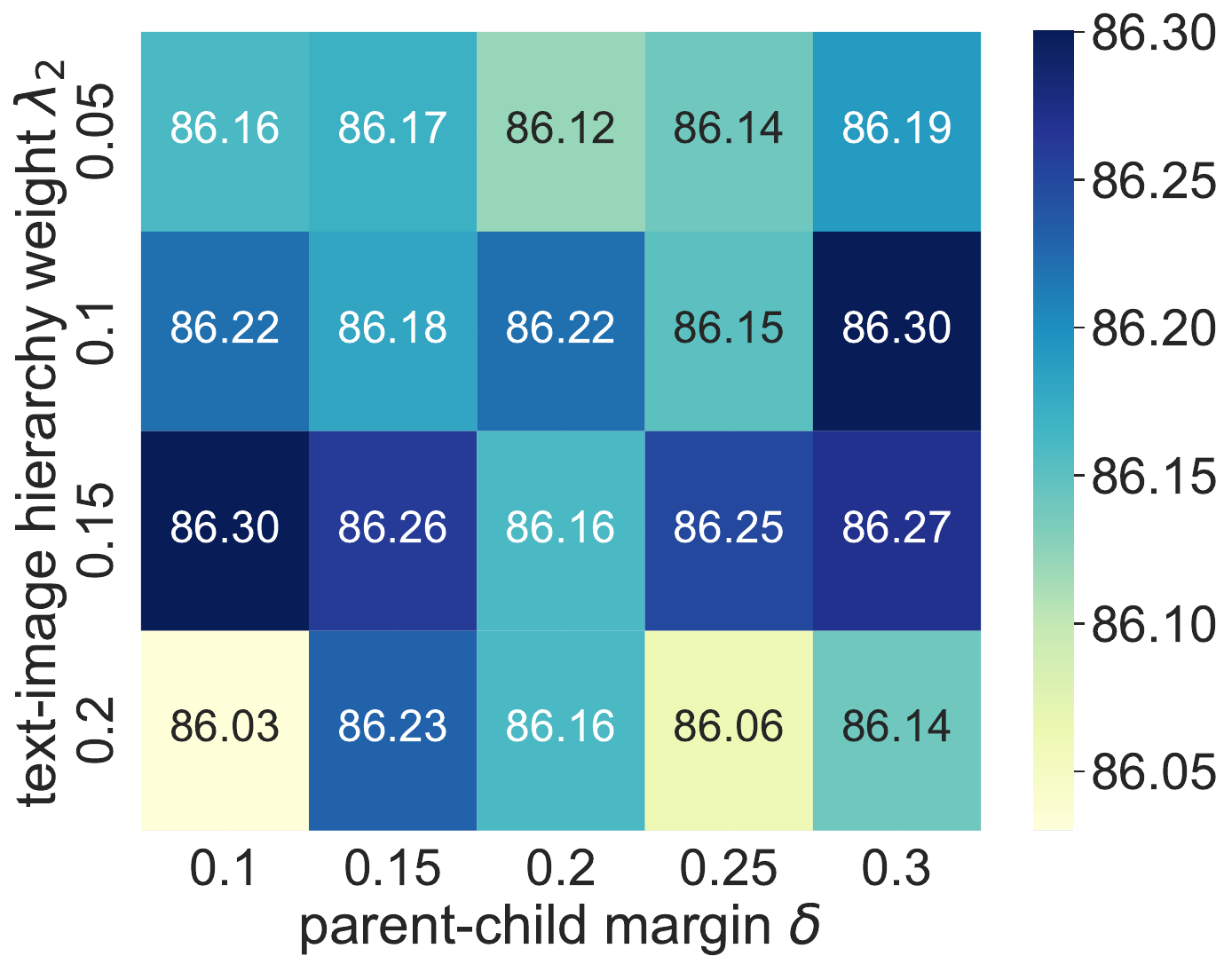}
            }%
        }%
        \hspace{0.015\textwidth}%
        \adjustbox{valign=t}{%
            \subcaptionbox{
                Sensitivity to $c$ and $\kappa$
                \label{fig:app_sensitivity_b}
            }[0.31\textwidth][c]{%
                \includegraphics[width=\linewidth]{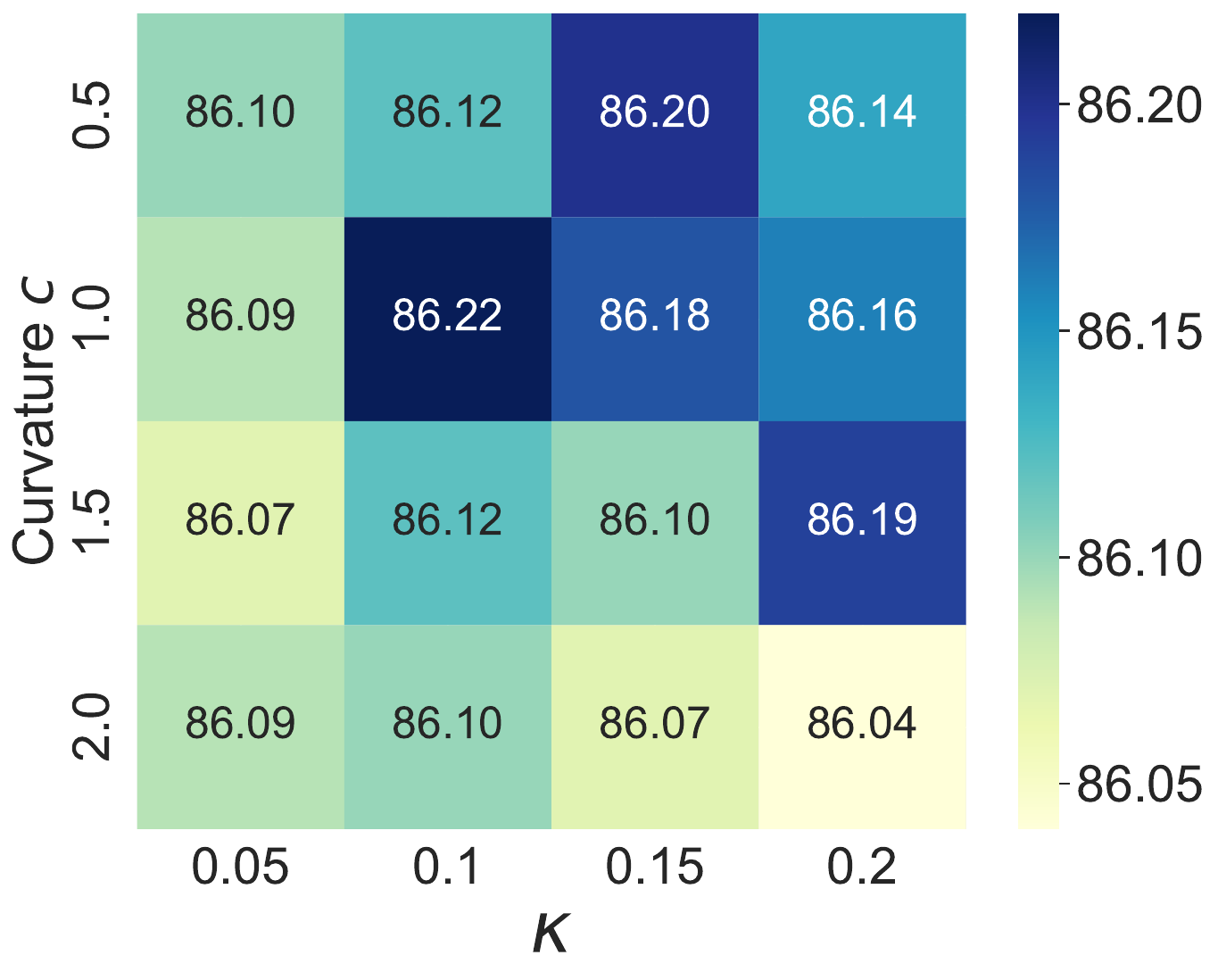}
            }%
        }%
        \hspace{0.015\textwidth}%
        \adjustbox{valign=t}{%
        \captionsetup[subfigure]{skip=18pt}
            \subcaptionbox{
                Trainable parameters
                \label{tab:app_params}
            }[0.32\textwidth][c]{%
                \scriptsize
                \setlength{\tabcolsep}{3pt}
                \renewcommand{\arraystretch}{0.9}
                \resizebox{\linewidth}{!}{%
                    \begin{tabular}{@{}lc@{}}
                        \toprule
                        Method & Trainable Params \\
                        \midrule
                        L2P & 161,330 \\
                        DualPrompt & 333,412 \\
                        CODA-Prompt & 3,916,900 \\
                        PROOF & 1,310,720 \\
                        RAPF & 262,144 \\
                        \name & 786,432 \\
                        \bottomrule
                    \end{tabular}
                }
            }%
        }%
    }

    \caption{\small
    Parameter sensitivity analysis on CUB B0 Inc20.
    \textbf{(a)} Impact of the text-to-image hierarchical loss weight $\lambda_2$ and the parent-child separation margin $\delta$.
    \textbf{(b)} Impact of the foundational geometric priors: hyperbolic curvature $c$ and cone aperture $\kappa$.
    \textbf{(c)} Number of trainable parameters on ImageNet-R B0 Inc20.
    }
    \label{fig:app_sensitivity_all}
    \vspace{-3mm}
\end{figure*}

First, we analyze the \emph{text-to-image hierarchical loss weight} $\lambda_2$ from Eq.~\ref{eq:hier}, which controls how strongly visual features are anchored to their corresponding class text nodes within the parent cone, and the \emph{parent--child separation margin} $\delta$ from Eq.~\ref{eq:text_hier_loss}, which enforces a minimum geodesic distance between parent and child text embeddings to prevent representation collapse. 
We evaluate on the CUB B0 Inc20 setting, sweeping $\lambda_2$ over $\{0.05, 0.10, 0.15, 0.20\}$ and $\delta$ over $\{0.10, 0.15, 0.20, 0.25, 0.30\}$. 
Figure~\ref{fig:app_sensitivity_a} reports the average validation accuracy. 
The results show that \name remains stable under different choices of $\lambda_2$ and $\delta$.

Furthermore, to examine whether our framework is sensitive to geometric configurations across different semantic granularities, we study the foundational geometric parameters: the \emph{curvature} $c$ and the \emph{cone half-aperture parameter} $\kappa$. 
The curvature controls the volume expansion rate of the Lorentz model, while $\kappa$ determines the strictness of the partial-order entailment cones. 
We sweep $c \in \{0.5, 1.0, 1.5, 2.0\}$ and $\kappa \in \{0.05, 0.10, 0.15, 0.20\}$. 
Figure~\ref{fig:app_sensitivity_b} reports the average validation accuracy, which remains tightly bounded between 86.04\% and 86.22\% across all configurations. 
These results indicate that \name does not require exhaustive dataset-specific tuning and remains stable under a broad range of hyperparameter choices.

\section{Efficiency and Overhead Analysis}
\label{sec:app_params}
\subsection{Trainable Parameter Analysis}
\begin{figure}[ht]
    \centering
    \begin{subfigure}[t]{0.49\textwidth}
	\centering
    \includegraphics[width=1\linewidth]{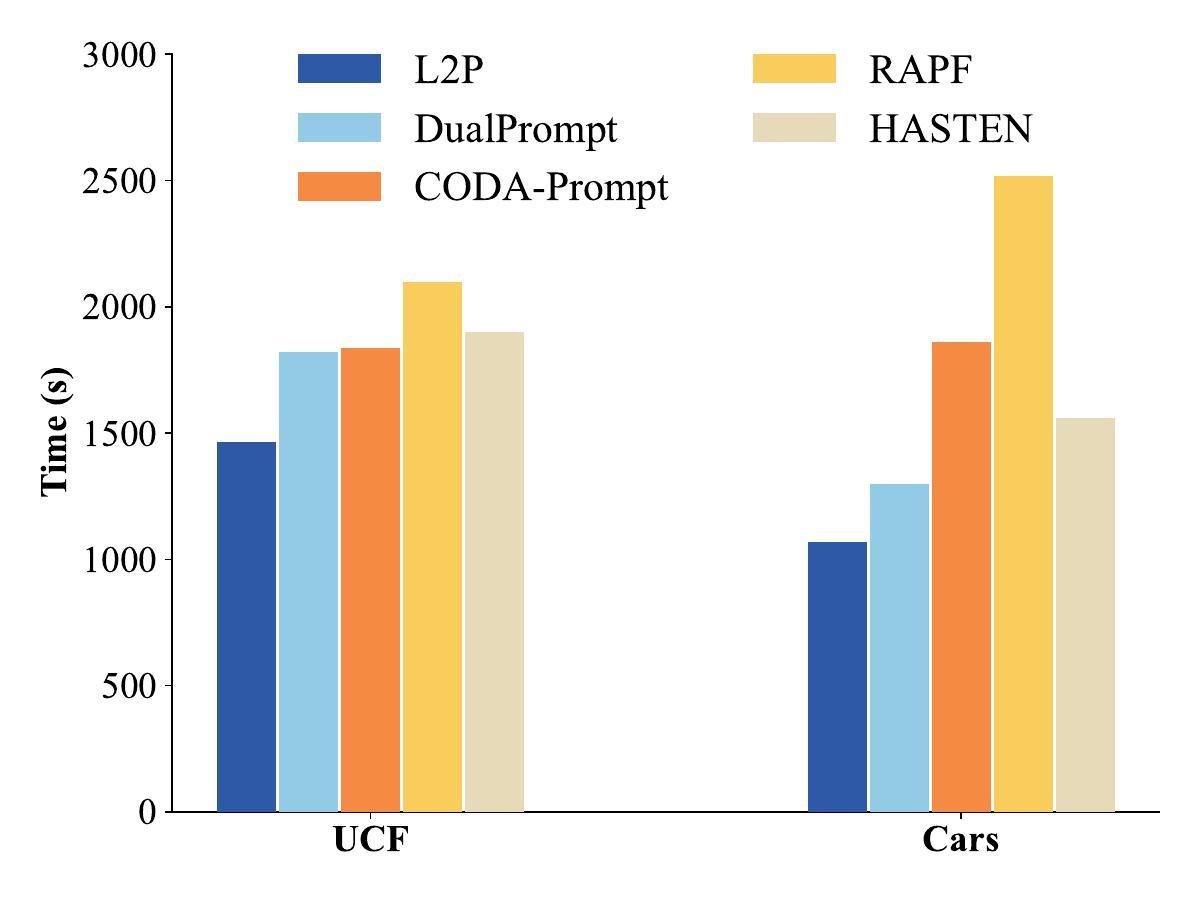}
    \caption{Running time comparison. \name achieves the best performance while maintaining a training time comparable to other compared methods.}
	\label{fig:app_runtime}
    \end{subfigure}%
    \hfill
    \begin{subfigure}[t]{0.49\textwidth}
	\centering
    \includegraphics[width=1\linewidth]{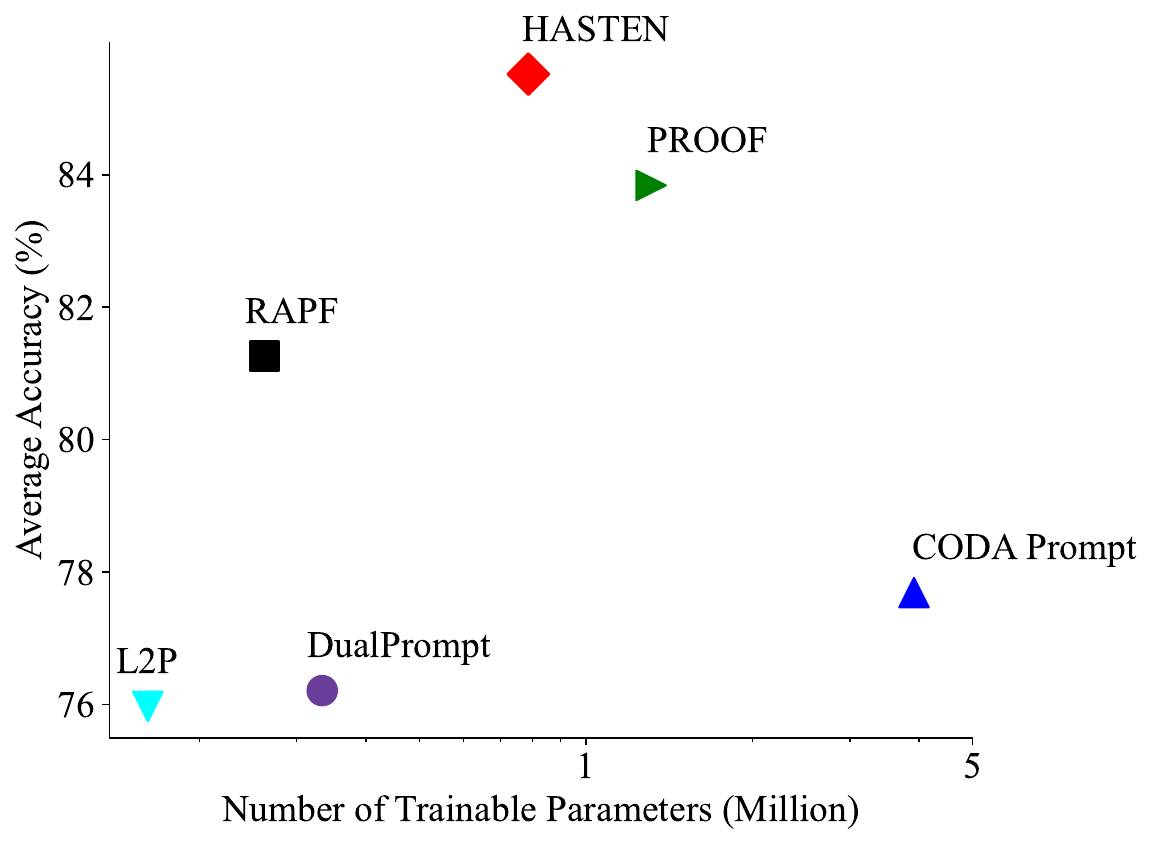}
    \caption{Comparison of accuracy and trainable parameters of different methods on ImageNet-R B0 Inc20.}
	\label{fig:app_param_acc}
    \end{subfigure}
\caption{Efficiency analysis. \textbf{(a)} Running time comparison across methods. \textbf{(b)} Trade-off between accuracy and the number of trainable parameters on ImageNet-R B0 Inc20.}
    \label{fig:app_efficiency}
\end{figure}

In this paper, we design \name by extending a hierarchy-aware module per task and incorporating a task-shared hyperbolic mapping layer. During inference, the features are aggregated into the final Euclidean feature.
Specifically, when learning task $b$, we freeze the hierarchy-aware modules from all previous tasks
$\{1,\dots,b{-}1\}$ and only train the newly added module for task $b$. Therefore, the number of \emph{active}
(trainable) parameters at every stage remains constant: $2 \times d \times d$ for the current hierarchy-aware modules
($H_i^b$ and $H_t^b$), plus $d \times d$ for the global hyperbolic mapping layer $\mathrm{TP}$, yielding $3 \times d
\times d$ in total. Moreover, as illustrated in Section~4.2 of the main paper, because all hierarchy-aware modules are
linear layers, the frozen modules accumulated from previous tasks can be \emph{reparameterized} by simply summing
their weights, \ie, $\sum_{p=1}^{b-1} H_i^p$ and $\sum_{p=1}^{b-1} H_t^p$, each collapsing into a single $d \times d$
matrix. This means the \emph{stored} parameter size of the hierarchy-aware modules stays at $2 \times d \times d$
regardless of how many tasks have been learned, rather than growing linearly with the number of tasks. Combined with
the $d \times d$ parameters of $\mathrm{TP}$, the total parameter footprint of \name is always $3 \times d \times d$.
We further report the number of trainable parameters for representative methods in Figure~\ref{tab:app_params}. In
addition, we visualize the relationship between trainable parameter size and average accuracy in
Figure~\ref{fig:app_param_acc}, showing that \name achieves strong performance with a comparable parameter budget. As
shown in the table, \name has a comparable number of trainable parameters to other competitors, while achieving the
best performance.

\subsection{Running Time Comparison}
In this section, we present a running time comparison of different methods. To ensure a fair evaluation, all methods are evaluated under the same experimental setting. Figure~\ref{fig:app_runtime} reports the total running time on UCF B0 Inc10 and Cars B0 Inc10, including both training and inference/evaluation. As shown in the figure, \name achieves the best performance while maintaining competitive running time compared with baselines such as CODA-Prompt and RAPF.

\begin{table}[t]
    \caption{\small Overhead comparison on \textbf{CUB B0 Inc20}. 
    All methods start from the same pre-trained CLIP backbone and are evaluated under the same hardware environment, using a \textbf{single RTX 3090 GPU} per method.
    ``Total Time'' denotes the end-to-end running time from the start of continual training to the completion of the final evaluation.
    ``Peak GPU'' denotes the peak training GPU memory of the current process, and ``Infer.'' reports the average evaluation latency per test image aggregated over all sessions.
    Lower is better for all overhead metrics, while higher is better for $\bar{\mathcal{A}}$.}
    \label{tab:overhead_cub_b0_inc20}
    \centering
    \resizebox{\columnwidth}{!}{%
    \begin{NiceTabular}{@{}lcccccc @{}}
        \toprule
        Method 
        & $\bar{\mathcal{A}}$ 
        & Total Time (min)$\downarrow$
        & Train Time (min)$\downarrow$
        & Eval Time (min)$\downarrow$
        & Peak GPU (GB)$\downarrow$
        & Infer. (ms/img)$\downarrow$ \\
        \midrule
        SimpleCIL~\cite{zhou2023revisiting} & 83.81 & \textbf{1.24} & \textbf{0.55} & \textbf{0.69} & \textbf{0.92} & \textbf{3.20} \\
        DualPrompt~\cite{wang2022dualprompt} & 69.49 & 61.77 & 59.82 & 1.95 & 6.95 & 9.06 \\
        CODA-Prompt~\cite{smith2023coda} & 73.27 & 66.43 & 64.26 & 2.17 & 15.53 & 10.09 \\
        L2P~\cite{wang2022learning} & 70.57 & 73.66 & 71.60 & 2.07 & 2.85 & 9.60 \\
        RAPF~\cite{huang2024class} & 79.26 & 16.95 & 16.08 & 0.87 & 12.25 & 4.06 \\
        PROOF~\cite{zhou2023learning} & 82.86 & 46.15 & 44.57 & 1.58 & 1.30 & 7.34 \\
        CLG-CBM~\cite{yu2025language} & 79.33 & 35.44 & 34.75 & 0.70 & 1.17 & 3.23 \\
        \midrule
        \rowcolor{LightCyan}
        \name (Ours) & \textbf{86.18} & 32.11 & 30.73 & 1.38 & 1.31 & 6.41 \\
        \bottomrule
    \end{NiceTabular}}
    \vspace{-3mm}
\end{table}

\subsection{Overhead Analysis}
\label{sec:app_overhead_cub}

Table~\ref{tab:overhead_cub_b0_inc20} reports the computational overhead on CUB B0 Inc20. 
This supplementary efficiency study is conducted on a single NVIDIA RTX 3090 GPU for each method, while the main experiments follow the hardware setting described in the main paper. 
All methods use the same pre-trained CLIP backbone and are evaluated under the same hardware environment for a fair comparison. 
Among the compared methods, \name achieves the best average accuracy ($86.18\%$), while keeping the overall computational cost within a practical range.

\noindent\textbf{Total runtime.}
\name completes the continual learning process in $32.11$ minutes, including both training and evaluation. 
Although it is slower than lightweight baselines such as SimpleCIL ($1.24$ min) and RAPF ($16.95$ min), it is faster than CLG-CBM ($35.44$ min), PROOF ($46.15$ min), DualPrompt ($61.77$ min), CODA-Prompt ($66.43$ min), and L2P ($73.66$ min). 
This indicates that the hierarchy-aware training and TP stabilization introduce moderate training overhead while improving accuracy.

\noindent\textbf{Inference cost.}
During evaluation, \name requires $6.41$ ms per image on average. 
This is higher than SimpleCIL ($3.20$ ms), CLG-CBM ($3.23$ ms), and RAPF ($4.06$ ms), but lower than PROOF ($7.34$ ms), DualPrompt ($9.06$ ms), L2P ($9.60$ ms), and CODA-Prompt ($10.09$ ms). 
Therefore, the additional hierarchy-aware mapping and hyperbolic logits introduce a moderate inference cost compared with lightweight baselines, while remaining efficient among several optimization-based CLIP CIL methods.

\noindent\textbf{Memory usage.}
For peak training GPU memory, \name uses $1.31$ GB, which is close to PROOF ($1.30$ GB) and CLG-CBM ($1.17$ GB), and much lower than L2P ($2.85$ GB), DualPrompt ($6.95$ GB), RAPF ($12.25$ GB), and CODA-Prompt ($15.53$ GB). 
Although it uses slightly more memory than SimpleCIL ($0.92$ GB), its memory footprint remains lightweight among the compared methods.

Overall, these results indicate that the additional hierarchy-aware modeling in \name does not introduce prohibitive computational cost, and the method maintains a practical efficiency profile while achieving stronger incremental learning performance.

\noindent\textbf{Additional overhead}. 
Beyond the CUB overhead table, we further examine the SVD cost in the B0 Inc10 settings used in Figure~\ref{fig:app_runtime}. 
Each B0 Inc10 benchmark contains 10 incremental tasks, and \name performs SVD only once after each task to refresh the projection subspace. 
For CLIP ViT-B/16, the pre-TP covariance matrix is \(512\times512\). 
Benchmarking SVD on random covariance-like matrices of this size gives an average cost of \(51.82\pm4.41\) ms per decomposition, resulting in about \(0.52\) s total SVD overhead. 
This accounts for only \(0.027\%\) of the total running time on UCF and \(0.033\%\) on Cars, indicating that the projection update is computationally negligible. 
Separately, for the given class order, the stage-wise semantic trees are constructed or expanded outside the training loop and used as fixed hierarchical supervision; thus, LLM querying introduces no training-loop or inference overhead.

\input{table/fig2}
\input{table/fig1}
\section{Full Results on Benchmarks}
\label{sec:app_full_results}

In this section, we provide the complete benchmark results for all compared methods. 
In addition to the complete quantitative comparison in Table~1, we present the full set of incremental learning curves. 
While the main paper reports three representative learning trends, here we include results for all benchmark settings. 
Specifically, Figure~\ref{fig:supp-benchmark-b0} shows the incremental performance under the B0 setting, and Figure~\ref{fig:supp-benchmark-b50} shows the results under the half-base setting. 
Across datasets and splits, \name consistently outperforms competing approaches.

\section{Details of Image Traversals and Text Set Construction}
\label{sec:app_hierarchy_details}
In this section, we present additional image traversal examples on the UCF101 dataset and provide a detailed explanation of the traversal protocol and the text sets used in these visualizations. All traversals are computed on images randomly sampled from the UCF101 test split, with qualitative results shown in Figure~\ref{fig:app_travel}.

\noindent\textbf{Text Set Construction}: 
To properly evaluate the semantic transition from specific to generic concepts, we construct a candidate text pool. For each UCF101 class, we generate a subset of text that includes the original class name, along with short descriptions and related concepts generated by GPT-5. These textual descriptions are intentionally curated to span multiple levels of semantic granularity---ranging from highly specific visual actions to broader categorical terms. The global text pool used for the retrieval during traversal is the union of the text subsets from all classes, supplemented by the universal root node (``entity'' for \mame, and \texttt{[ROOT]} for CLIP).

\noindent\textbf{Traversal and Retrieval Protocol}: 
As described in the main paper, we build an image-to-root traversal in each model’s feature geometry: the Lorentz manifold $\mathcal{B}^d$ for \mame\ and the Euclidean unit hypersphere for CLIP. We uniformly sample 100 points along the trajectory. At each sampled point $\mathbf{l}_t$, we retrieve one text from a hierarchical text set $\mathcal{T}$ (including \texttt{[ROOT]}). For \mame, retrieval is structure-constrained. A text node is a valid ancestor candidate only if it satisfies the
entailment-cone constraint:
\begin{equation}
\mathcal{P}(\mathbf{l}_t)=\left\{\mathbf{p}\in\mathcal{T}\ \middle|\ \mathrm{ext}(\mathbf{p},\mathbf{l}_t)\le
\mathrm{aper}(\mathbf{p})\right\}.
\end{equation}
We then select the nearest valid ancestor in hyperbolic space:
\begin{equation}
\mathbf{p}_t^*=\arg\min_{\mathbf{p}\in\mathcal{P}(\mathbf{l}_t)} d_{\mathcal{B}}(\mathbf{l}_t,\mathbf{p}).
\end{equation}
For CLIP, no structural constraint is used; we directly retrieve
\begin{equation}
\mathbf{u}_t^*=\arg\max_{\mathbf{u}\in\mathcal{T}} \cos(\mathbf{l}_t,\mathbf{u}).
\end{equation}
Therefore, \name performs geometry-aware parent retrieval, while CLIP performs unconstrained nearest-text retrieval.

\noindent\textbf{Analysis of the Results}: 
The retrieved sequences show how semantics change as an embedding moves toward the root. In \mame, image and text embeddings are organized by parent-centered entailment cones, so each sampled point is matched only to entailment-valid ancestors. This geometry enforces a partial order and yields a gradual shift from specific to increasingly generic concepts. In CLIP, retrieval is unconstrained: points are matched only by cosine similarity in a flat Euclidean space. Because CLIP is trained for global image-text alignment rather than ancestor-descendant ordering, intermediate points often remain closest to the original leaf-level concept and switch only near a decision boundary (often close to \texttt{[ROOT]}). Thus, the downward arrows ($\downarrow$) indicate repeated nearest-neighbor outcomes over many consecutive samples, rather than stable intermediate ancestors. This exposes its lack of inherent hierarchical properties, a semantic gap that \name effectively resolves via its hierarchy-aware hyperbolic representation.

Below, we provide two concrete examples illustrating how the class-specific text subsets are generated via GPT-5:

\begin{figure*}
\footnotesize
\newcolumntype{Z}{>{\centering\arraybackslash}X}
\setlength{\extrarowheight}{4pt} 
\setlength{\tabcolsep}{3pt} 

          \tiny
\centering

\begin{minipage}[t]{0.23\linewidth}
\centering
\includegraphics[width=\linewidth]{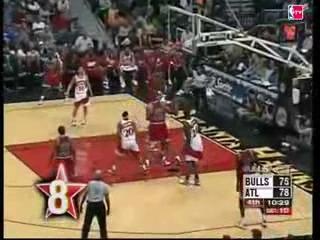}
\setlength{\tabcolsep}{0pt}
\begin{tabularx}{\linewidth}[t]{ZZ}
\rowcolor{Gray!40} \textbf{Ours} & \textbf{CLIP} \\
\hline
\rowcolor{Apricot!100} \emph{Basketball Dunk} & $\downarrow$ \\
\hline
\rowcolor{Apricot!80} \emph{an athlete sprinting to the rim and slamming the ball with a two-handed dunk.} & $\downarrow$ \\
\hline
\rowcolor{Apricot!60} \emph{a skateboarder rolling down a street and popping an ollie over a curb.} & $\downarrow$ \\
\hline
\textbf{``entity''} & \texttt{\textbf{[ROOT]}} \\
\hline
\end{tabularx}
\end{minipage}
\hfill
\begin{minipage}[t]{0.23\linewidth}
\centering
\includegraphics[width=\linewidth]{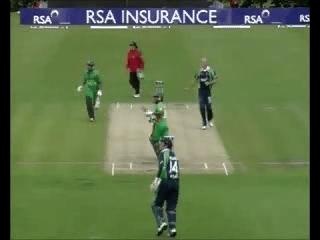}
\setlength{\tabcolsep}{0pt}
\begin{tabularx}{\linewidth}[t]{ZZ}
\rowcolor{Gray!40} \textbf{Ours} & \textbf{CLIP} \\
\hline
\rowcolor{Apricot!100} \emph{Cricket Bowling} & \emph{individual\_sport} \\
\hline
\rowcolor{Apricot!80} \emph{individual\_sport} & $\downarrow$ \\
\hline
\rowcolor{Apricot!60} \emph{a person enjoying a light outdoor pastime.} & $\downarrow$ \\
\hline
\rowcolor{Apricot!60} \emph{a bowler approaching the crease and delivering a seam-up ball toward the wicket.} & $\downarrow$ \\
\hline
\textbf{``entity''} & \texttt{\textbf{[ROOT]}} \\
\hline
\end{tabularx}
\end{minipage}
\hfill
\begin{minipage}[t]{0.23\linewidth}
\centering
\includegraphics[width=\linewidth]{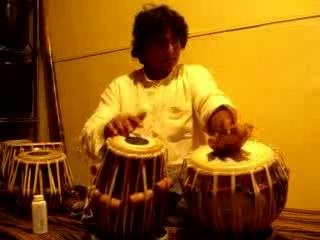}
\setlength{\tabcolsep}{0pt}
\begin{tabularx}{\linewidth}[t]{ZZ}
\rowcolor{Gray!40} \textbf{Ours} & \textbf{CLIP} \\
\hline
\rowcolor{Apricot!100} \emph{Playing Tabla} & \emph{Playing Tabla} \\
\hline
\rowcolor{Apricot!80} \emph{a tabla player producing distinct bols with fingertips and palms.} & $\downarrow$ \\
\hline
\rowcolor{Apricot!60} \emph{Playing Sitar} & $\downarrow$ \\
\hline
\rowcolor{Apricot!40} \emph{a performer beating the double-headed dhol drum with sticks.} & $\downarrow$ \\
\hline
\textbf{``entity''} & \texttt{\textbf{[ROOT]}} \\
\hline
\end{tabularx}
\end{minipage}
\hfill
\begin{minipage}[t]{0.23\linewidth}
\centering
\includegraphics[width=\linewidth]{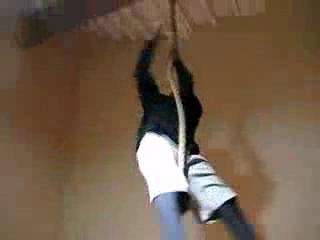}
\setlength{\tabcolsep}{0pt}
\begin{tabularx}{\linewidth}[t]{ZZ}
\rowcolor{Gray!40} \textbf{Ours} & \textbf{CLIP} \\
\hline
\rowcolor{Apricot!100} \emph{Rope Climbing} & \emph{Rock Climbing Indoor} \\
\hline
\rowcolor{Apricot!80} \emph{Rock Climbing Indoor} & $\downarrow$ \\
\hline
\rowcolor{Apricot!60} \emph{an athlete climbing a thick rope using legs and arms in a gym.} & $\downarrow$ \\
\hline
\rowcolor{Apricot!40} \emph{a skier carving turns down a snowy slope with poles in hand.} & $\downarrow$ \\
\hline
\textbf{``entity''} & \texttt{\textbf{[ROOT]}} \\
\hline
\end{tabularx}
\end{minipage}

\begin{minipage}[h]{0.23\linewidth}
\centering
\includegraphics[width=\linewidth]{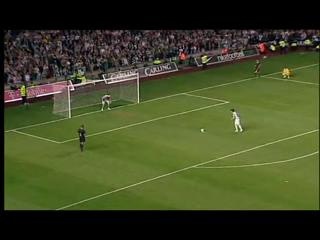}
\setlength{\tabcolsep}{0pt}
\begin{tabularx}{\linewidth}[t]{ZZ}
\rowcolor{Gray!40} \textbf{Ours} & \textbf{CLIP} \\
\hline
\rowcolor{Apricot!100} \emph{Soccer Penalty} & \emph{Soccer Penalty} \\
\hline
\rowcolor{Apricot!80} \emph{Field Hockey Penalty} & $\downarrow$ \\
\hline
\rowcolor{Apricot!60} \emph{a player taking a penalty stroke and aiming low into the corner of the goal.} & $\downarrow$ \\
\hline
\rowcolor{Apricot!40} \emph{a person practicing a sport indoors.} & $\downarrow$ \\
\hline
\textbf{``entity''} & \texttt{\textbf{[ROOT]}} \\
\hline
\end{tabularx}
\end{minipage}
\hfill
\begin{minipage}[h]{0.23\linewidth} 
\centering
\includegraphics[width=\linewidth]{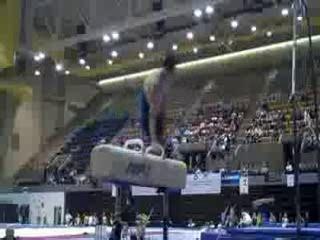}
\setlength{\tabcolsep}{0pt}
\begin{tabularx}{\linewidth}[t]{ZZ}
\rowcolor{Gray!40} \textbf{Ours} & \textbf{CLIP} \\
\hline
\rowcolor{Apricot!100} \emph{Pommel Horse} & \emph{Pommel Horse} \\
\hline
\rowcolor{Apricot!84} \emph{gymnastics} & \emph{Balance Beam} \\
\hline
\rowcolor{Apricot!68} \emph{Balance Beam} & $\downarrow$ \\
\hline
\rowcolor{Apricot!52} \emph{apparatus\_gymnastics} & $\downarrow$ \\
\hline
\rowcolor{Apricot!36} \emph{a gymnast holding a balanced pose on apparatus.} & $\downarrow$ \\
\hline
\textbf{``entity''} & \texttt{\textbf{[ROOT]}} \\
\hline
\end{tabularx}
\end{minipage}
\hfill
\begin{minipage}[h]{0.23\linewidth}
\centering
\includegraphics[width=\linewidth]{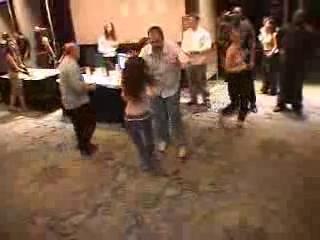}
\setlength{\tabcolsep}{0pt}
\begin{tabularx}{\linewidth}[t]{ZZ}
\rowcolor{Gray!40} \textbf{Ours} & \textbf{CLIP} \\
\hline
\rowcolor{Apricot!100} \emph{Salsa Spin} & \emph{Salsa Spin} \\
\hline
\rowcolor{Apricot!80} \emph{a dancer spinning quickly while maintaining salsa timing with a partner.} & \emph{a person jumping with arms and legs spreading in a rhythmic exercise.} \\
\hline
\rowcolor{Apricot!40} \emph{a person jumping with arms and legs spreading in a rhythmic exercise.} & $\downarrow$ \\
\hline
\textbf{``entity''} & \texttt{\textbf{[ROOT]}} \\
\hline
\end{tabularx}
\end{minipage}
\hfill
\begin{minipage}[h]{0.23\linewidth}
\centering
\includegraphics[width=\linewidth]{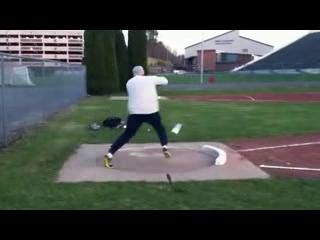}
\setlength{\tabcolsep}{0pt}
\begin{tabularx}{\linewidth}[t]{ZZ}
\rowcolor{Gray!40} \textbf{Ours} & \textbf{CLIP} \\
\hline
\rowcolor{Apricot!100} \emph{Shotput} & \emph{Shotput} \\
\hline
\rowcolor{Apricot!80} \emph{Javelin Throw} & $\downarrow$ \\
\hline
\rowcolor{Apricot!60} \emph{a person warming up before exercise.} & $\downarrow$ \\
\hline
\rowcolor{Apricot!60} \emph{an athlete practicing a sport on a field or court.} & $\downarrow$ \\
\hline
\textbf{``entity''} & \texttt{\textbf{[ROOT]}} \\
\hline
\end{tabularx}
\end{minipage}

\caption{\small {Image traversals with \name and CLIP:} \mame's path through hyperbolic space reveals a smooth transition from specific to generic text descriptions, reflecting a more systematic and fine-grained visual-textual semantic hierarchy than CLIP.}
\label{fig:app_travel}
\end{figure*}

\blue{Hammer Throw}:
\begin{displayquote}
	{\bf Q}: Could you analyze ``Hammer Throw'' by providing a series of descriptions that start as very specific and become progressively more general? Alongside this, please also provide a list of related adjectives, similarly ordered from specific to general. \\
	{\bf A}: An athlete spinning within a circle and releasing the hammer for maximum distance. An athlete accelerating down the runway and launching in a field event. A person practicing a sport outdoors. A person performing a physical activity. Adjectives: powerful, explosive, technical, coordinated, solo, athletic, outdoor.
\end{displayquote}

\blue{Knitting}:
\begin{displayquote}
	{\bf Q}: Could you analyze ``Knitting'' by providing a series of descriptions that start as very specific and become progressively more general? Alongside this, please also provide a list of related adjectives, similarly ordered from specific to general.\\
	{\bf A}: Hands looping yarn with knitting needles to form a patterned fabric. Hands crafting with yarn, fabric, or small tools. Routine indoor activity using common tools. A person engaged in everyday human activity. Adjectives: precise, controlled, rhythmic, handheld, quiet, indoor, recreational.
\end{displayquote}

\section{Quantitative Verification of Hierarchical Structure}
  \label{sec:app_hierarchy_verification}

To quantitatively verify that the learned hyperbolic embeddings faithfully capture the intended hierarchical
structure, we analyze the geometric properties of the representations on the test sets of CIFAR-100 (B0\,Inc10),
CUB-200 (B0\,Inc20), and UCF-101 (B0\,Inc10). Crucially, {all features are extracted after the model has completed the
final incremental learning stage}, demonstrating \mame's robustness against catastrophic forgetting across the entire
task sequence.

Following the Lorentz model adopted in Section~4.2, we define the \emph{radius} of an embedding
$\mathbf{x}\in\mathcal{B}^d$ as its geodesic distance to the origin $\mathbf{o}=[\mathbf{0},1]$
via Eq. 4, \ie, $r(\mathbf{x})=d_{\mathcal{B}}(\mathbf{x},\mathbf{o})$. In a well-structured
hyperbolic hierarchy, the entailment cone mechanism constrains each child to lie within the cone of its parent, which
geometrically pushes children to higher-radius regions. Consequently, concepts should expand outward radially as they
become more specific, and semantically related nodes should cluster tightly.

As reported in Table~\ref{tab:hierarchy_metrics}, \name strictly preserves these geometric expectations across all
datasets, even after the full sequence of incremental updates:
\begin{enumerate}[leftmargin=1.5em,itemsep=2pt]
  \item \textbf{Radial Expansion by Specificity}: The average radius increases with tree depth across all
datasets. For instance, on the deep CUB-200 hierarchy the radius expands monotonically from $0.10$ at Depth~1 to
$0.83$ at Depth~5; UCF-101 smoothly expands from $0.14$ at Depth~1 to $0.78$ at Depth~6. Furthermore, decomposing the
radius by node type reveals a consistent ordering: Virtual parent nodes (abstract categories) reside closest to the
origin, Real leaf-class prototypes lie at intermediate radii, and Image-level sample embeddings are mapped nearest to
the boundary (\emph{Virtual}~$<$~\emph{Real}~$<$~\emph{Image} across all datasets). This directly reflects the
entailment cone geometry, where the cone aperture $\mathrm{aper}(\mathbf{x})$ shrinks with increasing
$\|\mathbf{x}\|$, causing deeper and more specific nodes to occupy progressively higher-radius regions of
$\mathcal{B}^d$.
  \item \textbf{Semantic Topology Preservation}: Beyond the radial ordering, we verify structural integrity by
comparing the average pairwise geodesic distance between sibling nodes (sharing the same parent in the semantic tree)
and non-relative nodes. Across all datasets, siblings are consistently embedded closer together than non-relatives
(e.g., $0.27$ vs.\ $0.32$ on CIFAR-100; $0.42$ vs.\ $0.47$ on UCF-101), confirming that the learned Lorentz embedding
preserves the semantic topology throughout the incremental learning process.
\end{enumerate}
These results confirm that \name effectively translates the discrete partial-order semantic tree into a continuous,
geometrically faithful hierarchical representation, and successfully maintains this structure across sequential tasks
without collapsing.

  \begin{table*}[htbp]
      \centering
      \setlength{\tabcolsep}{6pt}
      \renewcommand{\arraystretch}{1.2}
      \caption{Quantitative hierarchy verification computed {after the final incremental task}. The radius is defined as
   the geodesic distance to the Lorentz origin $\mathbf{o}=[\mathbf{0},1]$ via Eq. 4. Structural
  distances compare the pairwise geodesic proximity of sibling nodes versus non-relative nodes. (A dash (–) indicates that the dataset tree does not reach this depth.)}
      \label{tab:hierarchy_metrics}
      \resizebox{\textwidth}{!}{
      \begin{tabular}{l | cccccc | ccc | cc}
          \toprule
          \multirow{2}{*}{\textbf{Dataset}} & \multicolumn{6}{c|}{\textbf{Average Radius by Tree Depth}} &
  \multicolumn{3}{c|}{\textbf{Average Radius by Node Type}} & \multicolumn{2}{c}{\textbf{Structural Distance}} \\
          \cmidrule(lr){2-7} \cmidrule(lr){8-10} \cmidrule(lr){11-12}
          & D\,1 & D\,2 & D\,3 & D\,4 & D\,5 & D\,6 & Virtual & Real & Image & Sibling & Non-Rel \\
          \midrule
          \textbf{CIFAR-100} & 0.13 & 0.29 & 0.29 & 0.67 & 0.68 & --   & 0.26 & 0.30 & 0.68 & 0.27 & 0.32 \\
          \textbf{CUB-200}   & 0.10 & 0.12 & 0.48 & 0.80 & 0.83 & --   & 0.32 & 0.53 & 0.82 & 0.51 & 0.54 \\
          \textbf{UCF-101}   & 0.14 & 0.17 & 0.35 & 0.72 & 0.75 & 0.78 & 0.25 & 0.44 & 0.75 & 0.42 & 0.47 \\
          \bottomrule
      \end{tabular}
      }
  \end{table*}

\section{Broader Impacts}
\label{sec:app_broader_impacts}
\name studies how hierarchical information can be used to improve class-incremental learning, enabling models to learn new classes while better preserving previously learned knowledge. Since the method is exemplar-free, it avoids storing past training samples, which may benefit privacy-sensitive or storage-constrained continual learning scenarios. By using semantic hierarchies as structured supervision, \name provides a way to incorporate class relationships into continual learning without requiring manually stored exemplars. The main practical consideration is that the semantic hierarchy should be reliable for the target domain, since inaccurate hierarchy construction may affect downstream predictions. In real deployments, this can be mitigated by validating the generated hierarchy, using curated domain taxonomies when available, or incorporating light human review for high-stakes applications.

\newpage

%% file: table/fig2.tex
\begin{figure*}[t]
	\centering
	\begin{subfigure}{0.32\linewidth}
		\includegraphics[width=1\columnwidth]{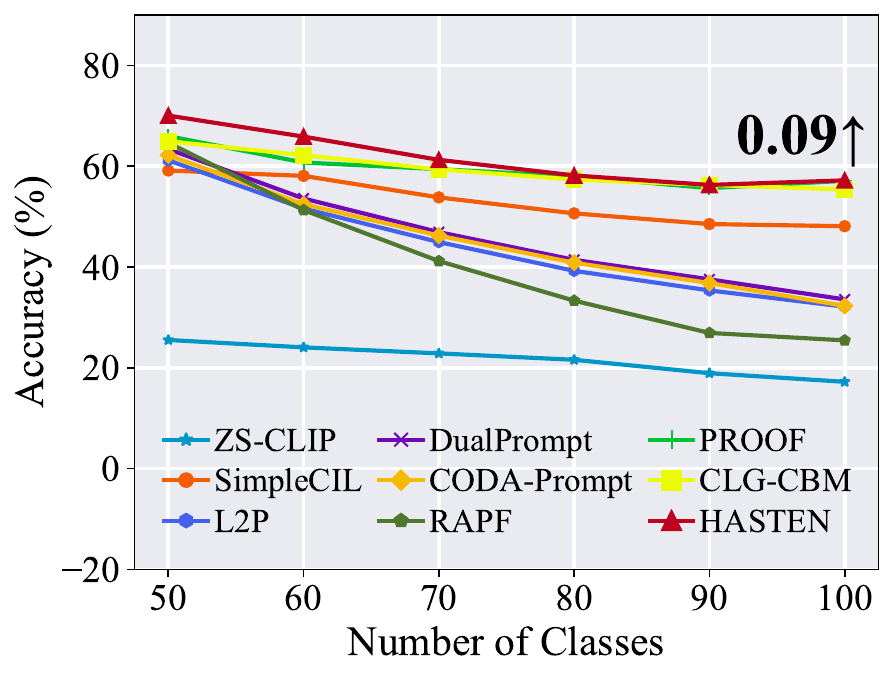}
		\caption{Aircraft Base50 Inc10}
		\label{fig:benchmark-aircraft50}
	\end{subfigure}
	\hfill
	\begin{subfigure}{0.32\linewidth}
		\includegraphics[width=1\linewidth]{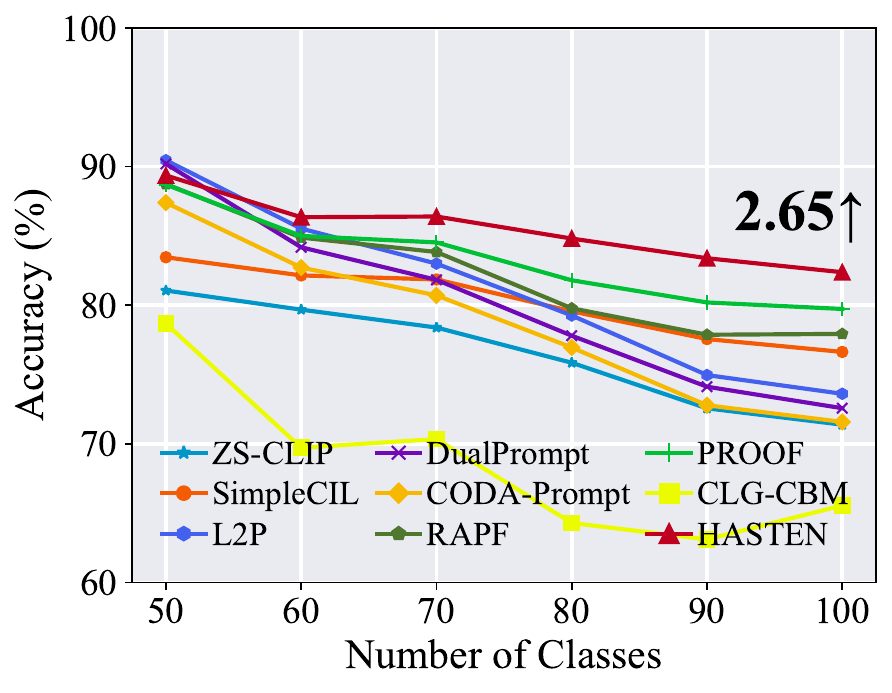}
		\caption{CIFAR100 Base50 Inc10}
		\label{fig:benchmark-cifar50}
	\end{subfigure}
	\hfill
	\begin{subfigure}{0.32\linewidth}
		\includegraphics[width=1\linewidth]{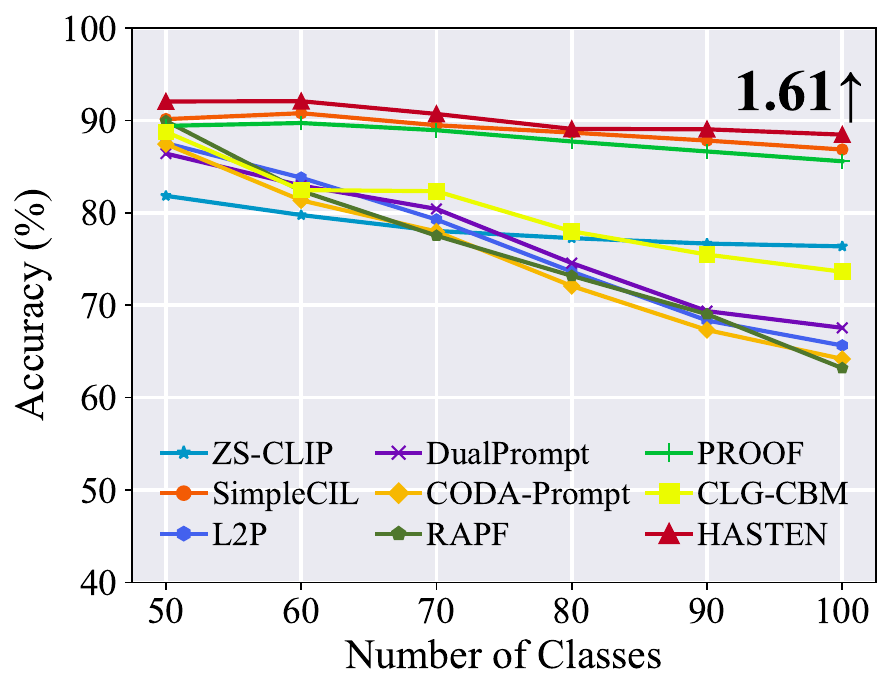}
		\caption{Cars Base50 Inc10}
		\label{fig:benchmark-cars50}
	\end{subfigure}
	\\
	\begin{subfigure}{0.32\linewidth}
		\includegraphics[width=1\linewidth]{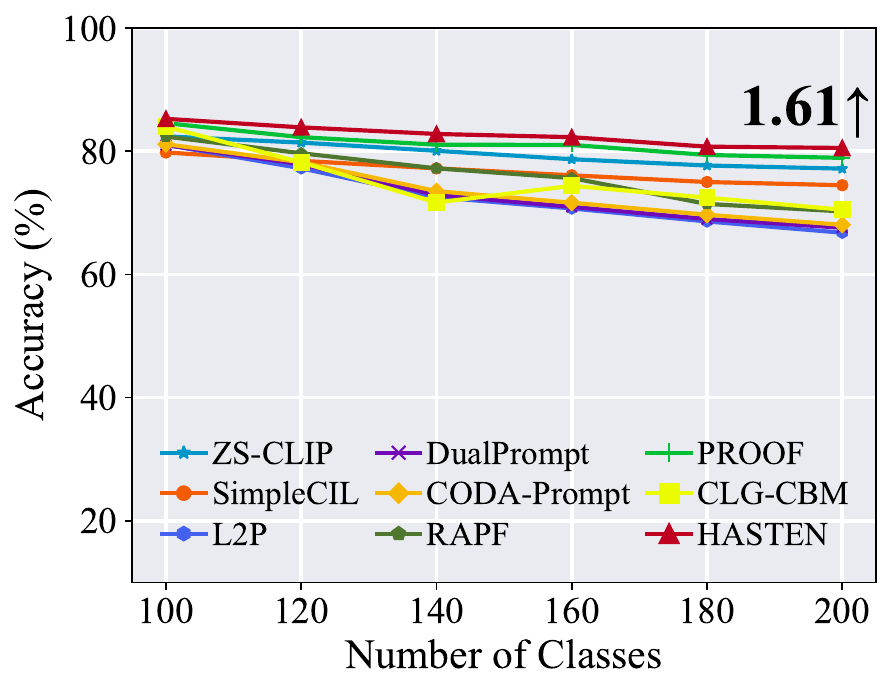}
		\caption{ImageNet-R Base100 Inc20}
		\label{fig:benchmark-imagenetr100}
	\end{subfigure}
	\hfill
	\begin{subfigure}{0.32\linewidth}
		\includegraphics[width=1\linewidth]{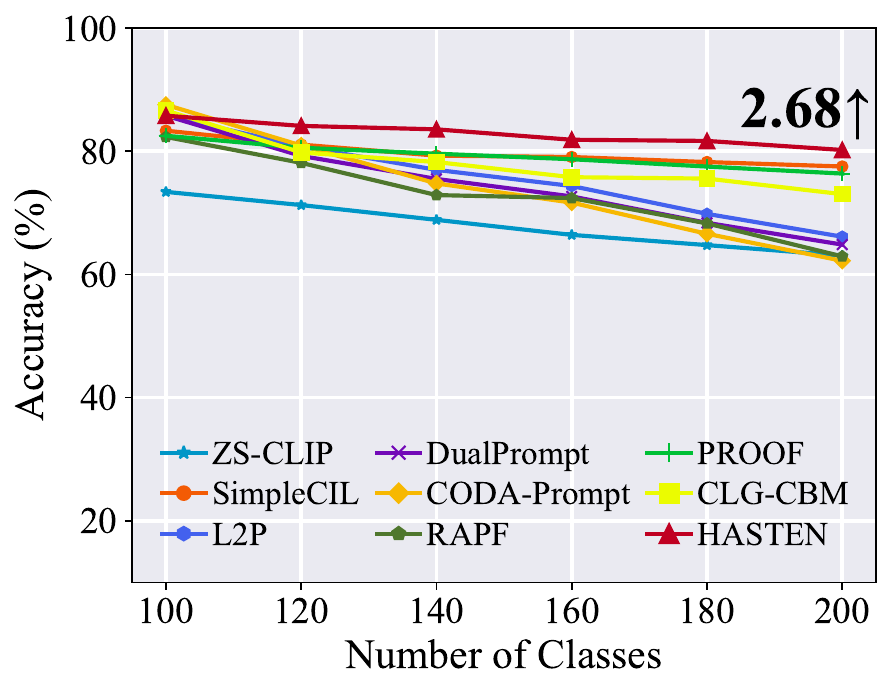}
		\caption{CUB Base100 Inc20}
		\label{fig:benchmark-cub100}
	\end{subfigure}
	\hfill
	\begin{subfigure}{0.32\linewidth}
		\includegraphics[width=1\columnwidth]{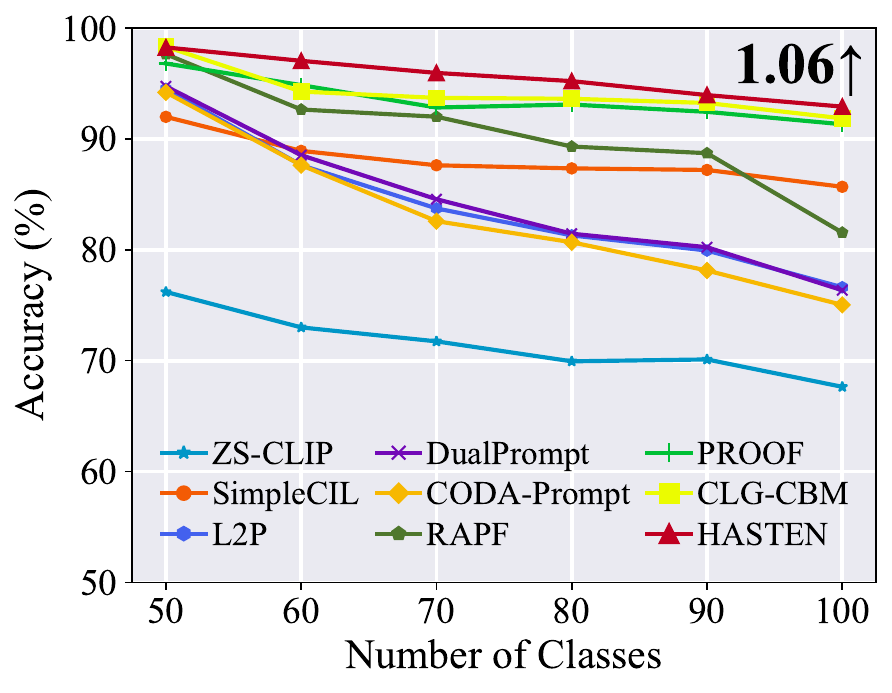}
		\caption{UCF Base50 Inc10}
		\label{fig:benchmark-ucf50}
	\end{subfigure}
	\\
	\begin{subfigure}{0.32\linewidth}
		\includegraphics[width=1\linewidth]{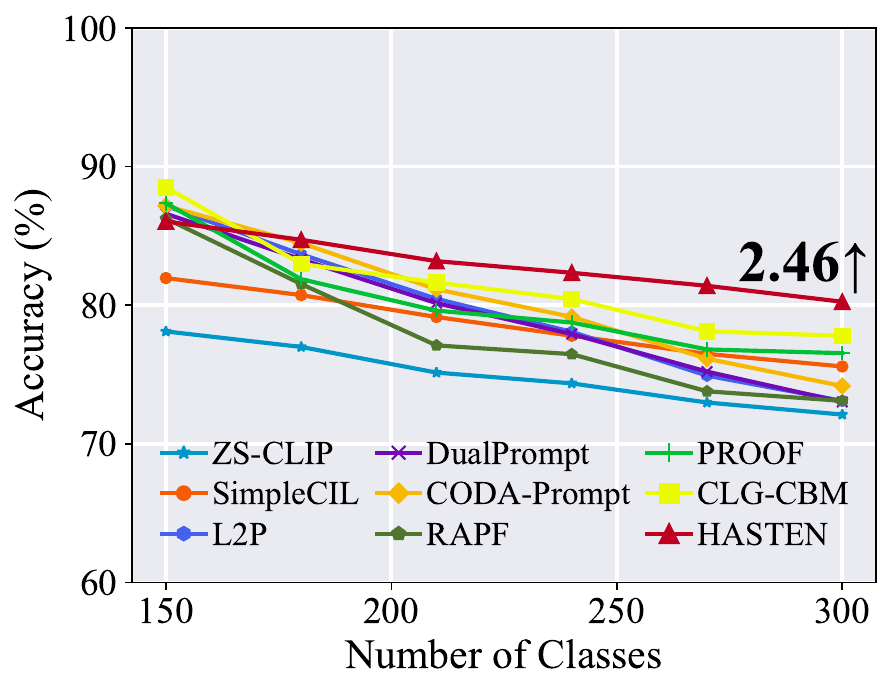}
		\caption{SUN Base150 Inc30}
		\label{fig:benchmark-sun150}
	\end{subfigure}
	\hfill
	\begin{subfigure}{0.32\linewidth}
		\includegraphics[width=1\linewidth]{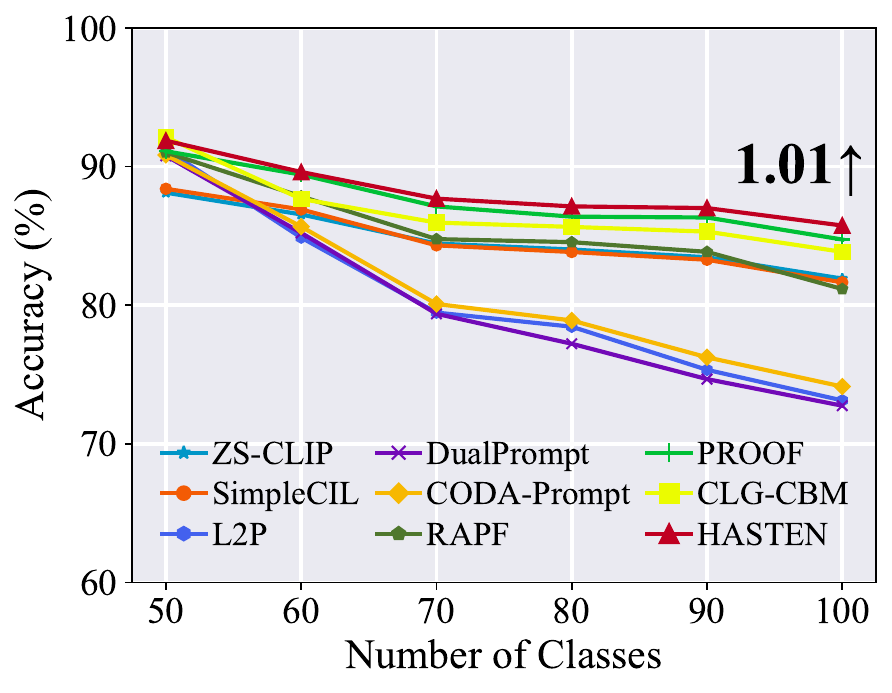}
		\caption{Food Base50 Inc10}
		\label{fig:benchmark-food50}
	\end{subfigure}
	\hfill
	\begin{subfigure}{0.32\linewidth}
		\includegraphics[width=1\columnwidth]{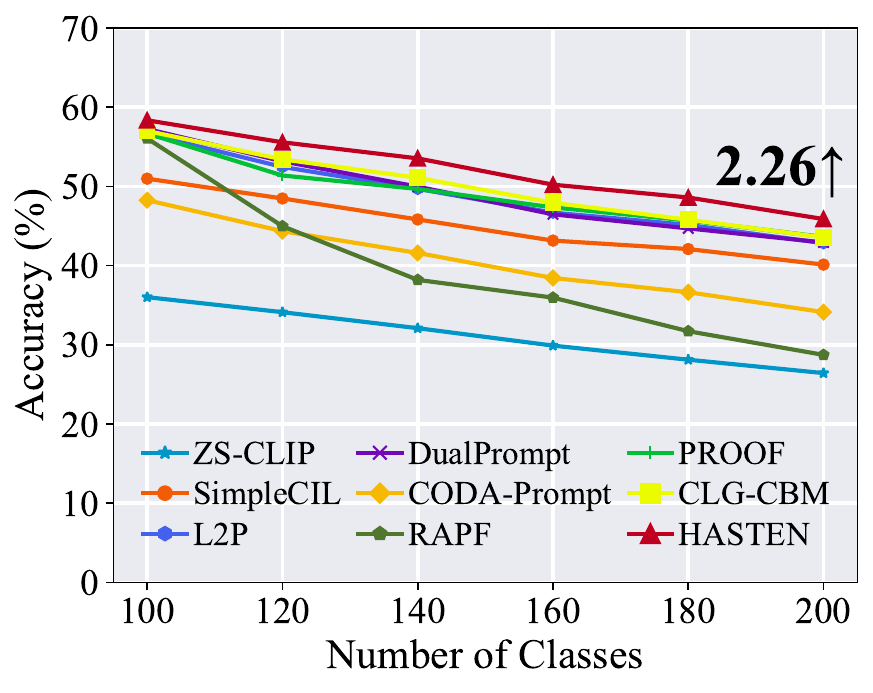}
		\caption{ObjectNet Base100 Inc20}
		\label{fig:benchmark-objectnet100}
	\end{subfigure}
	\caption{ 	Incremental performance of different methods on half-base setting. We report the performance gap after the last incremental stage of \name and the runner-up method at the end of the line. All methods use the same CLIP pre-trained weights. }
	\label{fig:supp-benchmark-b50}
\end{figure*}

%% file: table/fig1.tex
\begin{figure}[t]
	\centering
	\begin{subfigure}{0.32\linewidth}
		\includegraphics[width=1\columnwidth]{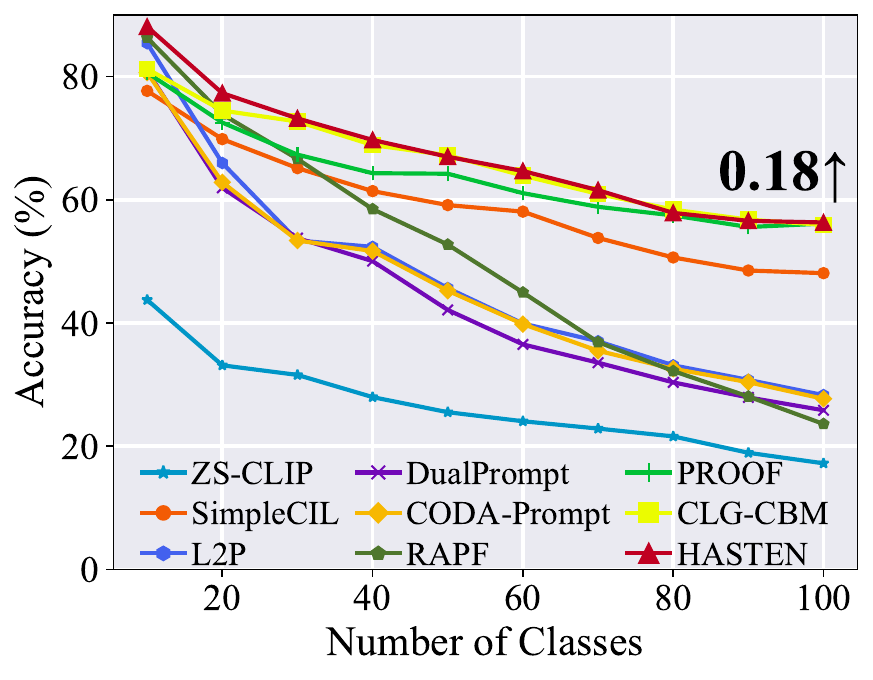}
		\caption{Aircraft Base0 Inc10}
	\end{subfigure}
	\hfill
	\begin{subfigure}{0.32\linewidth}
		\includegraphics[width=1\linewidth]{pics/cifar}
		\caption{CIFAR100 Base0 Inc10}
	\end{subfigure}
	\hfill
	\begin{subfigure}{0.32\linewidth}
		\includegraphics[width=1\linewidth]{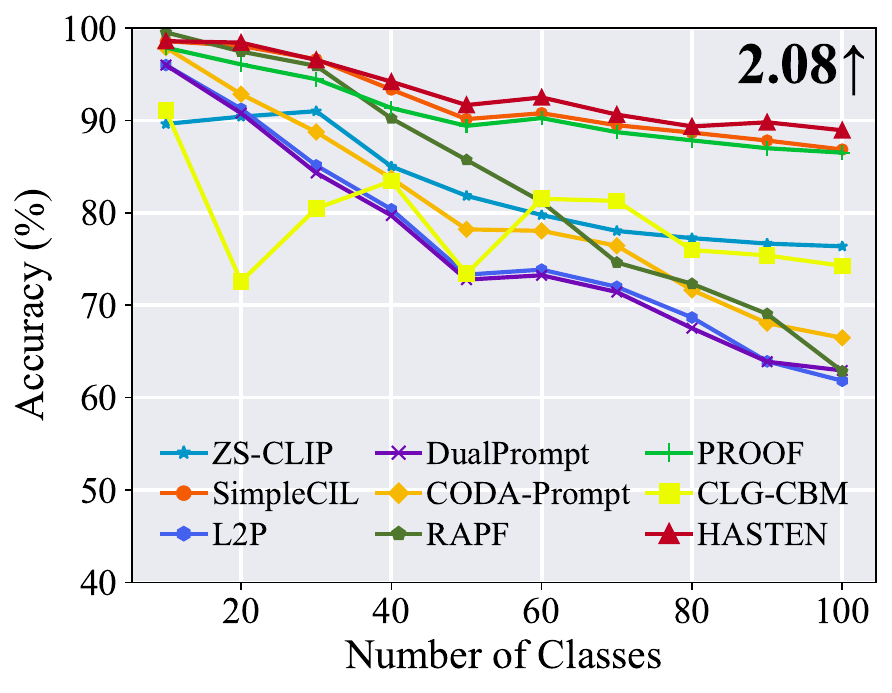}
		\caption{Cars Base0 Inc10}
	\end{subfigure}
	\\
	\begin{subfigure}{0.32\linewidth}
		\includegraphics[width=1\linewidth]{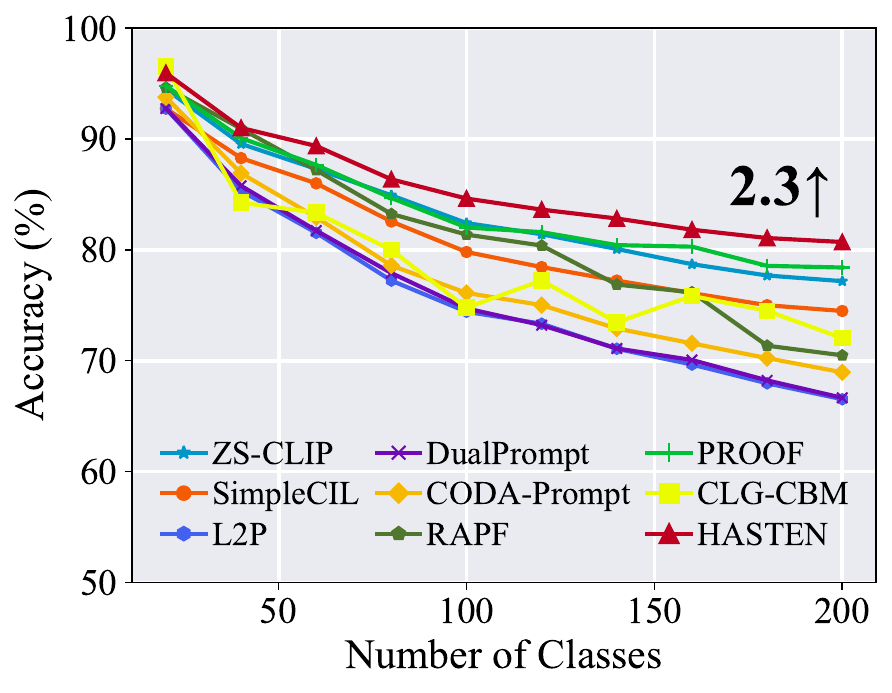}
		\caption{ImageNet-R Base0 Inc20}
	\end{subfigure}
	\hfill
	\begin{subfigure}{0.32\linewidth}
		\includegraphics[width=1\linewidth]{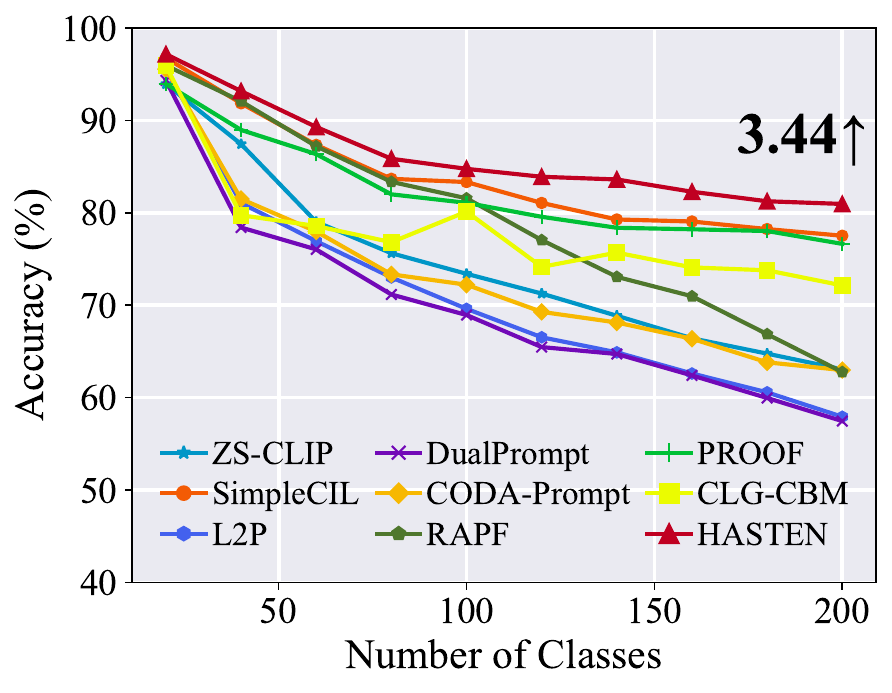}
		\caption{CUB Base0 Inc20}
	\end{subfigure}
	\hfill
	\begin{subfigure}{0.32\linewidth}
		\includegraphics[width=1\columnwidth]{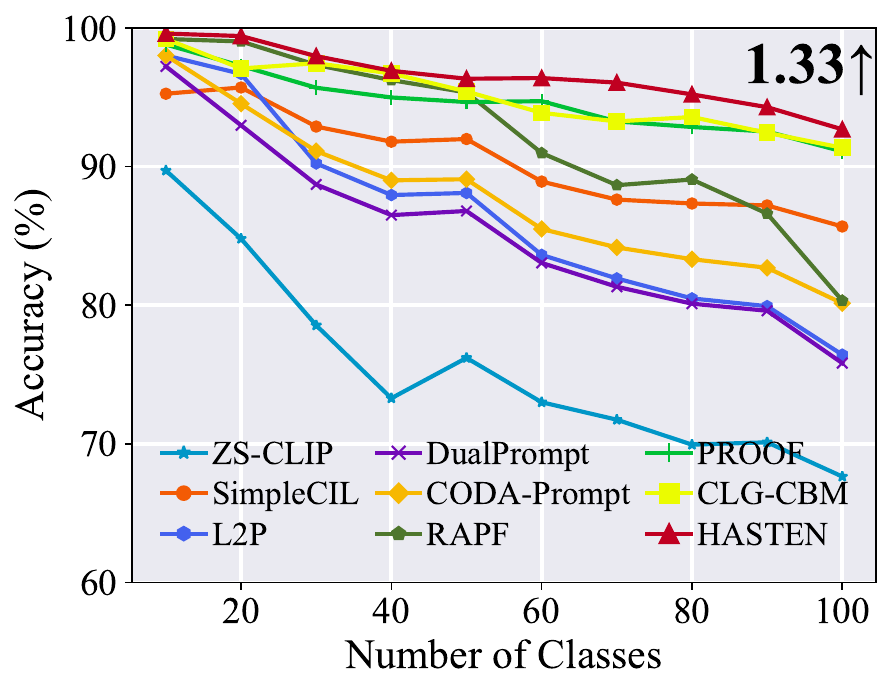}
		\caption{UCF Base0 Inc10}
	\end{subfigure}
	\\
	\begin{subfigure}{0.32\linewidth}
		\includegraphics[width=1\linewidth]{pics/sun}
		\caption{SUN Base0 Inc30}
	\end{subfigure}
	\hfill
	\begin{subfigure}{0.32\linewidth}
		\includegraphics[width=1\linewidth]{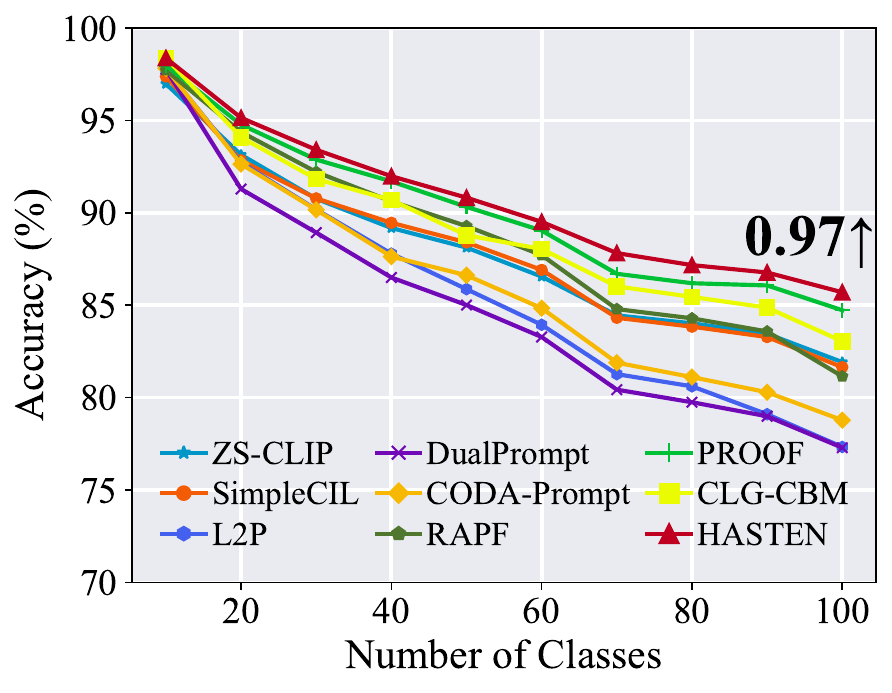}
		\caption{Food Base0 Inc10}
	\end{subfigure}
	\hfill
	\begin{subfigure}{0.32\linewidth}
		\includegraphics[width=1\columnwidth]{pics/objectnet}
		\caption{ObjectNet Base0 Inc20}
	\end{subfigure}
	\caption{ 	Incremental performance of different methods on B0 setting. We report the performance gap after the last incremental stage of \name and the runner-up method at the end of the line.    All methods use the same CLIP pre-trained weights. }
	\label{fig:supp-benchmark-b0}
\end{figure}

%% file: checklist.tex
\section*{NeurIPS Paper Checklist}

\begin{enumerate}

\item {\bf Claims}
    \item[] Question: Do the main claims made in the abstract and introduction accurately reflect the paper's contributions and scope?
    \item[] Answer: \answerYes{} 
    \item[] Justification: We state the contributions and scope in the abstract and introduction, and discuss limitations in Section~6.

\item {\bf Limitations}
    \item[] Question: Does the paper discuss the limitations of the work performed by the authors?
    \item[] Answer: \answerYes{} 
    \item[] Justification: We included it in the paper.

\item {\bf Theory assumptions and proofs}
    \item[] Question: For each theoretical result, does the paper provide the full set of assumptions and a complete (and correct) proof?
    \item[] Answer: \answerYes{}.
    \item[] Justification: We state a standard geometric property of Lorentz hyperbolic space and provide its derivation in Appendix~\ref{app:hyperbolic_hierarchy} to justify its use for hierarchical representation.

    \item {\bf Experimental result reproducibility}
    \item[] Question: Does the paper fully disclose all the information needed to reproduce the main experimental results of the paper to the extent that it affects the main claims and/or conclusions of the paper (regardless of whether the code and data are provided or not)?
    \item[] Answer: \answerYes{} 
    \item[] Justification: We provide all the necessary details to reproduce the results. Detailed algorithmic descriptions are provided in Appendix~\ref{sec:alg}.

\item {\bf Open access to data and code}
    \item[] Question: Does the paper provide open access to the data and code, with sufficient instructions to faithfully reproduce the main experimental results, as described in supplemental material?
    \item[] Answer: \answerYes{} 
    \item[] Justification: The source code is publicly available at: \url{https://github.com/LAMDA-CL/NeurIPS2026-HASTEN}.

\item {\bf Experimental setting/details}
    \item[] Question: Does the paper specify all the training and test details (e.g., data splits, hyperparameters, how they were chosen, type of optimizer) necessary to understand the results?
    \item[] Answer: \answerYes{} 
    \item[] Justification: The experiment details are available in the paper.

\item {\bf Experiment statistical significance}
    \item[] Question: Does the paper report error bars suitably and correctly defined or other appropriate information about the statistical significance of the experiments?
    \item[] Answer: \answerYes{} 
    \item[] Justification: We report mean accuracy with one-standard-deviation bands over multiple random class-order splits in the appendix. The variability reflects different independent runs under different class orders.

\item {\bf Experiments compute resources}
    \item[] Question: For each experiment, does the paper provide sufficient information on the computer resources (type of compute workers, memory, time of execution) needed to reproduce the experiments?
    \item[] Answer: \answerYes{} 
    \item[] Justification: The information of computer resources is provided.
    
\item {\bf Code of ethics}
    \item[] Question: Does the research conducted in the paper conform, in every respect, with the NeurIPS Code of Ethics \url{https://neurips.cc/public/EthicsGuidelines}?
    \item[] Answer: \answerYes{} 
    \item[] Justification: The research conforms to the NeurIPS Code of Ethics.

\item {\bf Broader impacts}
    \item[] Question: Does the paper discuss both potential positive societal impacts and negative societal impacts of the work performed?
    \item[] Answer: \answerYes{} 
    \item[] Justification: We discuss both positive and negative societal impacts of the work in the appendix.
    
\item {\bf Safeguards}
    \item[] Question: Does the paper describe safeguards that have been put in place for responsible release of data or models that have a high risk for misuse (e.g., pre-trained language models, image generators, or scraped datasets)?
    \item[] Answer: \answerNA{} 
    \item[] Justification: No data or models with a high risk of misuse are released.

\item {\bf Licenses for existing assets}
    \item[] Question: Are the creators or original owners of assets (e.g., code, data, models), used in the paper, properly credited and are the license and terms of use explicitly mentioned and properly respected?
\item[] Answer: \answerYes{}
\item[] Justification: All the datasets used are publicly available, discussed, and properly cited in Section~\ref{sec:exp} and Appendix~\ref{app:sub_method}.

\item {\bf New assets}
    \item[] Question: Are new assets introduced in the paper well documented and is the documentation provided alongside the assets?
    \item[] Answer: \answerNA{} 
    \item[] Justification: The paper does not release new datasets, models, or benchmarks.

\item {\bf Crowdsourcing and research with human subjects}
    \item[] Question: For crowdsourcing experiments and research with human subjects, does the paper include the full text of instructions given to participants and screenshots, if applicable, as well as details about compensation (if any)? 
    \item[] Answer: \answerNA{} 
    \item[] Justification: The paper does not involve crowdsourcing nor research with human subjects.

\item {\bf Institutional review board (IRB) approvals or equivalent for research with human subjects}
    \item[] Question: Does the paper describe potential risks incurred by study participants, whether such risks were disclosed to the subjects, and whether Institutional Review Board (IRB) approvals (or an equivalent approval/review based on the requirements of your country or institution) were obtained?
    \item[] Answer: \answerNA{} 
    \item[] Justification: The paper does not involve human-subject research.

\item {\bf Declaration of LLM usage}
    \item[] Question: Does the paper describe the usage of LLMs if it is an important, original, or non-standard component of the core methods in this research? Note that if the LLM is used only for writing, editing, or formatting purposes and does \emph{not} impact the core methodology, scientific rigor, or originality of the research, declaration is not required.
    \item[] Answer: \answerYes{} 
    \item[] Justification: We describe the use of LLMs for semantic tree construction in Section~\ref{subsec:semantic_tree} and Appendix~\ref{sec:app_prompt}.

\end{enumerate}